\documentclass[10pt,journal,compsoc]{IEEEtran}
\usepackage[breaklinks=true,colorlinks,bookmarks=true]{hyperref}
\usepackage{xspace}
\usepackage{graphicx}
\usepackage{booktabs}
\usepackage{multirow}
\usepackage{color}
\usepackage{colortbl}
\usepackage{amssymb}
\usepackage{amsmath}
\usepackage{amsthm}
\usepackage{cite}
\usepackage[switch]{lineno} 
\usepackage{makecell}
\usepackage{tabularx}
\usepackage{hyperref}
\hypersetup{
    colorlinks=true,
    linkcolor=blue,
    citecolor=blue,
    urlcolor=blue
}
\usepackage{subcaption}
\usepackage{caption}

\usepackage{url}

\newcommand{\cmark}{\ding{51}}
\newcommand{\xmark}{\ding{55}}

\usepackage[table]{xcolor}
\usepackage[section]{placeins}
\definecolor{mygray}{gray}{.92}
\definecolor{mycyan}{cmyk}{.3,0,0,0}
\definecolor{LightCyan}{rgb}{0.95,1,1}
\definecolor{darkgreen}{RGB}{0,100,0}

\definecolor{orange}{rgb}{1,0.8,0.6}
\definecolor{yellow}{rgb}{1,1,0.6}

\usepackage{xcolor}
\usepackage{pifont}

\definecolor{mygray}{gray}{.9}

\usepackage{array}
\newcolumntype{x}[1]{>{\centering\arraybackslash\hspace{0pt}}p{#1}}

\ifCLASSINFOpdf
\else
\fi

\newcommand{\xiao}[1]{{\color{purple}[xiao: #1]}}

\newcommand{\boldparagraph}[1]

\usepackage{booktabs}
\usepackage{threeparttable}
\usepackage{makecell}
\usepackage{multirow}
\usepackage{graphicx}
\usepackage{adjustbox}
\usepackage{rotating}
\usepackage{array}
\usepackage{overpic}
\usepackage{enumitem}
\usepackage{longtable,xcolor,colortbl}
\usepackage{threeparttablex}
\definecolor{LightGray}{gray}{0.9}
\definecolor{Gray}{gray}{0.7}
\definecolor{LightGray}{gray}{0.92}
\definecolor{LightGreen}{RGB}{0, 111, 0}
\definecolor{LightRed}{RGB}{200, 0, 0}

\begin{document}
\title{A Survey: Learning Embodied Intelligence from Physical Simulators and World Models}

\author{\normalsize{
Xiaoxiao Long$^*$, Qingrui Zhao$^*$, Kaiwen Zhang$^*$, Zihao Zhang$^*$, Dingrui Wang$^*$, Yumeng Liu$^*$, Zhengjie Shu$^*$, \\ 
Yi Lu$^*$, Shouzheng Wang$^*$, Xinzhe Wei$^*$, Wei Li, Wei Yin, Yao Yao, Jia Pan, Qiu Shen, Ruigang Yang, \\ 
Xun Cao$^{\dag}$, Qionghai Dai
\IEEEcompsocitemizethanks{
\IEEEcompsocthanksitem $^*$ indicates equal contributions. $^\dag$ indicates corresponding authors.
\IEEEcompsocthanksitem Xiaoxiao Long, Qingrui Zhao, Yi Lu, Yao Yao, Qiu Shen, Wei Li and Xun Cao are with the Nanjing University (e-mail: caoxun@nju.edu.cn).
\IEEEcompsocthanksitem Yumeng Liu, Zhengjie Shu and Jia Pan are with the University of Hong Kong.
\IEEEcompsocthanksitem Shouzheng Wang is with the Central South University. 
\IEEEcompsocthanksitem Wei Yin is with the Horizon Robotics.
\IEEEcompsocthanksitem Zihao Zhang and Xinzhe Wei are with the Institute of Computing Technology, Chinese Academy of Sciences (ICT). Xinzhe Wei is also with the University of Chinese Academy of Sciences.
\IEEEcompsocthanksitem Ruigang Yang and Zhengjie Shu are with the Shanghai Jiao Tong University. 
\IEEEcompsocthanksitem Dingrui Wang is with the Technical University of Munich.
\IEEEcompsocthanksitem Kaiwen Zhang and Qionghai Dai are with the Tsinghua University.
}
}
}

% \markboth{IEEE TRANSACTIONS ON PATTERN ANALYSIS AND MACHINE INTELLIGENCE}
% {Long \MakeLowercase{\textit{\etal}}: World Models}

\IEEEtitleabstractindextext{%
\begin{abstract}
The pursuit of artificial general intelligence (AGI) has placed embodied intelligence at the forefront of robotics research. Embodied intelligence focuses on agents capable of perceiving, reasoning, and acting within the physical world.
Achieving robust embodied intelligence requires not only advanced perception and control, but also the ability to ground abstract cognition in real-world interactions. Two foundational technologies, physical simulators and world models, have emerged as critical enablers in this quest. Physical simulators provide controlled, high-fidelity environments for training and evaluating robotic agents, allowing safe and efficient development of complex behaviors. In contrast, world models empower robots with internal representations of their surroundings, enabling predictive planning and adaptive decision-making beyond direct sensory input. This survey systematically reviews recent advances in learning embodied AI through the integration of physical simulators and world models. We analyze their complementary roles in enhancing autonomy, adaptability, and generalization in intelligent robots, and discuss the interplay between external simulation and internal modeling in bridging the gap between simulated training and real-world deployment. By synthesizing current progress and identifying open challenges, this survey aims to provide a comprehensive perspective on the path toward more capable and generalizable embodied AI systems. We also maintain an active repository that contains up-to-date literature and open-source projects at{\url{https://github.com/NJU3DV-LoongGroup/Embodied-World-Models-Survey}}.
\end{abstract}

}
\maketitle
% \IEEEdisplaynontitleabstractindextext
% \IEEEpeerreviewmaketitle

\section{Introduction}
\label{sec:introduction}
% \IEEEPARstart{W}{elcome} to the updated and simplified documentation to using the IEEEtran \LaTeX \ class file. 
%\ZQR{需要一个总图，介绍智能机器人（仿人机器人）以及模拟器、世界模型之间的关系}

\subsection{Overview}

With the rapid advancement of artificial intelligence~\cite{achiam2023gpt,liu2024deepseek} and robotics technology\cite{chi2023diffusion,kim2024openvla}, the interaction between intelligent agents and the physical world has increasingly become a central focus of research. The pursuit of artificial general intelligence (AGI), systems that can match or exceed human cognitive abilities across diverse domains, faces a key question: \textit{how to ground abstract reasoning in real-world understanding and action?} 

\begin{figure}[htb]
	\centering
	\includegraphics[width=\linewidth]{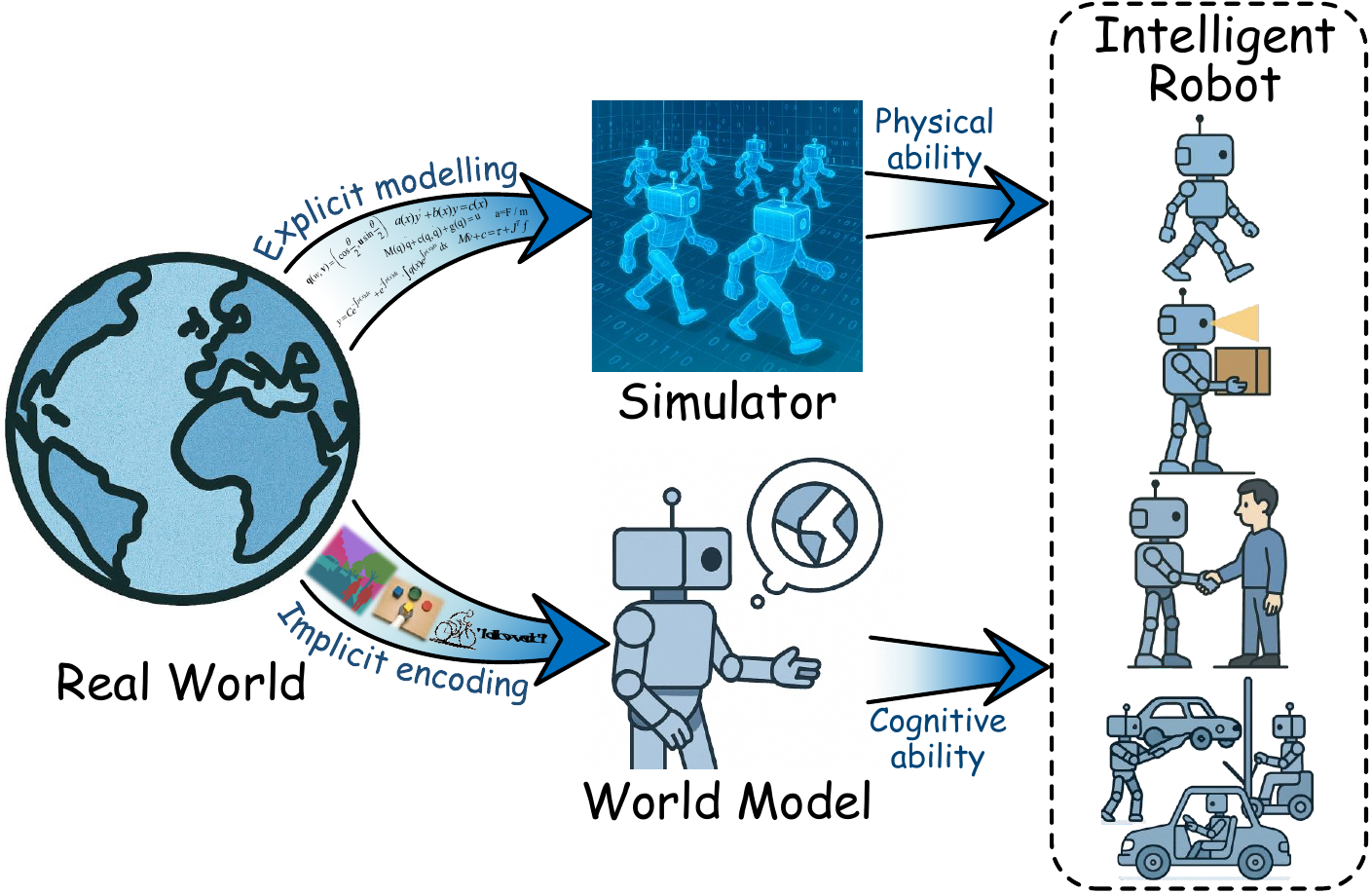}
	\caption{Physical simulator and world model play vital roles for embodied intelligence. Simulator provides an explicit modeling of the real world, offering a controlled environment where robots can train, test, and refine their behaviors. World model offers internal representations of the environment, enabling robots to autonomously simulate, predict, and plan actions within their cognitive framework.}
\end{figure}

Intelligent robots have emerged as essential embodied agents paving the way toward AGI by providing the physical medium that bridges computational intelligence with interaction in real-world environments. Unlike disembodied intelligence systems that operate purely on symbolic or digital data, embodied intelligence emphasizes the importance of perception, action, and cognition through physical interaction with the environment. This paradigm allows robots to adjust their behavior and cognition continuously based on feedback from the physical world while executing tasks, making robotics not merely an application of AI, but a essential component of the path toward general intelligence. 

\begin{figure*}[tp]
	\centering
	\includegraphics[width=\linewidth]{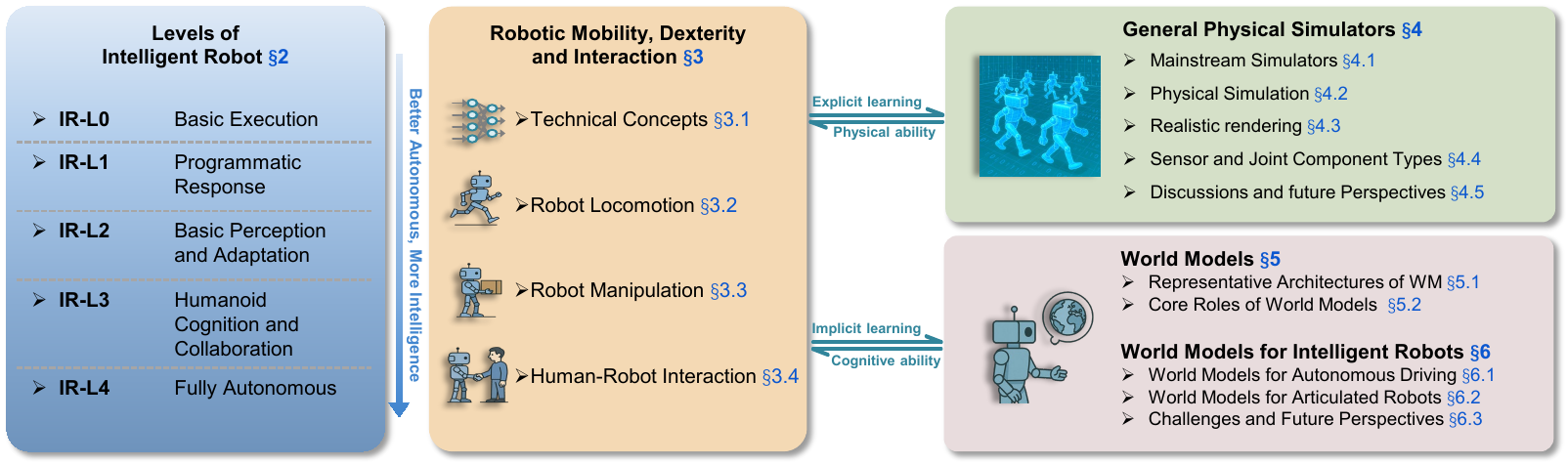}
	\caption{Survey outline: We categorize intelligent robot development into five levels (IR-L0 to IR-L4) and review progress and key techniques in robotic mobility, manipulation, and interaction (§\ref{sec:robots_mdi}), the use of physical simulators for learning and control algorithm verification (§\ref{sec:simulator}), and the design and use of world models as internal representations for learning, planning, and decision-making (§\ref{sec:world_models}–§\ref{sec:world_models_for_AD_robot}), emphasizing both explicit and implicit learning pathways.}
    \label{fig:paper_structure}
\end{figure*}

% The significance of embodied intelligence goes beyond the technicality of physical tasks; it provides a foundation for creating robots that can truly understand and reason about the world in a more human-like manner. The ability to act and perceive within a physical body makes these robots capable of more robust learning, since they can test hypotheses, adjust their strategies, and gain experience through real-time interaction. This physical engagement enhances their problem-solving abilities, as it enables them to continuously refine their understanding of how actions influence outcomes in a complex, uncertain environment. The combination of sensory input, motor control, and cognitive processing forms a closed-loop system that is fundamental for robots to achieve true autonomy and adaptability.

The significance of embodied intelligence extends beyond the execution of physical tasks. By acting and perceiving within a physical body~\cite{tong2024advancements}, robots can robustly learn from experience, test hypotheses, and refine their strategies through ongoing interaction. This closed-loop integration of sensory input, motor control, and cognitive processing forms the foundation for true autonomy and adaptability, allowing robots to reason about and respond to the world in a more human-like manner~\cite{fu2024humanplus}.

As intelligent robots are increasingly deployed in real-world scenarios, such as elderly care~\cite{broekens2009assistive}, medical assistance~\cite{dupont2021decade}, disaster rescue~\cite{murphy2016disaster} and education~\cite{belpaeme2018social}. Their ability to operate autonomously and safely in dynamic, uncertain environments becomes paramount. However, the diversity of applications and the rapid pace of technological progress have created a pressing need for a systematic framework to evaluate and compare robot capabilities. Establishing a scientifically sound grading system for robot intelligence not only clarifies the technological development roadmap, but also provides essential guidance for regulation, safety assessment, and ethical deployment.

To address this need, recent research has explored various frameworks for quantifying robot capabilities, such as the DARPA Robotics Challenge evaluation scheme\cite{darpa2015}, the ISO 13482 standard for service robot safety\cite{iso13482}, and reviews on autonomy levels\cite{huang2005framework, beer2014toward}. Nevertheless, a comprehensive grading system that integrates the dimensions of intelligent cognition, autonomous behavior, and social interaction remains lacking.

In this work, we propose a capability grading model for intelligent robots, systematically outlining five progressive levels (IR-L0 to IR-L4) from basic mechanical execution to advanced, fully autonomous social intelligence. This classification encompasses key dimensions such as autonomy, task handling ability, environmental adaptability, and societal cognition, providing a unified framework to assess and guide the development of intelligent robots across the full spectrum of technological evolution.

Central to enabling intelligent behavior in robots are two key technologies: physical simulators and world models. Both play vital roles in improving the robot’s control capabilities and extending its potential. Simulators, such as Gazebo\cite{1389727} or MuJoCo\cite{todorov2012mujoco}, provide an explicit modeling of the physical world, offering a controlled environment where robots can train, test, and refine their behaviors before being deployed in real-world scenarios. These simulators serve as training grounds where the robot’s actions can be predicted, tested, and fine-tuned without the high costs and risks of real-world experimentation.

Unlike simulators, world models offer internal representations of the environment, enabling robots to autonomously simulate, predict, and plan actions within their cognitive framework. Following NVIDIA's definition, world models are "generative AI models that understand the dynamics of the real world, including physics and spatial properties"\cite{nvidiaWorldModels}. This concept gained significant attention with Ha and Schmidhuber's seminal work\cite{Ha2018}, which demonstrated how agents could learn compact environmental representations for internal planning. 

The synergy between simulators and world models enhances robots' autonomy, adaptability, and task performance across diverse scenarios. This paper will explore the interplay between robot control algorithms, simulators, and world models. By examining how simulators provide a structured, external environment for training and how world models create internal representations for more adaptive decision-making, we aim to provide a comprehensive understanding of how these components work together to enhance the capabilities of intelligent robots.

%The concept of world models in AI can be traced back to early work in model-based reinforcement learning, where agents were given a learned model of their environment to simulate potential actions. A major breakthrough occurred in 2018 with the development of the “World Models” framework by David Ha and Jürgen Schmidhuber~\cite{Ha2018}. By integrating neural networks with probabilistic models, they demonstrated how an agent could learn a compact representation of the environment and plan actions entirely within the model’s internal framework, without relying on direct interaction with the world.

% While simulators focus on explicit, external modeling of the environment, world models represent a more abstract, internalized approach to understanding the world. This distinction highlights a critical aspect of robot control algorithms, which rely on both real-time sensory input and predictive models to execute tasks. Modern robot control systems leverage simulators for initial training and validation and world models for efficient decision-making in complex, dynamic environments. The synergy between these two technologies enhances the robot’s autonomy, adaptability, and ability to perform tasks in environments that may be too unpredictable or difficult to simulate entirely.

\subsection{Scope and Contributions}
{\bf Scope.} This survey offers a comprehensive analysis of the interplay among robot control algorithms, simulators, and world models, mainly focusing on developments from 2018 to 2025. Our coverage includes both traditional physics-based simulators and emerging world models, emphasizing their impact on autonomous driving and robots.

This survey distinguishes itself from existing literature by providing a comprehensive examination of the synergistic relationship between physical simulators and world models in advancing embodied intelligence. While previous surveys have typically focused on individual components (such as robotics simulators\cite{duan2022survey, liu2024aligning, collins2021review} and world models\cite{zhu2024sora, ding2024understanding, mai2024efficient}) our work bridges these domains to reveal their complementary roles in intelligent robot development.

\noindent
\textbf{Contribution.} The main contributions of this survey are:

\begin{itemize}
    \item \textbf{Levels of intelligent robots:} Propose a comprehensive five-level grading standard (IR-L0 to IR-L4) for humanoid robot autonomy evaluation across four key dimensions: autonomy, task handling ability, environmental adaptability, and societal cognition ability.
    \item \textbf{Analysis of recent techniques of robot learning:} Systematically review recent developments in intelligent robotics across legged locomotion (bipedal walking, fall recovery), manipulation (dexterous control, bimanual coordination), and human-robot interaction (cognitive collaboration, social embeddedness).
    \item \textbf{Analysis of current physical simulators:} Provide comprehensive comparative analysis of mainstream simulators (Webots, Gazebo, MuJoCo, Isaac Gym/Sim/Lab etc.) covering physical simulation capabilities, rendering quality, and sensor support.
    \item \textbf{Recent advancements of World Models:} We first revisit the main architectures of world models and their potential roles, for example, serve as controllable simulators, dynamic models, and reward models for embodied intelligence.
    % including recurrent state-space models, autoregressive models, diffusion-based models.
    Furthermore, we comprehensively discuss the recent world models designed for specific applications such as autonomous driving and articulated robots.
    % to video diffusion-based approaches, examining applications as controllable simulators, dynamic models, and reward models for embodied intelligence.
    % \item \textbf{Domain-Specific Applications:} Analyze world model implementations in intelligent robot systems including autonomous driving and articulated robots.
\end{itemize}

\subsection{Structure}

A summary of the paper structure is illustrated in Fig.~\ref{fig:paper_structure}, which is presented as follows: 
\setlist[itemize]{itemsep=0pt, parsep=0pt, topsep=0pt, partopsep=0pt}
\begin{itemize}
    \item Sec.~\ref{sec:introduction} introduces the importance of embodied AI and outlines how physical simulators and world models contribute to the development of embodied AI.
    \item  Sec.~\ref{sec:levels_robots} provides a comprehending grading system for intelligent robots in five levels.    \begin{itemize}
        \item Sec.~\ref{level_criteria} Level Criteria.
        \item Sec.~\ref{level_factors} Level Factors.
        \item Sec.~\ref{classification_levels} Classification Levels.
    \end{itemize}
    \item  Sec.~\ref{sec:robots_mdi} reviews current progresses in intelligent robotic tasks in terms of mobility, dexterity and human-robot interaction. 
    \begin{itemize}
        \item Sec.~\ref{subsec:robot_techniques} Related Robotic Techniques.
        \item Sec.~\ref{subsec:robotic_loco} Robotic Locomotion.
        \item Sec.~\ref{subsec:robotic_manipulation} Robotic Manipulation.
        \item Sec.~\ref{subsec:robotic_hri} Human-Robot Interaction.
    \end{itemize}
    \item Sec.~\ref{sec:simulator} discusses pros and cons of mainstream simulators in contemporary robotics research. 
    \begin{itemize}
        \item Sec.~\ref{subset:mainstream_simulators} Main stream Simulators.
        \item Sec.~\ref{subsec:simulator_properties} Physical Properties of Simulators.
        \item Sec.~\ref{sec:rendering_capabilities} Rendering Capabilities.
        \item Sec.~\ref{sec:sensor_and_joint} Sensor and Joint Component Types.
        \item Sec.~\ref{subsec:discussion} Discussions and Future Perspectives.
    \end{itemize}
    \item Sec.~\ref{sec:world_models} introduces the representative architectures and core roles of world models. 
    \begin{itemize}
        \item Sec.~\ref{subsec:architecure_of_WM} Representative Architectures of World Models.
        \item Sec.~\ref{sec:wm_role} Core Roles of World Models.
    \end{itemize}
    \item Sec.~\ref{sec:world_models_for_AD_robot} further reviews applications and challenges of world models for intelligent agents including autonomous driving and articulated robots.  
    \begin{itemize}
        \item Sec.~\ref{subsec:WM_for_AD} World Models for Autonomous Driving.
        \item Sec.~\ref{subsec:WM_for_AR} World Models for Articulated Robots.
        \item Sec.~\ref{subsec:challeng_WM} Challenges and Future Perspectives.
    \end{itemize}
\end{itemize}

% \section{人形机器人硬件设计}
% \xiao{需要对机器人硬件设计有一些介绍，同时介绍一下当下中外主流的机器人种类，比如波士顿、unitree的机器人}
\section{Levels of Intelligent Robot}
\label{sec:levels_robots}
With the rapid development in fields such as artificial intelligence, mechanical engineering, sensor fusion, and human-computer interaction, intelligent robots are gradually transitioning from laboratories to real-world scenarios, including elderly care~\cite{broekens2009assistive}, medical assistance~\cite{dupont2021decade}, disaster rescue~\cite{murphy2016disaster} and education~\cite{belpaeme2018social}. Unlike traditional industrial robots, intelligent robots emphasize the completion of complex cognition, perception, and execution tasks based on human-like structures. The dynamic nature and uncertainty they face in real environments make capability assessment a critical issue. Therefore, establishing a scientifically sound capability grading system not only helps clarify the technological development roadmap but also provides guidance for robot regulation and safety assessment.

Currently, several studies have attempted to quantify robot capabilities, such as the DARPA Robotics Challenge framework for evaluating the ability to perform complex tasks \cite{darpa2015}, the ISO 13482 standard for service robot safety grading \cite{iso13482}, and reviews on robot levels of autonomy \cite{huang2005framework, beer2014toward}. However, a comprehensive grading system that combines the dimensions of "intelligent cognition" and "autonomous behavior" is still lacking. To address this, this paper proposes and systematically outlines a capability grading model for intelligent robots from IR-L0 to IR-L4, covering the full technological evolution path from mechanical operation levels to advanced social interaction levels.
\begin{figure*}[t]
    \centering
    \includegraphics[width=\linewidth]{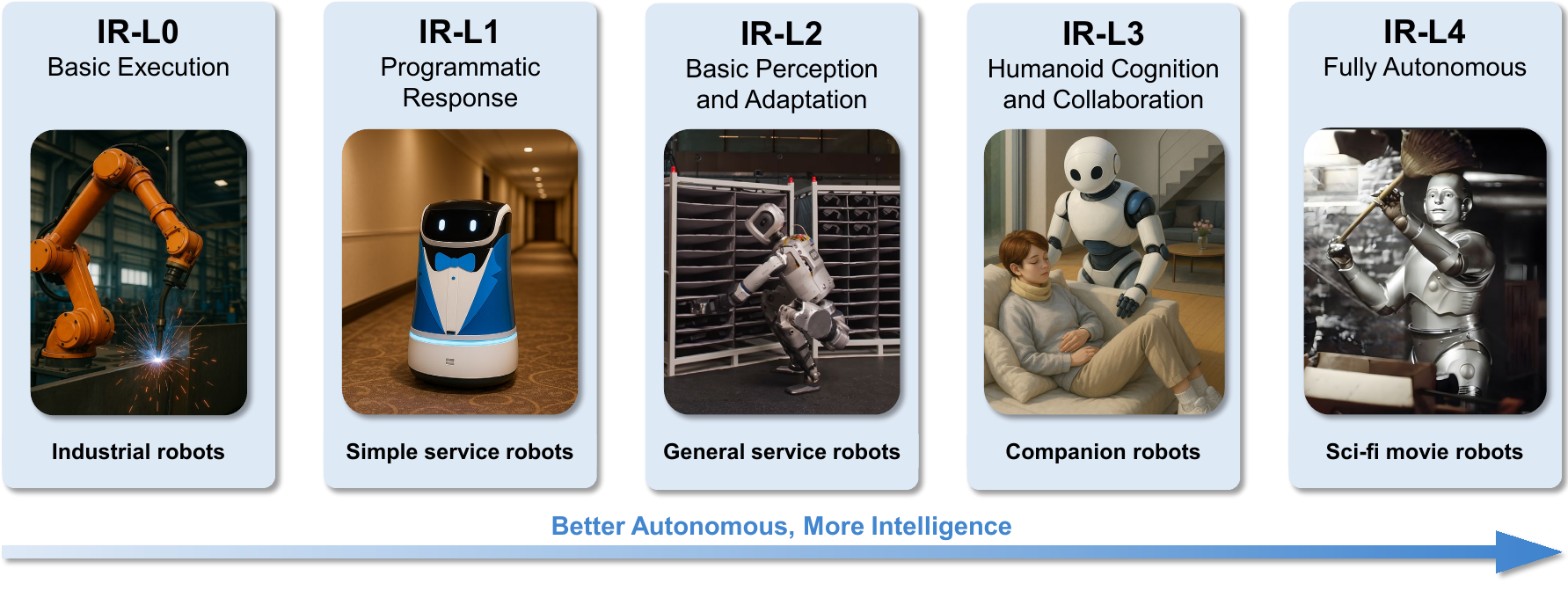}
    \caption{Levels of Intelligent Robotics: From Basic Execution to Full Autonomy. }
    \label{fig: levels}
\end{figure*}
\subsection{Level Criteria}
~\label{level_criteria}

\begin{table*}[t] 
\centering 
\renewcommand{\arraystretch}{1.8}  % 调整行间距，2.5大约相当于两行高
\begin{tabular}{>{\centering\arraybackslash}m{0.06\textwidth}|>{\centering\arraybackslash}m{0.16\textwidth}|>{\centering\arraybackslash}m{0.18\textwidth}|>{\centering\arraybackslash}m{0.24\textwidth}|>{\centering\arraybackslash}m{0.22\textwidth}} 
\hline 
 \cellcolor{gray!20}\textbf{Level} &   
 \cellcolor{gray!20}\textbf{Autonomy} &   
 \cellcolor{gray!20}\textbf{Task Handling Ability} &   
 \cellcolor{gray!20}\textbf{Environmental Adaptability} &   
 \cellcolor{gray!20}\textbf{Societal Cognition Ability} \\ 
\hline 
\textbf{IR-L0} &   Human Control    &   Basic Tasks    &    Controlled Only    &    No Social Cognition    \\ 
\hline 
 \cellcolor{gray!10}\textbf{IR-L1} &   \cellcolor{gray!10} Human Supervised    &    \cellcolor{gray!10}Complex Navigation    &     \cellcolor{gray!10}Predictable Environments    &     \cellcolor{gray!10}Basic Recognition    \\ 
\hline 
\textbf{IR-L2} &   Human Assisted    &   Dynamic Collaboration    &    Adaptive Learning    &    Simple Interaction    \\ 
\hline 
 \cellcolor{gray!10}\textbf{IR-L3} &   \cellcolor{gray!10} Conditional Autonomy    &   \cellcolor{gray!10} Multitasking    &     \cellcolor{gray!10}Dynamic Adaptation    &    \cellcolor{gray!10} Emotional Intelligence    \\ 
\hline 
\textbf{IR-L4} &   Full Autonomy    &   Innovation    &    Universal Flexibility    &    Advanced Social Intelligence    \\ 
\hline 
\end{tabular} 
\renewcommand{\arraystretch}{1}  % 恢复默认行间距
\caption{Relationship between graded levels and level factors.} 
\label{tab: Relationship between classified levels and classification factors} 
\end{table*}

This standard classifies robots based on: their ability to perform tasks in various environments, the depth of autonomous decision-making, interaction complexity, and ethical cognition. The following core dimensions are covered:

\begin{itemize}
    \item The robot's ability to complete tasks independently, ranging from full reliance on human control to full autonomy.
    \item The difficulty of tasks the robot can handle, from simple repetitive labor to innovative problem-solving.
    \item The robot’s ability to work in dynamic or extreme environments.
    \item The robot's capacity to understand, interact with, and respond to social situations within human society.
\end{itemize}

\subsection{Level Factors}
~\label{level_factors}
The intelligent level of robots is graded based on the following five factors.
\begin{itemize}
    \item \textbf{Autonomy}: This factor is based on the robot's ability to autonomously make decisions across various tasks. 
    \item \textbf{Task Handling Ability}: This factor is based on the complexity of the tasks the robot can perform.
    \item \textbf{Environmental Adaptability}: This factor is based on the robot’s performance in different environments.
    \item \textbf{Societal Cognition Ability}: This factor is based on the level of intelligence exhibited by robots in social scenarios.
\end{itemize}

The relationship between graded levels and level factors are listed in table \ref{tab: Relationship between classified levels and classification factors}.

% \usepackage{multirow}
% \usepackage{multirow}

\subsection{Classification Levels}
~\label{classification_levels}
\subsubsection{\textbf{IR-L0:} Basic Execution Level}

IR-L0 represents the foundational execution level in this system, characterized by a completely non-intelligent, program-driven attribute. Robots at this level are focused on executing highly repetitive, mechanized, deterministic tasks, such as industrial welding and fixed-path material handling. The "low perception - high execution" operational mode makes the robot entirely relies on predefined programmatic instructions or real-time teleoperation. It lacks environmental perception, state feedback, or autonomous decision-making capability, forming a unidirectional closed-loop system with "command input - mechanical execution" \cite{siciliano2010robotics}.
The potential technical requirements are summarized as follows:
\begin{itemize} 
    \item \textbf{Hardware:} High-precision servomotors and rigid mechanical structures, with motion controllers based on PLC or MCU. 
    \item \textbf{Perception:} Extremely limited, typically involving only limit switches, encoders, etc. 
    \item \textbf{Control Algorithms:} Mainly based on predefined scripts, action sequences, or teleoperation, without real-time feedback loops. 
    \item \textbf{Human-Robot Interaction:} None, or limited to simple buttons/teleoperation.
\end{itemize}

\subsubsection{\textbf{IR-L1:} Programmatic Response Level}

IR-L1 robots feature limited rule-based reactive capabilities, enabling them to execute predefined task sequences such as those performed by cleaning and reception robots. These systems utilize fundamental sensors, including infrared, ultrasonic, and pressure sensors, to trigger specific behavioral patterns.
They cannot process complex or unforeseen events and can  demonstrate operational stability only in closed-task environments with clearly defined rules. They embody a "limited perception–limited execution" paradigm, representing the beginning of basic robotic intelligence \cite{murphy2000}.
The potential technical requirements are summarized as follows:
\begin{itemize}
    \item \textbf{Hardware:} Integration of basic sensors (infrared, ultrasonic, pressure) with moderately enhanced processor capabilities.
    \item \textbf{Perception:} Detection of obstacles, boundaries, and simple human movements.
    \item \textbf{Control Algorithms:} Rule engines and finite state machines (FSM), supplemented by basic SLAM or random walk algorithms.
    \item \textbf{Human-Robot Interaction:} Basic voice and touch interfaces supporting simple command-response protocols.
    \item \textbf{Software Architecture: }Embedded real-time operating systems with elementary task scheduling capabilities.
\end{itemize}

\subsubsection{\textbf{IR-L2:} Basic Perception and Adaptation Level}

IR-L2 robots introduce preliminary environmental awareness and autonomous capabilities, representing a significant advancement in robotic intelligence. Their characteristics include fundamental responsiveness to environmental changes and the ability to transition between multiple task modes. For instance, a service robot at this level can execute distinct tasks such as "water delivery" or "navigation guidance" based on voice commands while simultaneously avoiding obstacles during path execution. These systems require integrated perception modules (cameras, microphone arrays, LiDAR) and basic behavior-decision frameworks, such as Finite State Machines (FSM) or Behavior Trees \cite{colledanchise2016behavior}.

While human supervision remains essential, IR-L2 robots demonstrate substantially greater execution flexibility compared to IR-L1 systems, signifying their progression toward genuine "contextual understanding."
The potential technical requirements are summarized as follows:
\begin{itemize}
    \item \textbf{Hardware:} Multimodal sensor arrays (cameras, LiDAR, microphone arrays) coupled with enhanced computational resources.
    \item \textbf{Perception:}  Visual processing, auditory recognition, and spatial localization capabilities enabling basic object identification and environmental mapping.
    \item \textbf{Control Algorithms:} Finite state machines, behavior trees, SLAM implementations, path planning, and obstacle avoidance systems.
    \item \textbf{Human-Robot Interaction:} Speech recognition and synthesis capabilities supporting comprehension and execution of basic commands.
    \item \textbf{Software Architecture:} Modular design framework facilitating parallel task execution with preliminary priority management systems.
\end{itemize}

\subsubsection{\textbf{IR-L3: } Humanoid Cognition and Collaboration Level}

IR-L3 robots demonstrate autonomous decision-making capabilities in complex, dynamic environments while supporting sophisticated multimodal human-robot interaction. These systems can infer user intent, adapt their behavior accordingly, and operate within established ethical constraints. For example, in eldercare application, IR-L3 robots analyze speech patterns and facial expressions to detect emotional state changes in elderly patients, responding with appropriate comforting actions or emergency alerts.
The potential technical requirements are summarized as follows:
\begin{itemize}
    \item \textbf{Hardware:} High-performance computing platforms integrated with comprehensive multimodal sensor suites (depth cameras, electromyography sensors, force-sensing arrays).
    \item \textbf{Perception:} Multimodal fusion of vision, speech, and tactile inputs; affective computing for emotion recognition and dynamic user modeling.
    \item \textbf{Control Algorithms:} Deep learning architectures (CNNs, Transformers) for perception and language understanding; reinforcement learning for adaptive policy optimization; planning and reasoning modules for complex task workflow management.
    \item \textbf{Human-Robot Interaction:} Multi-turn natural language dialogue support; facial expression recognition and feedback; foundational empathy and emotion regulation capabilities.
    \item \textbf{Software Architecture:} Service-oriented, distributed frameworks enabling task decomposition and collaborative execution; integrated learning and adaptation mechanisms.
    \item \textbf{Safety and Ethics:} Embedded ethical governance systems to prevent unsafe or non-compliant behaviors.
\end{itemize}

\subsubsection{\textbf{IR-L4:} Fully Autonomous Level
}
IR-L4 represents the pinnacle of intelligent robotics: systems with complete autonomy in perception, decision-making, and execution, capable of independent operation in any environment without human intervention. These robots possess self-evolving ethical reasoning, advanced cognition, empathy, and long-term adaptive learning capabilities. Beyond handling open-ended tasks, they engage in sophisticated social interactions, including multi-turn natural language dialogue, emotional understanding, cultural adaptation, and multi-agent collaboration.
The potential technical requirements are summarized as follows:
\begin{itemize}
    \item \textbf{Hardware:} Highly biomimetic structures featuring full-body, multi-degree-of-freedom articulation; distributed high-performance computing platforms.
    \item \textbf{Perception:} Omnidirectional, multi-scale, multimodal sensing systems; real-time environment modeling and intent inference.
    \item \textbf{Control Algorithms:} General Artificial Intelligence (AGI) frameworks integrating meta-learning, generative AI, and embodied intelligence; autonomous task generation and advanced reasoning capabilities.
    \item \textbf{Human-Robot Interaction: }Natural language understanding and generation; complex social context adaptation; empathy and ethical deliberation.
    \item \textbf{Software Architecture:} Cloud-edge-client collaborative systems; distributed agent architectures supporting self-evolution and knowledge transfer.
    \item \textbf{Safety and Ethics: }Embedded dynamic ethical decision systems constraining behavior and enabling morally sound choices in ethical dilemmas.
\end{itemize}

\section{Robotic Mobility, Dexterity and Interaction}
\label{sec:robots_mdi}
Among the various embodiments of intelligent robots, humanoid robots—characterized by their human-like appearance—stand out for their ability to seamlessly integrate into human-centered environments and provide meaningful assistance. As such, they serve as crucial physical representations of embodied intelligence. 

Recently, the rapid advancement of machine learning technologies has led to significant breakthroughs in robot whole-body control and general manipulation. This chapter begin by outlining fundamental technical approaches in intelligent robotics, then review the latest developments in robot locomotion and manipulation, and finally, examine ongoing research aimed at enabling natural and intuitive human-robot interactions.

\subsection{Related Robotic Techniques}
\label{subsec:robot_techniques}

\subsubsection{Model Predictive Control, MPC}
Model Predictive Control (MPC) \cite{rawlings2017model} is a powerful control strategy that has gained significant traction in the field of humanoid robotics over the past two decades. At its core, MPC is an optimization-based approach that predicts the future behavior of a system using a dynamic model, and computes control actions by solving an optimization problem at each time step. This allows the controller to explicitly handle constraints on inputs and states, making it particularly suitable for complex, high-dimensional systems like humanoid robots \cite{katayama2023model}.

Tom Erez et al. introduced a comprehensive real-time MPC system that applied MPC to the full dynamics of a humanoid robot, enabling it to perform complex tasks such as standing, walking, and recovering from disturbances\cite{erez2013integrated}. Building upon this foundation, in 2015, Jonas Koenemann, Andrea Del Prete, Yuval Tassa, Emanuel Todorov et al. implemented a complete MPC and applied it in real-time on the physical HRP-2 robot, marking the first time such a whole-body model predictive controller was applied in real-time on a complex dynamic robot\cite{koenemann2015whole}.

\subsubsection{Whole-Body Control, WBC}
Whole body control (WBC) in humanoid robotics is a comprehensive framework that enables robots to coordinate all of their joints and limbs simultaneously to achieve different motions. The basic methods of whole body control typically involve formulating the robot’s motion and force objectives as a set of prioritized tasks, such as maintaining balance, following a desired trajectory, or applying a specific force with the hand. These tasks are then translated into mathematical constraints and objectives, which are solved using optimization techniques or hierarchical control frameworks\cite{goswami2019humanoid}. 

During implementation, WBC typically employs techniques such as dynamic modeling, inverse kinematics solving, and optimization algorithms to ensure that the robot achieves desired motion behaviors while satisfying physical constraints. In the early 2000s, Oussama Khatib and his collaborators introduced the operational space formulation for controlling redundant manipulators and later extended it to humanoid robots\cite{sentis2006whole}. Optimization-based WBC offers strong flexibility, allowing for modular addition or removal of constraints and resolving conflicting constraints by setting different task hierarchies or soft task weightings \cite{escande2014hierarchical, kuindersma2016optimization, hopkins2015compliant}. In recent years, with the development of artificial intelligence such as reinforcement learning, researchers have proposed frameworks like ExBody2 \cite{ji2024exbody2} and HugWBC\cite{xue2025unified}, which train control policies in simulated environments and transfer them to actual robots, achieving more natural and expressive whole-body motion control.

\subsubsection{Reinforcement Learning}
Reinforcement learning (RL)\cite{kober2013reinforcement} is a branch of machine learning that has become increasingly influential in the field of humanoid robotics. The core insight of RL is that an agent—such as a humanoid robot—can learn to perform complex tasks by interacting with its environment and receiving feedback in the form of rewards or penalties. Unlike traditional control methods\cite{nguyen2011model,yamaguchi1999development} that require explicit programming or modeling of behaviors, RL enables robots to autonomously discover optimal actions through trial and error, making it particularly well-suited for high-dimensional, dynamic, and uncertain environments that humanoid robots often encounter\cite{ha2024learning}.

The application of reinforcement learning to humanoid robotics dates back to the late 1990s and early 2000s. In 1998, Masahiro Morimoto and Kenji Doya introduced a reinforcement learning (RL) approach enabling a simulated two-joint, three-link robot to autonomously learn a dynamic standing-up motion from a lying position\cite{morimoto1998reinforcement}. Since then, RL has been used to achieve complex behaviors for humanoid robots, DeepLoco\cite{peng2017deeploco} and other works\cite{yu2018learning, heess2017emergence} did an extensive exploration of the ability of deep RL in bipedal tasks, but they have not shown to be be suitable for physical robots. In 2019, Xie et al. used iterative reinforcement learning and Deterministic Action Stochastic State (DASS) tuples to progressively refine the reward function and policy architecture, enabling robust dynamic walking for a physical Cassie bipedal robot\cite{xie2019iterative}.

\subsubsection{Imitation Learning}
Imitation learning (IL) is a paradigm in robotics where robots learn to perform tasks by observing and mimicking demonstrations, typically provided by humans or other agents. The core insight behind imitation learning is that it bypasses the need for explicit programming or hand-crafted reward functions, allowing humanoid robots to acquire complex behaviors more efficiently and intuitively. By leveraging demonstrations, robots can learn skills such as walking, manipulation, or social interaction, which are otherwise difficult to specify through traditional control or reinforcement learning methods\cite{chi2023diffusion,ze20243d,zhao2023learning,liu2024rdt}.

In humanoid robot motion control, IL often utilizes human motion capture data that has been re-targeted (Retargeting), or reference gaits generated from model-based trajectory planning (such as natural walking, running, etc.), and encourages the robot to follow these reference trajectories in simulation to achieve more natural and stable motion gaits \cite{huang2024diffuseloco, peng2021amp, zhang2024whole}.

While IL effectively utilizes existing knowledge for learning, it faces challenges such as high costs for obtaining expert demonstration data, insufficient data diversity, quality concerns, and time-consuming processes. Moreover, strategies trained on limited demonstration data often suffer from poor generalization, making them difficult to adapt to new environments or tasks. The operational skills learned may also be relatively narrow. To address these challenges, researchers and companies are focusing on developing more efficient data collection hardware platforms or teleoperation technologies to expand data \cite{zhao2023learning, wang2024dexcap,cheng2024open}, while also exploring new types of training data, such as extracting human actions from video data\cite{allshire2025visual}.

\subsubsection{Visual-Language-Action Models, VLA}

Visual-Language-Action (VLA) models represent a cross-modal artificial intelligence framework that integrates visual perception, language understanding, and action generation. The core concept involves leveraging the reasoning capabilities of Large Language Models (LLMs) to directly map natural language instructions to actions of physical robotic.

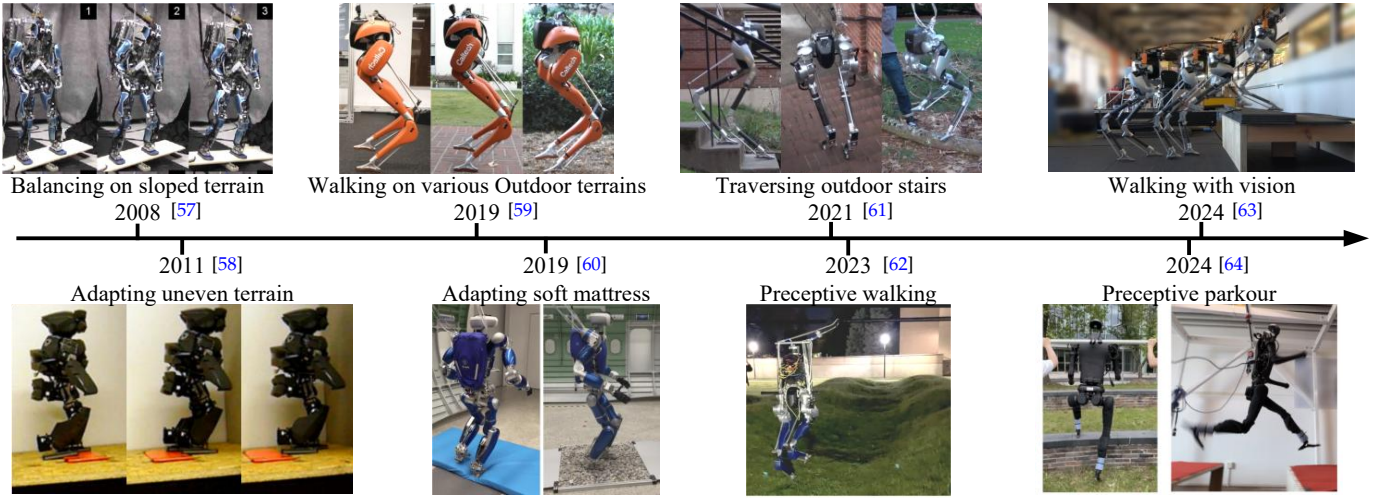
\begin{figure*}[ht]
	\centering
   \begin{overpic}[width=\linewidth]{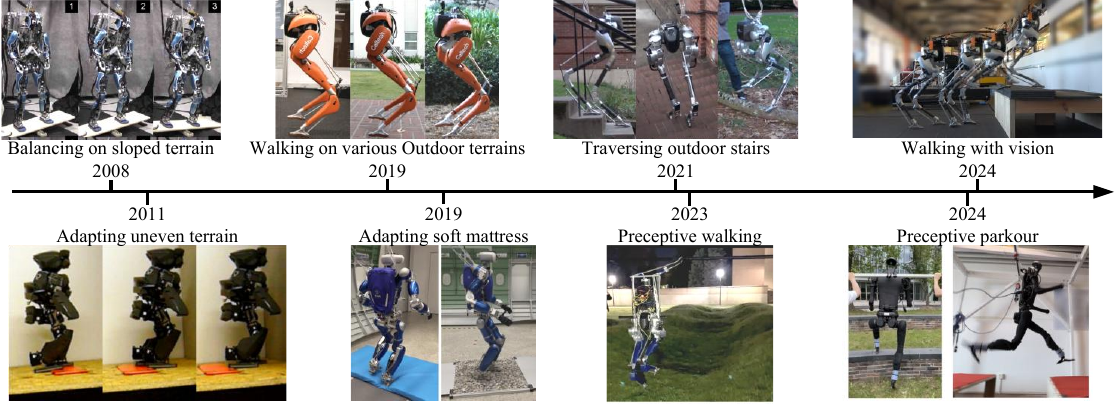} % Include your image here
    \put(12,20.7){\scriptsize \cite{hyon2008compliant}}
    \put(15, 16.8){{\scriptsize \cite{yi2011practical}}}
    \put(36.6, 20.7){\scriptsize \cite{reher2019dynamic}}
    \put(41.6, 16.8){\scriptsize \cite{mesesan2019dynamic}}
    \put(62.5, 20.7){\scriptsize \cite{siekmann2021blind}}
    \put(64, 16.8){\scriptsize \cite{huang2023efficient}}
    \put(89.5, 20.7){\scriptsize \cite{duan2024learning}}
    \put(88.5, 16.8){\scriptsize \cite{zhuang2024humanoid}}
  \end{overpic}
     \caption{Timeline of advancements in unstructured environment adaption of humanoid robot.}
\end{figure*}

In 2023, Google DeepMind introduced RT-2, which first applied this paradigm to robot control by discretizing robot control instructions into language-like tokens, achieving end-to-end visual-language-action mapping \cite{zitkovich2023rt}. By leveraging internet-scale visual-language data for pre-training, robots could understand previously unseen semantic concepts and generate reasonable action sequences through chain-of-thought reasoning. Subsequently, numerous end-to-end VLA models emerged \cite{kim2024openvla}\cite{zhen20243d}\cite{yang2025magma}\cite{black2410pi0}\cite{pertsch2025fast}\cite{shi2025hi}\cite{wen2024tinyvla}, further advancing the application and development of VLA models in robotics.

Although current VLA models have made significant progress, several critical challenges still remain. These models often struggle to reliably handle tasks or environments they have not encountered before. Additionally, real-time inference constraints limit their responsiveness in dynamic situations. Furthermore, biases within training datasets, difficulties in semantic grounding across modalities, and the high computational complexity of system integration continue to hinder further development\cite{sapkota2025vision}.

\subsection{Robotic Locomotion}
\label{subsec:robotic_loco}
The goal of robotic locomotion is to achieve natural movement patterns, including walking, running, and jumping. By integrating multiple domains such as perception, planning, and control, robots with locomotion abilities could be categorized within our IR-L2 level. This integration enables robots to adapt dynamically to varying terrain, external disturbances, and unforeseen events, promoting robust and fluent bipedal motion. Moreover, the ability to autonomously recover from unexpected incidents reduces dependence on human intervention, paving the way toward greater intelligence and autonomy. This section will explore recent advancements in legged locomotion and discuss strategies for protecting and recovering from falls.

\subsubsection{Legged Locomotion}

Bipedal robots exhibit unique advantages in navigating complex terrains, emulating human behaviors, and seamlessly integrating into human-centric environments. Research in the field of bipedal locomotion control can be divided into two tasks: \textbf{Unstructured Environment Adaption}, which emphasizes the ability to maintain stable walking in complex, unknown or dynamic environments, and \textbf{High Dynamic Movements}, which focuses on achieving a balance between stability and agility in high-speed, dynamic movements such as running and jumping.

\noindent
\textbf{Unstructured Environment Adaption. } “Unstructured environments” typically refer to complex natural or man-made terrains, such as rugged mountain trails, gravel-strewn ground, slippery grass, staircases, and other unpredictable obstacles. 
Early efforts in bipedal walking stabilization mainly used position-controlled humanoid robots. In 2008, Sang-Ho Hyon\cite{hyon2008compliant} introduced a passivity-based contact force control framework, enabling the SARCOS humanoid robot\cite{safonova2003optimizing} to actively balance on indoor terrains with varying heights and time-varying inclinations. 
Subsequent advancements explored diverse strategies to improve locomotion stability, including online learning for terrain adaptation\cite{yi2010online}, compliance control with terrain estimation\cite{yi2011practical}, and the integration of Linear Inverted Pendulum Model (LIPM) with foot force control\cite{kajita2010biped}, as demonstrated by Kajita et al. in 2010 on the HRP-4C humanoid robot.

Above mentioned methods only enabled humanoid robots with limited terrain adaptability, such as walking on thin beams, pavement, or flat but inclined ramp. This is due to the high gear-ratio of position controlled robot joints. These joints suffers from high impedance, which could easily cause damage under large impact applied to the output shaft or end-effector (hand and foot)\cite{stephens2010dynamic}. 
To achieve better adaptability in unknown environments, modern humanoid and quadruped robots use force-controlled joints with low gear ratios, offering improved compliance and smoother responses under large impacts\cite{sombolestan2021adaptive, henze2016passivity}. 

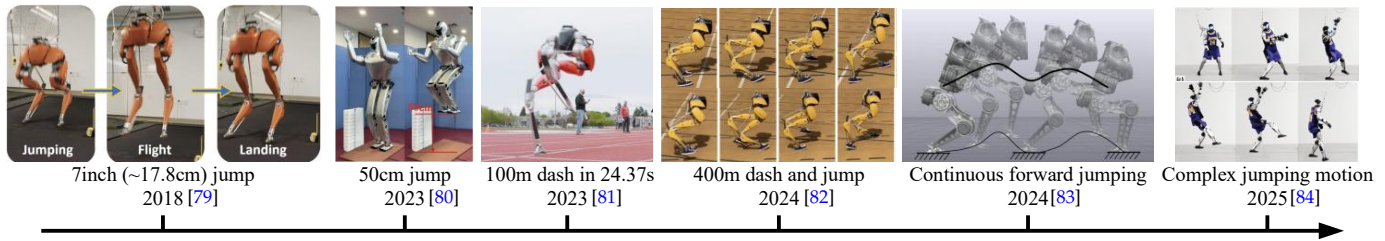
\begin{figure*}[t]
	\centering
    \begin{overpic}[width=\linewidth]{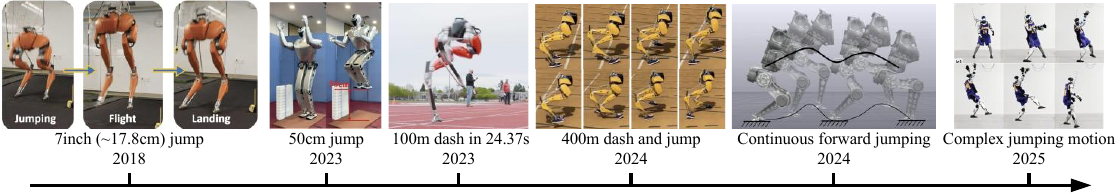} % Include your image here
    \put(13, 2.8){\scriptsize \cite{xiong2018bipedal}}
    \put(30.6, 2.8){{\scriptsize \cite{qihaoxiang_jump}}}
    \put(42.5, 2.8){\scriptsize \cite{crowley2023optimizing}}
    \put(58, 2.8){\scriptsize \cite{li2024reinforcement}}
    \put(76, 2.8){\scriptsize \cite{he2024cdm}}
    \put(93.5, 2.8){\scriptsize \cite{he2025asap}}
  \end{overpic}
 \caption{Timeline of advancements in high dynamic movements of humanoid robot. }
\end{figure*}

With development of force-controlled humanoid robots and increased computational power, researchers are able to develop and implement more sophisticated control algorithms\cite{le2024fast}. This further enhanced robots' adaptability to diverse environments. Jacob Reher et al. introduced a comprehensive full-body dynamic controller that explicitly accounts for the passive spring mechanisms in the Cassie bipedal robot\cite{reher2019dynamic}. Their approach successfully achieved stable bipedal walking on various terrains, including outdoor grass. George Mesesan et al. combines Divergent Component of Motion (DCM) for center of mass trajectory planning with a passivity-based whole-body controller (WBC) to compute joint torques\cite{mesesan2019dynamic}. Demonstrated dynamic walking over soft gym mats with the TORO robot.

Apart from proprioceptive blind walk, researchers also explored integrating exteroceptive sensing and path planning modules to tackle more complex environments. Jiunn-Kai Huang et al. integrates a low-frequency path planner with a high-frequency reactive controller that generates smooth, feedback-driven motion commands\cite{huang2023efficient}. This enables the Cassie Blue robot to autonomously traverse more complex terrains, such as the University of Michigan's Wave Field. 

Learning-based methods also performed promising stability under outdoor complex environments. In 2020, Joonho Lee et al.\cite{lee2020learning} demonstrated the first successful real-world application of reinforcement learning to legged locomotion, outperforming traditional approaches in outdoor environments\cite{carpentier2021recent}. Jonah Siekmann et al. achieved blind stair traverse with the Cassie robot\cite{siekmann2021blind}. They used domain randomization methods to vary stair dimensions and robot dynamics, which enabled the learned policy to transfer successfully to real-world scenarios.
Researchers also leveraged depth cameras and LiDAR to construct height maps \cite{duan2024learning}, Perceptive Internal Models (PIM) \cite{long2024learning}, or end-to-end policy \cite{zhuang2024humanoid}, significantly enhancing robot mobility across diverse terrains. Now robots can traverse stairs, overcome hurdles, and even jump across gaps up to 0.8 meters wide.

\noindent
\textbf{High Dynamic Movements.}
Highly dynamic movements such as running and jumping impose greater demands on the control systems of bipedal robots. During rapid motion, robots must manage quick support transitions, posture adjustments, and precise force control within very short duration.

Early studies employed simplified dynamic models, such as the Spring-Loaded Inverted Pendulum (SLIP) \cite{shahbazi2016unified}, Linear Inverted Pendulum Model (LIPM) \cite{pratt2012capturability}, and Single Rigid Body Model (SRBM) \cite{shen2022convex}, to reduce computational complexity and enable real-time control. Xiaobin Xiong and Aaron D. Ames developed a simplified spring-mass model controlled via a Control Lyapunov Function-based Quadratic Program (CLF-QP) \cite{xiong2018bipedal}, successfully achieving an 18 cm vertical jump with the Cassie robot. Qi et al. \cite{qihaoxiang_jump} proposed a Center of Pressure (CoP)-guided angular momentum controller based on LIPM, enabling vertical jumps of up to 0.5 meters by stabilizing angular momentum during flight. More recently, He et al. \cite{he2024cdm} introduced a Centroidal Dynamics Model (CDM) coupled with a MPC framework, termed CDM-MPC, to achieve continuous jumping motions on the KUAVO humanoid robot.

RL-based methods have also been applied to dynamic tasks. Learned implicit robot dynamics have shown promising results in activities such as running \cite{crowley2023optimizing}, jumping \cite{li2024reinforcement}, and parkour on discrete terrain \cite{zhuanghumanoid}, significantly broadening the capabilities of bipedal locomotion.

Training highly dynamic motions from scratch often requires tedious reward function design and parameter tuning. Imitation learning, leveraging extensive human motion datasets \cite{AMASS:ICCV:2019}, has been employed to achieve expressive and dynamic robot behaviors. Adversarial Motion Priors (AMP) \cite{Escontrela_AMP} derive style-based rewards from motion capture data, enhancing the naturalness of robot movements. Frameworks such as Exbody \cite{Cheng-RSS-24, ji2024exbody2}, OmniH2O \cite{heomnih2o}, and ASAP \cite{he2025asap} realized natural and agile whole-body movements. ASAP specifically addressing simulation-to-reality gaps, enabled sophisticated movements like the Fadeaway Jump-Shot.

\subsubsection{Fall Protection and Recovery}

Humanoid robots are prone to instability and falls which could cause hardware damage or interrupt operation. Therefore, fall protection for humanoid robots and efficient recovery to standing posture after falls have become major research topics in humanoid robotics.

\noindent
\textbf{Model-based Methods. }Early model-based fall protection and recovery control for humanoid robots largely drew inspiration from biomechanics, mimicking the biomechanical features of human falling processes combined with optimization control methods to generate motion trajectories that reduce damage during falls and achieve stable standing recovery. 

UKEMI\cite{fujiwara2002ukemi} controlled the robot's posture during falls to distribute impact forces and reduce damage to critical components. They also designed specific joint movement patterns and control strategies to enable robot fall recovery \cite{fujiwara2003first}. Libo Meng et al. proposed a falling motion control method through biomechanical analysis of human falls, achieving impact protection control adaptable to different falling directions\cite{meng2015falling}. Dong et al. proposed a compliant control framework that allows the robot to adjust its stiffness and damping characteristics in response to external disturbances, similar to how humans modulate muscle stiffness to maintain balance\cite{huang2022resistant}. 
\begin{figure}[t]
	\centering
	\includegraphics[width=\linewidth]{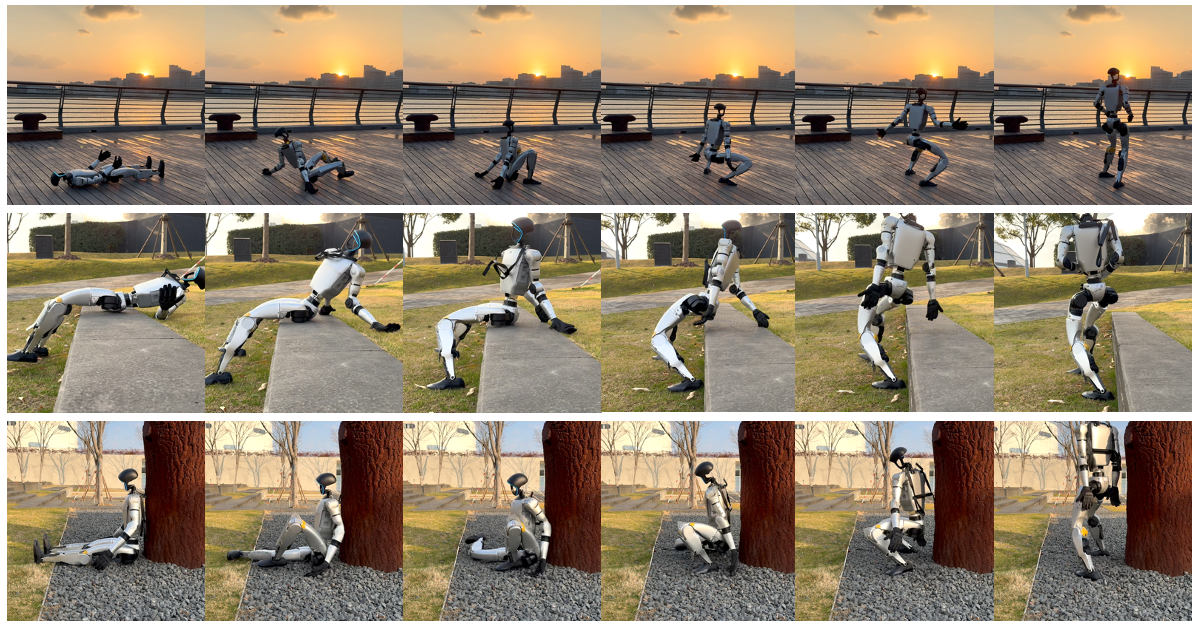}
 \caption{The HOST, proposed by Tao Huang et al.\cite{huang2025learning}, enables the Unitree G1 robot to stand-up from diverse postures in complex environments.}
\end{figure}

\noindent
\textbf{Learning-based Methods. }Learning-based methods, with their insensitivity to high-precision models and strong generalization capabilities, can better address this task. HiFAR \cite{chen2025hifar} trained humanoid robots to recover from falls through a multi-stage curriculum learning method that gradually increased scenario difficulty, achieving high success rates in fall recovery across various scenarios with real robots. HoST \cite{huang2025learning} enabled robust and natural standing actions on the Unitree G1 robot across various  environments with smoothing regularization and implicit action velocity limits. Embrace Collisions \cite{zhuang2025embrace} expanded the robot's ability to interact with environment through whole-body contacts, rather than being restricted to hands and feet alone. By imitating human actions such as roll-and-stand, side-lying, and other multi-contact behaviors, this approach extended the range of motion and adaptability of humanoid robots in real-world scenarios.

\subsection{Robotic Manipulation}
\label{subsec:robotic_manipulation}
Robotic manipulation tasks encompass a wide range of activities, from simple actions like picking up objects to complex sequences involving assembly or cooking. 
% Achieving advanced manipulation relies on the tight integration of both intelligent systems and sophisticated hardware. On the intelligent systems side, these tasks require several key components: accurate perception to identify and classify objects in the environment, autonomous decision-making and reasoning to determine optimal actions, and robust motion control for precise execution. On the hardware side, achieving advanced manipulation ability for humanoid robots requires seamless whole-body coordination mechanisms, with particular emphasis on dexterous hands for fine manipulation, dual-limb systems for effective bimanual tasks, and stable locomotion capabilities to maintain balance and mobility during complex operations. Only through the close synergy of these intelligent and physical components can humanoid robots interact safely and efficiently with dynamic and unstructured environments.

This section will review research progress in robot manipulation, focusing on the increasing complexity of coordination required for different tasks. We will begin with manipulation using a single end effector, such as a hand or gripper, progress to bimanual coordination involving two arms, and finally address whole-body manipulation tasks that require the integrated control of the entire robot.

\subsubsection{Unimanual Manipulation Task}
Unimanual manipulation refers to the use of a single end effector to interact with and manipulate objects, such as a parallel gripper or a dexterous robotic hand. Tasks in this category range from basic pick-and-place operations to more complex actions such as pushing, inserting, tool usage, and manipulating deformable or articulated objects. The complexity of these tasks depends on the capabilities of the end effector and the environment it interacts with.

% \noindent\textbf{Gripper-based manipulation} focuses on tasks performed by simple end effectors such as parallel-jaw or two-finger grippers. Typical tasks include grasping, pick-and-place, pushing, insertion, and basic tool use. Early industrial robots with grippers relied on precise models and pre-programmed motions to operate in structured environments \cite{zare2024survey, kragic2002survey, an2025dexterous}. As research progressed, learning-based approaches have enabled greater adaptability. For example, PoseCNN \cite{xiang2017posecnn} and NOCS \cite{wang2019normalized} advanced object pose estimation, while AffordanceNet \cite{do2018affordancenet} and Where2Act \cite{mo2021where2act} helped robots identify functional regions for manipulation. In cluttered environments, CollisionNet \cite{murali20206} and PerAct \cite{shridhar2023perceiver} addressed grasp planning and perception challenges, enabling more robust and flexible gripper-based manipulation.

\noindent\textbf{Gripper-based manipulation.} The parallel two-finger gripper \cite{an2025dexterous} is the most common end effector for manipulation tasks like grasping, placing, and tool use, relying on simple opening and closing actions.
Early research focused on precise physical models and pre-programming \cite{zare2024survey}, effective in structured settings like industrial automation with predetermined trajectories or Visual Servoing \cite{kragic2002survey} for feedback. These methods struggled with adaptability in unstructured environments, object diversity, and complex interactions (e.g., friction, deformation).

Learning-based approaches overcame these limitations. In perception, PoseCNN \cite{xiang2017posecnn} enabled instance-level 6D pose estimation, while NOCS \cite{wang2019normalized} advanced category-level estimation for generalization. Functional affordance learning progressed with AffordanceNet \cite{do2018affordancenet} identifying manipulable regions via supervised learning and Where2Act \cite{mo2021where2act} using self-supervised simulation interactions. Imitation learning evolved with Neural Descriptor Fields (NDFs) \cite{simeonov2022neural} enhancing policy generalization, Diffusion Policy \cite{chi2023diffusion} leveraging diffusion models for multimodal actions, and RT-2 \cite{brohan2023rt} integrating foundation models to interpret complex instructions.

Task-oriented manipulation expanded with these advances. Beyond basic grasping \cite{chi2023diffusion}, robots handle cluttered environments with collision avoidance (e.g., CollisionNet \cite{murali20206}, PerAct \cite{shridhar2023perceiver}), manipulate deformable and articulated objects \cite{arriola2020modeling, gu2023survey, she2021cable, yang2023neural}, and perform bimanual coordination \cite{zhao2023learning}, dexterous hand manipulation \cite{wang2024dexcap}, and whole-body control \cite{jiang2025behavior}.
This evolution underscores gripper-based manipulation’s growing capability for diverse, complex tasks.

% \noindent\textbf{Dexterous hand manipulation} aims to endow robots with human-like precision and versatility through multi-fingered end effectors. Early research focused on hardware innovation, such as the Utah/MIT Hand \cite{jacobsen1986design} and BarrettHand \cite{townsend2000barretthand}, which provided more degrees of freedom and adaptive grasping. With the rise of data-driven methods, robots equipped with dexterous hands have demonstrated remarkable progress in in-hand manipulation, complex grasping, and tool use. Recent works such as UGG \cite{lu2024ugg}, DexVIP \cite{mandikal2022dexvip}, and DexGraspVLA \cite{zhong2025dexgraspvla} combine advanced perception with learning-based control, enabling dexterous hands to generalize across diverse objects and tasks, and to perform challenging manipulations in unstructured environments.
\begin{figure}[t]
	\centering
	\includegraphics[width=0.9\linewidth]{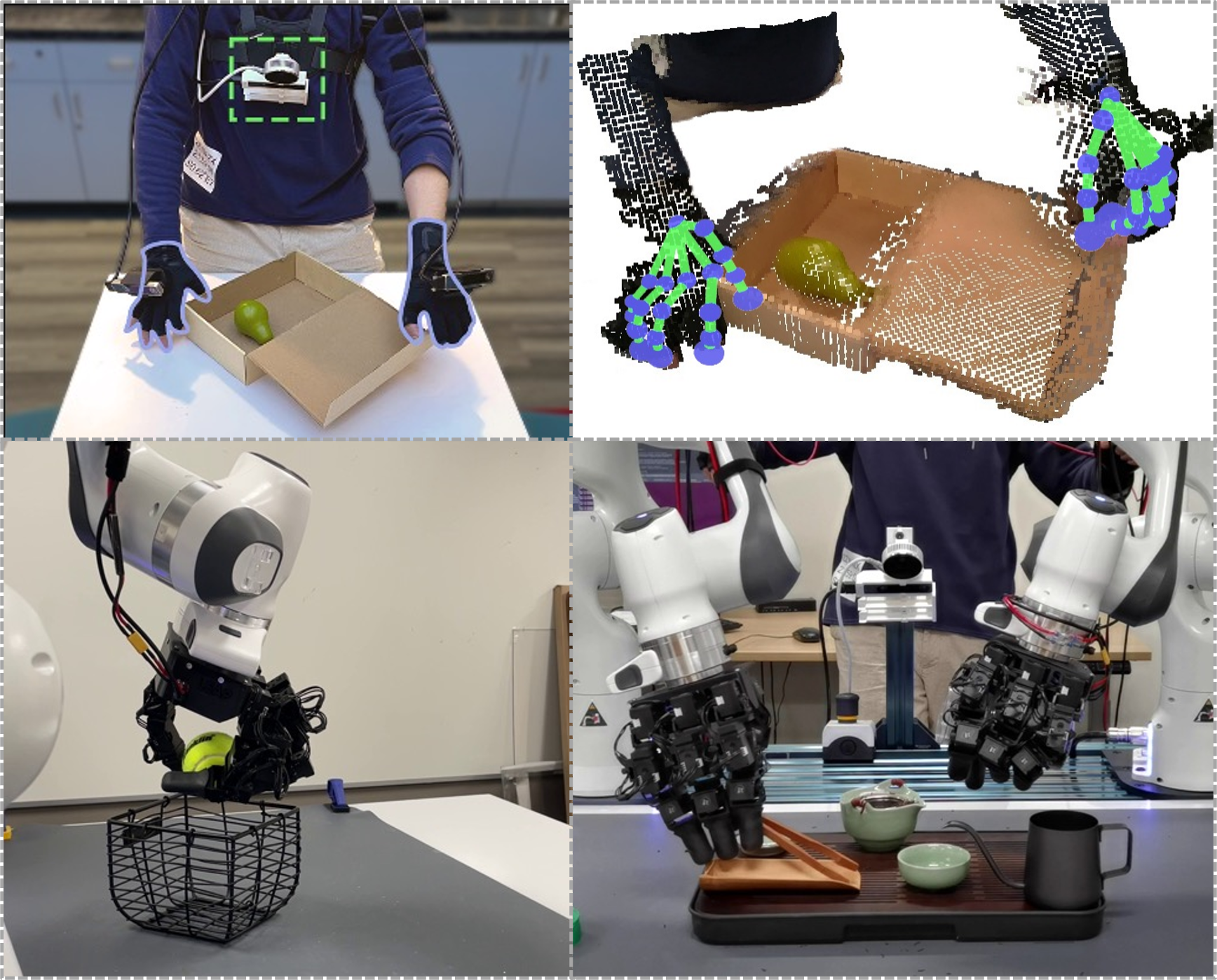}
 \caption{DexCap\cite{wang2024dexcap} is a portable motion capture system that collects human hand movements, enabling robots to complete tasks from simple picks to complex operations like Tea preparing.}
\end{figure}
\noindent\textbf{Dexterous hand manipulation.}
Dexterous manipulation aims to enable robots to interact with the physical world in complex, precise ways akin to human hands, which is a core challenge in robotics for decades~\cite{okamura2000overview}. This field seeks to achieve human-like versatility and precision, addressing tasks that require intricate control and adaptability.

Early work in dexterous manipulation concentrated on hardware design and theoretical foundations. Pioneering designs like the Utah/MIT Hand~\cite{jacobsen1986design} and Shadow Hand\cite{ShadowRobotHand} explored high degrees of freedom and biomimetic structures (e.g., tendon-driven mechanisms), while the BarrettHand~\cite{townsend2000barretthand}  showcased underactuated designs for adaptive grasping. Concurrently, theoretical contributions\cite{napier1956prehensile} from Napier, classifying human grasp patterns, and Salisbury and Craig, analyzing multi-fingered force control and kinematics\cite{salisbury1982articulated}, established the groundwork for future research.
Traditional model-based control methods struggled with high-dimensional state spaces and complex contact dynamics, limiting their real-world effectiveness. Learning-based methods, including two-stage and end-to-end approaches, have since become the mainstream, using machine learning to tackle these challenges.

Two-stage methods begin by generating grasping poses, followed by controlling the dexterous hand to achieve them. The key challenge lies in producing high-quality poses from visual observations, tackled through optimization-based~\cite{miller2004graspit,turpin2023fast,zhang2024dexgraspnet}, regression-based~\cite{li2022hgc,liu2020deep} or generation-based~\cite{scenediffuser,grasptta, lundell2021multi,lundell2021ddgc,wei2022dvgg, lu2024ugg,wei2024grasp,liu2024realdex,zhong2025dexgrasp,wei2024dro} strategies, often paired with motion planning. For instance, UGG~\cite{lu2024ugg} uses a diffusion model to unify pose and object geometry generation, while SpringGrasp~\cite{chen2024springgrasp} models uncertainty in partial observations to improve pose quality. Though these methods benefit from decoupled perception and control and simulation data, they remain sensitive to disturbances and calibration errors due to the absence of closed-loop feedback.

End-to-end methods directly model grasping trajectories using reinforcement learning or imitation learning. Reinforcement learning trains policies in simulation for real-world transfer~\cite{chen2022system,chen2023visual,yin2023rotating}, with examples like DexVIP~\cite{mandikal2022dexvip} and GRAFF~\cite{mandikal2021learning} integrating visual affordance cues with reinforcement learning. DextrAH-G~\cite{lum2024dextrah} and DextrAH-RGB~\cite{singh2024dextrah} achieve real-world generalization via large-scale simulation, though sim-to-real gaps and sample efficiency pose challenges. Imitation learning, driven by human demonstrations~\cite{mandikal2021learning,chen2022learning,wang2024dexcap}, excels in complex tasks but struggles with generalization. Innovations like SparseDFF~\cite{wang2023sparsedff} and Neural Attention Field~\cite{wang2024neural} enhance generalization using three-dimensional feature fields, while DexGraspVLA~\cite{zhong2025dexgraspvla} employs a vision-language-action framework: combining pre-trained vision-language models with diffusion-based action controllers. It achieves a 90.8\% success rate across 1,287 unseen object, lighting, and background combinations in zero-shot settings.

\begin{figure}[t]
	\centering
	\includegraphics[width=0.9\linewidth]{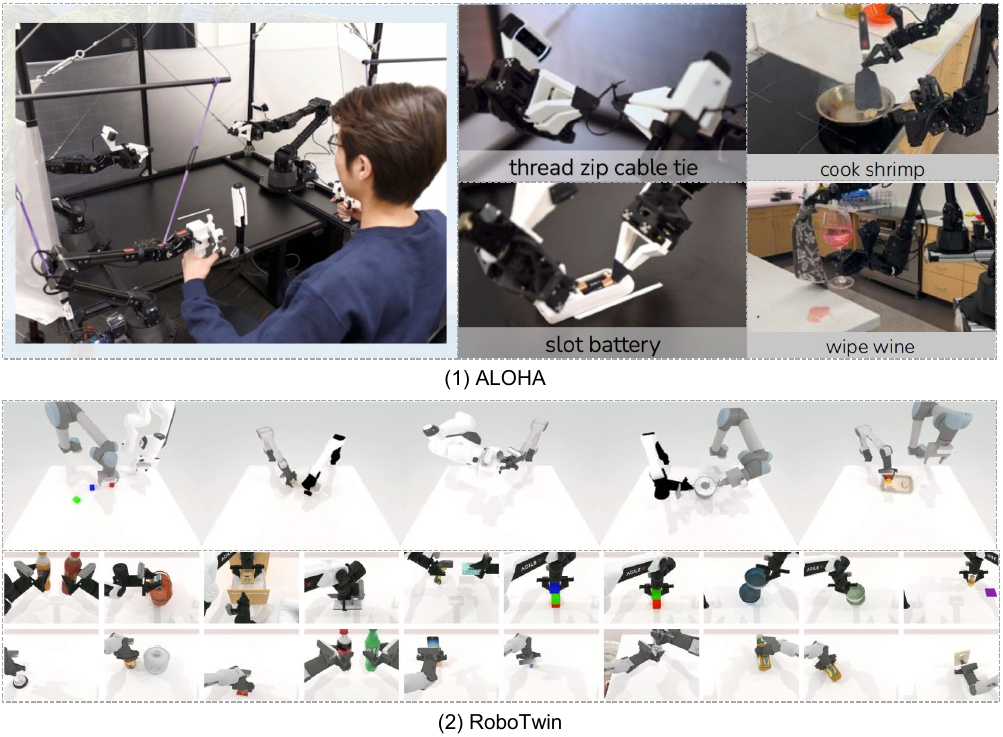}
 \caption{
 (1)The ALOHA series\cite{zhao2023learning, fu2024mobile} feature a low-cost, open-source hardware system that enables the learning of fine-grained, complex, and long-horizon mobile bimanual manipulation tasks, such as slot battery and cooking shrimp. This provides a robotic platform foundation for subsequent research. (2)RoboTwin 2.0\cite{chen2025robotwin20scalabledata} enhances dual-arm manipulation by generating simulated data in simulator. It offers 50 dual-arm tasks, 731 diverse objects, and 5 embodiments for comprehensive research and development.
 }
\end{figure}

\subsubsection{Bimanual Manipulation Task}
Bimanual manipulation refers to robotic tasks that require the coordinated use of two arms, enabling robots to perform complex operations such as cooperative transport, precise assembly, and the handling of flexible or deformable objects \cite{smith2012dual}. These bimanual tasks inherently present greater challenges compared to single-arm manipulation, including high-dimensional state-action spaces, potential for inter-arm and environmental collisions, and the necessity for effective bimanual coordination and dynamic role assignment.

Early research in this area addressed these challenges by introducing inductive biases or structural decompositions to simplify learning and control. For instance, BUDS \cite{grannen2023stabilize} decomposes bimanual manipulation tasks into stabilizer and executor functional roles, which not only reduces the complexity of the dual-arm action space but also enables effective collaboration between arms. This framework has demonstrated robust performance on tasks such as cutting vegetables, pulling zippers, and capping markers. In parallel, SIMPLe \cite{franzese2023interactive} utilizes Graph Gaussian Processes (GGP) to represent motion primitives for bimanual manipulation, ensuring trajectory stability and enabling the system to learn dual-arm coordination from single-arm demonstrations with the aid of kinematic feedback.

With advances in large-scale data collection and imitation learning, end-to-end approaches have become increasingly prominent in bimanual manipulation research. The ALOHA series \cite{zhao2023learning, fu2024mobile, zhao2024aloha} exemplifies this trend by leveraging off-the-shelf hardware and custom 3D-printed components to enable efficient collection of diverse, large-scale demonstration data for high-precision bimanual tasks. These datasets have facilitated the training of end-to-end neural networks with strong generalization capabilities. In particular, ACT \cite{zhao2023learning} combines action chunking with a Conditional Variational Autoencoder (CVAE) framework, allowing robots to learn effective policies from as little as 10 minutes of demonstration data and achieving high success rates in challenging tasks such as battery insertion and cup lid opening. Building on this, Mobile ALOHA \cite{fu2024mobile} introduces a mobile base and further simplifies the ACT pipeline, enabling efficient completion of mobile bimanual tasks. Furthermore, RDT-1B \cite{liu2024rdt} proposes a foundation model for bimanual manipulation based on the diffusion DiT architecture, unifying action representations across heterogeneous multi-robot systems. This approach addresses data scarcity by enabling zero-shot generalization to new tasks and platforms.

While these works have significantly advanced the field of bimanual manipulation, their focus has predominantly been on systems equipped with parallel-jaw grippers. In contrast, bimanual manipulation with dexterous robotic hands introduces additional challenges, particularly in terms of fine-grained coordination and high-dimensional control. Recently, several studies~\cite{li2025maniptrans,mandi2025dexmachinafunctionalretargetingbimanual,wei2024dro} have investigated RL-based methods to transfer human bimanual manipulation skills to robotic dexterous hands, thereby enabling more sophisticated manipulation capabilities.

In summary, bimanual manipulation research has evolved from approaches reliant on strong task-specific priors and structural simplifications to data-driven, end-to-end frameworks that leverage large-scale demonstrations. These advances have significantly improved the robustness, generality, and versatility of bimanual robotic systems, enabling them to tackle an increasingly diverse array of complex manipulation tasks. In addition, ongoing research aimed at adapting these approaches to dexterous robotic hands is expanding the capabilities of bimanual manipulation, enabling robots to perform increasingly intricate and human-like tasks.

\begin{figure}[t]
	\centering
	\includegraphics[width=1\linewidth]{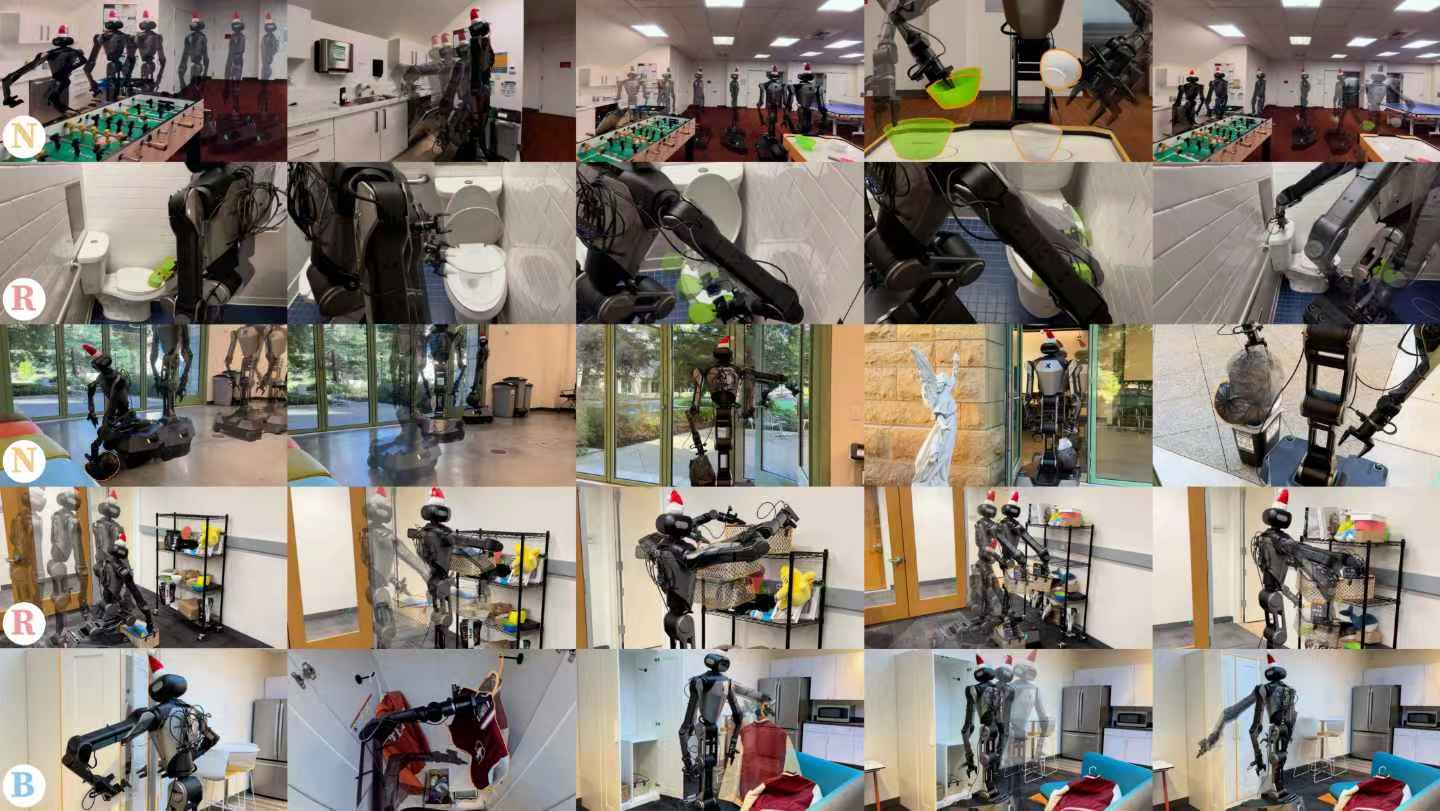}
 \caption{The BRS, introduced by Yunfan Jiang et al.\cite{jiang2025behavior}, enables a humanoid robot to perform a wide range of complex household chores that require Whole-Body Manipulation Control, such as cleaning a toilet, taking out trash, and organizing shelves.
}
\end{figure}

\begin{figure*}[t]
	\centering
	\includegraphics[width=0.8\linewidth]{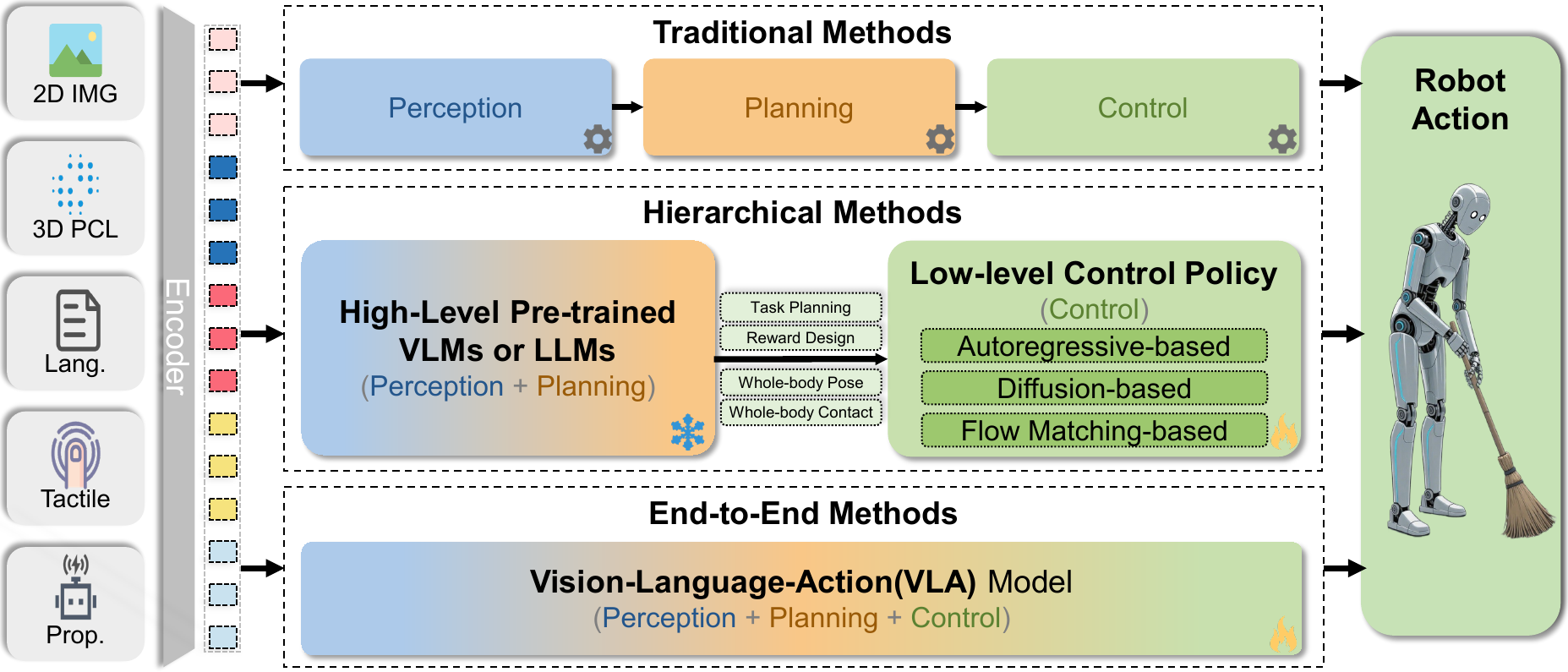}
    \caption{Illustration of three common humanoid robot manipulation frameworks: Traditional, Hierarchical and End-to-End, which shows \textbf{input $\longrightarrow$ action} data flows and structural differences.}
    \label{fig: FM humanoid manipulation.}
\end{figure*}

\subsubsection{Whole-Body Manipulation Control}

Whole-body manipulation refers to the ability of humanoid robots to interact with and manipulate objects using their entire body, including dual arms \cite{smith2012dual}, torso \cite{harada2003whole}, wheeled or legged base \cite{xiong2024adaptive}, and/or other components \cite{cheng2023legs}.

In recent years, significant progress has been made in learning-based whole-body manipulation for humanoid robots, focusing on enhancing robots' autonomy, adaptability, and interaction capabilities in complex environments. One of the trends involves leveraging large pre-trained models (such as LLMs, VLMs, and generative models) to enhance semantic understanding and generalization capabilities. For instance, TidyBot \cite{wu2023tidybot} utilizes the inductive capabilities of LLMs to learn personalized household tidying preferences from a small number of examples. MOO \cite{stone2023open} maps object descriptions from language instructions to visual observations through VLMs, achieving zero-shot operational generalization to unseen object categories. HARMON \cite{jiang2024harmon} combines human motion generation priors with VLM editing to generate diverse and expressive humanoid movements from natural language.

Visual demonstrations also guide learning manipulation skills.
OKAMI \cite{li2024okami} proposes an object-aware redirection method, enabling humanoid robots to imitate skills from a single human RGB-D video and adapt to different object layouts. iDP3 \cite{ze2024generalizable} achieves multi-scene task execution policies trained from single-scene teleoperation data through an improved 3D diffusion strategy. To achieve robust and dexterous whole-body control, OmniH2O\cite{heomnih2o} adopts a reinforcement learning Sim-to-Real approach, training whole-body control policies that coordinate movement and manipulation, and designs a universal kinematic interface compatible with both VR teleoperation and autonomous agents.
The HumanPlus \cite{fu2024humanplus} system combines a Transformer-based low-level control policy with visual imitation strategies, enabling whole-body action demonstration and autonomous learning of complex skills for humanoid robots using only a monocular RGB camera. This system can learn whole-body manipulation and locomotion skills in the real world, such as putting on shoes, standing up, and walking.
WB-VIMA \cite{jiang2025behavior} models the hierarchical structure of whole-body actions and the interdependencies between components within specific humanoid robot morphologies through autoregressive action denoising, predicting coordinated whole-body actions and effectively learning whole-body manipulation for completing challenging real-world household tasks.

\subsubsection{Foundation Models in Humanoid Robot Manipulation }

Foundation Models (FMs) refer to large-scale models pretrained on internet-scale data, including Large Language Models (LLMs), Vision Models (VMs), and Vision-Language Models (VLMs). With their powerful capabilities in semantic understanding, world knowledge integration, logical reasoning, task planning, and cross-modal representation, these models can be directly deployed or fine-tuned for a wide range of downstream tasks.

Foundation Models enable humanoid robots to perform operational tasks in complex, dynamic, and unstructured environments, typically involving complex environmental perception and modeling, abstract task understanding, and autonomous planning for long-sequence and multi-step tasks. Figure \ref{fig: FM humanoid manipulation.} illustrates two main technical paradigms for leveraging foundation models to drive humanoid robot operations.

\noindent
\textbf{Hierarchical Approach }utilizes pretrained language or vision-language foundation models as high-level task planning and reasoning engines to understand user instructions, parse scene information, and decompose complex tasks into sequences of sub-goals. These high-level outputs (typically actionable knowledge or image-language tokens) are then passed to low-level action policies (usually expert policies trained through imitation learning or reinforcement learning) to execute physical interaction actions. Transformer have become a common choice for such low-level policies due to their scalability.

This hierarchical architecture leverages the powerful semantic and logical reasoning capabilities of foundation models while combining the efficiency of lower-level policies in specific action execution, enabling robots to excel in multi-task processing and cross-scenario generalization \cite{ahn2022can,driess2023palm,huang2022inner,liang2023code,lin2024stiv}. The advantages of this approach lie in its modularity and interpretability, but it also faces challenges related to information bottlenecks and semantic gaps between high and low levels. For example, Figure AI demonstrated Helix, a hierarchical VLA model for dexterous manipulation and collaboration between two humanoid robots \cite{FigureAI2025}, while NVIDIA developed GR00T N1, a general foundation model for humanoid robots \cite{bjorck2025gr00t}.
The $\pi_0$ model \cite{black2024pi_0} integrates pretrained vision-language models with flow matching architecture to achieve universal control across various robotic platforms, efficiently executing complex dexterous tasks such as laundry folding and object classification.

\noindent
\textbf{End-to-End Approach} directly incorporates robot operation data into the training or fine-tuning process of foundation models, constructing end-to-end Vision-Language-Action (VLA) models \cite{black2410pi0,brohan2022rt,cheang2024gr,kim2024openvla}. These models directly learn the mapping from multimodal inputs (such as images and language instructions) to robot action outputs. Through pretraining or fine-tuning on large amounts of robot interaction data, VLA models can implicitly learn task planning, scene understanding, and action generation without explicit hierarchy. This end-to-end approach allows models to be holistically optimized for downstream deployment tasks, potentially achieving superior performance and faster response times, but typically requires substantial robot-specific data and offers relatively weaker model interpretability. For instance, Google DeepMind's RT (Robotics Transformer) series \cite{brohan2022rt,brohan2023rt} represents typical VLA models for manipulation.

\subsection{Human-Robot Interaction}
\label{subsec:robotic_hri}
Human-Robot Interaction (HRI) focuses on enabling robots to understand and respond to human needs and emotions, fostering  efficient cooperation, companionship, and personalized services from robots, with broad applications in home, healthcare, education, and entertainment \cite{zhang2024humanlike}. To accurately interpret and adapt to diverse human behaviors, robots need to possess human-like abilities such as multimodal perception, natural language processing, and coordinated control.

% To achieve effective human-robot interaction, it is essential for robots to possess human-like intelligence-related capabilities, such as multimodal perception, natural language processing, affective computing, and coordinated whole-body control. These technologies allow robots to understand, respond to, and adapt to human needs, supporting advanced interactions such as task collaboration, social companionship, and personalized services. In practice, robots are expected to accurately perceive, interpret, and respond to diverse human signals and behaviors, while coordinating their own actions to ensure natural and credible interactions. 

Research in human-robot interaction can be categorized into three primary dimensions: \textbf{Cognitive Collaboration}, \textbf{Physical Reliability}, and \textbf{Social Embeddedness}. These dimensions respectively focus on how robots perceive and understand human cognitive patterns, coordinate physical actions, and integrate effectively into human social contexts. 

Taking a robot navigating through a crowded environment as an example, Cognitive Collaboration involves the robot recognizing pedestrians' potential urgency; Physical Reliability is demonstrated through the robot's adjustments in speed and trajectory to avoid collisions; and Social Embeddedness is exemplified by the robot actively negotiating passage rights using linguistic cues or body language to establish temporary social protocols. Such comprehensive capabilities enable robots to integrate seamlessly into daily human activities, enhancing interaction naturalness and efficiency. The subsequent sections will elaborate on the theoretical foundations, current research advancements, major research directions, and representative literature associated with each dimension.

\subsubsection{\textbf{Cognitive Collaboration:} Understanding and Aligning with Human Cognition}
\begin{figure}
    \centering
    \includegraphics[width=\linewidth]{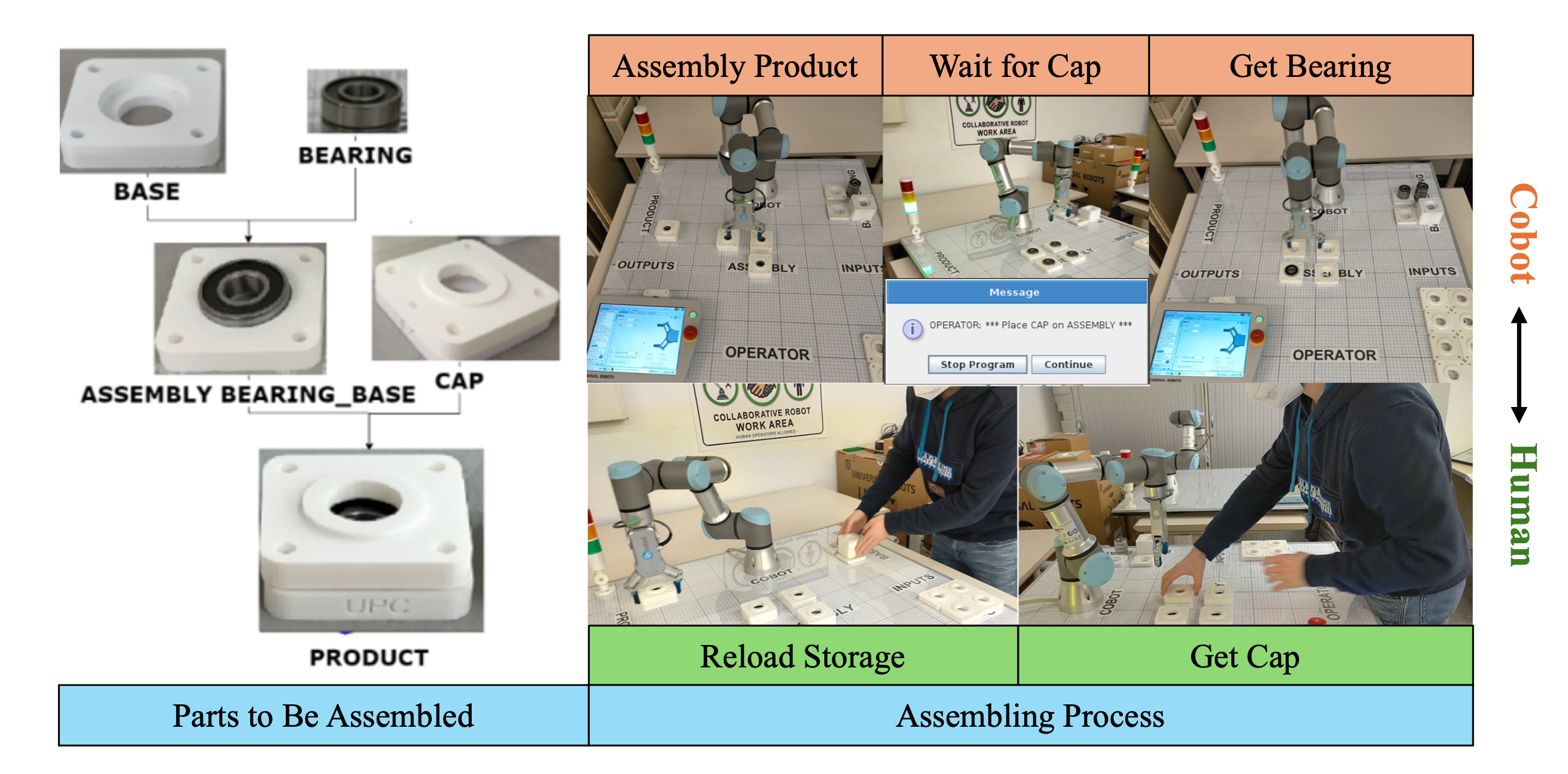}
    \caption{Cognitive interaction analysis enables effective human–robot collaboration in assembly tasks~\cite{chacon2021cognitive}.}
    \label{fig:CognitiveCollaboration}
\end{figure}
Cognitive Collaboration in Human-Robot Interaction refers to the bidirectional cognitive alignment between robots and humans, achieving natural and intuitive communication and cooperation. This collaboration emphasizes not only the robot's passive response to human behavior but also deep cognitive understanding and dynamic adaptation, forming a pattern similar to efficient collaboration among humans. Its core objective is to enable robots to understand not only explicit instructions from humans, such as voice commands and sign language instructions \cite{liang2024llava}, but also implicit intentions (such as emotions and context), dynamically adjusting their behavior to match human cognitive patterns and expectations. The realization of this capability is crucial for improving robots' adaptability in complex scenarios and natural interaction experiences.

Research shows that achieving cognitive collaboration depends on complex cognitive architectures and multimodal information processing capabilities. For example, Lemaignan et al. \cite{lemaignan2017artificial} explored key skills required for robot cognition in social human-robot interaction, including geometric reasoning, contextual assessment, and multimodal dialogue. Robots need to understand human intentions and collaborate with humans to complete joint tasks through these skills~\cite{chacon2021cognitive}. 

Additionally, multimodal intention learning has been identified as a key factor for achieving cognitive collaboration~\cite{semeraro2023human}. For example, integrating facial expressions and body movements to interpret the emotional tone and underlying intentions of spoken instructions can greatly reduce misunderstandings and enhance the naturalness of human-robot interactions~\cite{spezialetti2020emotion}.

Furthermore, cognitive collaboration requires robots to have a deep semantic understanding of the environment and interaction context. Laplaza et al.'s research demonstrates how to infer interaction intentions through contextual semantic analysis of human actions \cite{laplaza2024enhancing}. They proposed a model based on dynamic semantic analysis that can parse the potential goals of human actions in real-time and make predictions combined with environmental information, enabling robots to more accurately collaborate in completing tasks. 

In interaction tasks without direct human participation, cognitive collaboration also plays an important role. For example, in home service robot scenarios, robots can solve goal-oriented navigation tasks through semantic understanding of the environment. This task requires robots to locate specific target objects (such as cups, sofas, or televisions) or designated areas (such as bedrooms and bathrooms) in unknown environments. Works like L3mvn~\cite{yu2023l3mvn}, Sg-Nav~\cite{yin2024online}, Trihelper~\cite{zhang2024trihelper}, CogNav ~\cite{cao2024cognav} and UniGoal~\cite{yin2025unigoal} have improved robot performance in goal-oriented navigation tasks by leveraging large language models (LLMs) to simulate various human cognitive states, such as broad search and contextual search.

In summary, the advancement of cognitive collaboration in human-robot interaction relies on interdisciplinary efforts that encompass cognitive science, semantic understanding, and the development of large multimodal models involving language, vision, and audio perception. As cognitive collaboration capabilities continue to improve, robots will become increasingly adept at supporting various tasks in real-world scenarios.

\begin{figure}[tp]
    \centering
    \includegraphics[width=0.8\linewidth]{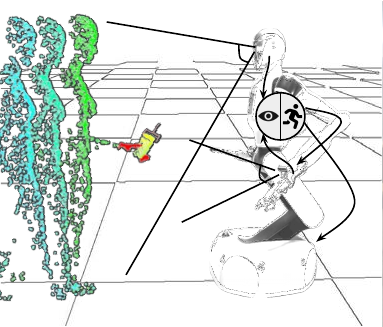}
    \caption{Physical Reliability are ensured through perception and planning~\cite{wang2025mobileh2r}.}
    \label{fig:PhysicalReliability}
\end{figure}
\subsubsection{\textbf{Physical Reliability:} Coordination and Safety of Physical Actions}
Physical reliability in human-robot interaction refers to the effective coordination of force, timing, and distance between humans and robots to ensure safe, efficient, and human-compatible task execution. Its core objective is to enable robots to dynamically respond to real-time changes in human actions, including adjust movement strategies, avoid physical conflicts, and ensure human safety during interaction. To achieve this, research focuses on two main directions: real-time control in physical interaction and large-scale generative dataset construction based on simulation platforms.

Ensuring physical reliability in human-robot interaction relies on advanced motion planning and control strategies that address both coordination and safety between humans and robots~\cite{liu2024application}. Sampling-based planners such as Probabilistic Road Map (PRM)~\cite{geraerts2004comparative} and Rapidly-exploring Random Tree (RRT)~\cite{lavalle1998rapidly}, as well as their extensions, have been widely used to generate collision-free and human-aware trajectories in shared workspaces~\cite{chen2021path,karaman2011sampling,kuffner2000rrt,yi2016homotopy,raessa2020human}. Optimization-based planners including CHOMP~\cite{ratliff2009chomp}, STOMP~\cite{kalakrishnan2011stomp}, ITOMP~\cite{park2012itomp}, TrajOpt~\cite{schulman2014motion,hayne2016considering} and GPMP~\cite{mukadam2018continuous} further improve trajectory quality by minimizing costs related to feasibility and smoothness in dynamic scenarios. These methods are well-suited for application in human-robot collaborative settings to ensure physical reliability~\cite{zhao2018considering,finean2021simultaneous}.

On the control side, impedance control and admittance control provide compliant and safe responses to physical contacts, while adaptive and hybrid controllers further enhance robustness against disturbances and uncertainties~\cite{tarbouriech2019admittance,yu2020cooperative,hameed2023control,sharkawy2022human}. These approaches ensure that robots act reliably and predictably, reducing the risk of injury and discomfort for human collaborators.

However, as robots are increasingly deployed in diverse and unstructured environments, achieving safe and robust physical collaboration requires more than just traditional control strategies. The integration of perception, intention prediction, and real-time adaptation has become essential to enable robots to respond to dynamic changes and complex human behaviors~\cite{ji2020towards,bonci2021human}.
Building upon these advances, recent research has explored imitation learning and reinforcement learning methods, enabling robots to acquire adaptive motion policies directly from data and experience~\cite{zhao2020actor,cui2024task,fan2024learning}. However, the effectiveness of these learning-based approaches depends heavily on the availability of high-quality interaction data. As a result, large-scale generative datasets produced through physical simulation have become a vital resource for enhancing the reliability and safety of robotic actions.
For instance, HandoverSim~\cite{chao2022handoversim} provides a simulation and benchmarking platform for human-robot object handover, leveraging physics engines and trajectory optimization to ensure collision-free and standardized safety evaluation. Building on this, GenH2R~\cite{wang2024genh2r} introduces a simulation environment with extensive 3D models and dexterous grasp generation, enabling the training of generalized handover strategies through imitation learning. Further, MobileH2R~\cite{wang2025mobileh2r} integrates expert demonstrations generated by CHOMP~\cite{ratliff2009chomp} to tackle the challenges of safe and efficient object transfer for mobile robots in dynamic scenarios.

Despite promising progress in current research, ensuring physical reliability in HRI still faces challenges such as high computational cost and limited robustness in complex scenarios. Future exploration is needed to develop more efficient algorithms and enhance adaptability, enabling reliable human-robot interaction in diverse and dynamic environments. 

\subsubsection{\textbf{Social Embeddedness:} Integration with Social Rules and Cultural Norms}
\begin{figure}[tp!]
    \centering
    \includegraphics[width=0.7\linewidth]{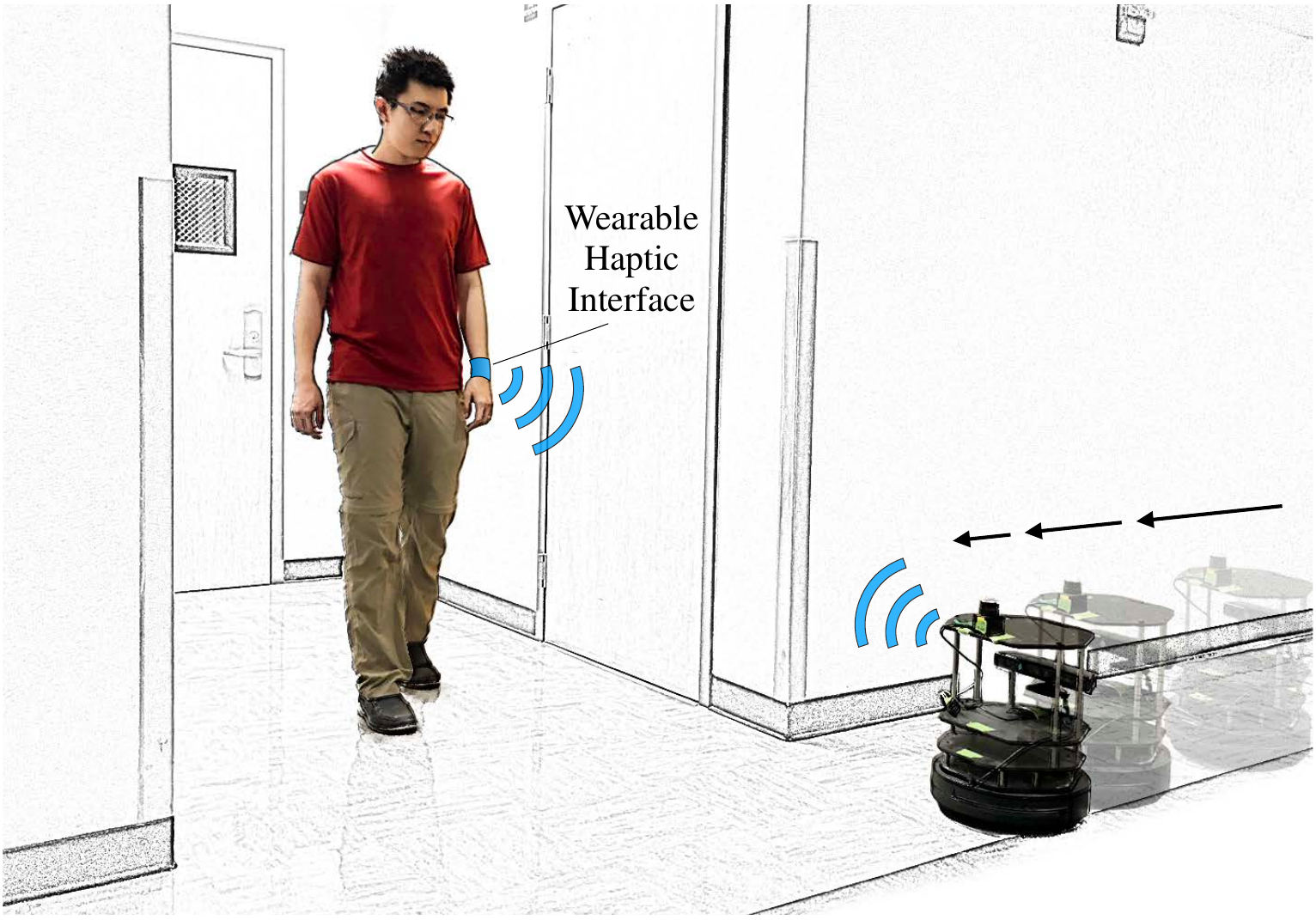}
    \caption{A social navigation scenario where the robot communicates its intent to a human via a wearable haptic interface~\cite{che2020efficient}.}
    \label{fig:SocialEmbeddedness}
\end{figure}

Social embeddedness in human-robot interaction refers to a robot’s ability to recognize and adapt to social norms, cultural expectations, and group dynamics, enabling seamless integration into human environments. This extends beyond task completion to include behaviors such as negotiation, etiquette, and emotional expression. As illustrated in Fig.~\ref{fig:SocialEmbeddedness}, the robot is shown engaging in negotiation with a human partner.
To facilitate the seamless integration of robots into social scenarios, recent research has explored a range of strategies that address both social space understanding and behavior understanding. These two complementary aspects are fundamental for achieving effective social embeddedness in human-robot interaction.

On one hand, social space understanding enables robots to interpret and adapt to the spatial dynamics of human groups. Through the application of spatial understanding in social scenarios, robots can better interpret collaborative or defensive behaviors through concepts such as peripersonal space~\cite{lloyd2009space,coello2021interrelation}. A typical application is enabling robots to navigate and interact more effectively within social environments, thereby providing individuals with visual impairments more appropriate and natural assistance~\cite{gharpure2008robot,fernandes2019review,agrawal2024shelfhelp}.

On the other hand, behavior understanding focuses on decoding the intricacies of human communication from both linguistic and non-linguistic perspectives. Linguistic research explores aspects such as dialogue modeling, conversational structure, and discourse analysis~\cite{hazarika2018conversational,lee2021graph,ghosal2019dialoguegcn,stolcke2000dialogue,lai2023werewolf}, while non-linguistic studies focus on the interpretation of gestures, gaze, and emotional expressions~\cite{lee2024towards,yang2024socially}. To capture these diverse social signals, a range of methods have been proposed for modeling and recognizing non-verbal behaviors, spanning from individual gesture analysis to multi-party, real-world interactions~\cite{liu2022ld,jia2024audio,jahangard2024jrdb,ghosh2024mrac,peng2024tong,cao2025socialgesture}.

Despite these advances, robust social embeddedness remains challenging. Robots must accurately balance task efficiency with social appropriateness, requiring improvements in multimodal perception, long-term adaptation, and the integration of social knowledge into decision-making. Future research should also address lifelong learning, cross-cultural adaptation, and the ethical implications of socially embedded robots, paving the way toward IR-L4 autonomy.

\section{General Physical Simulators}
\label{sec:simulator}
\begin{figure*}
    \centering
    \includegraphics[width=\linewidth]{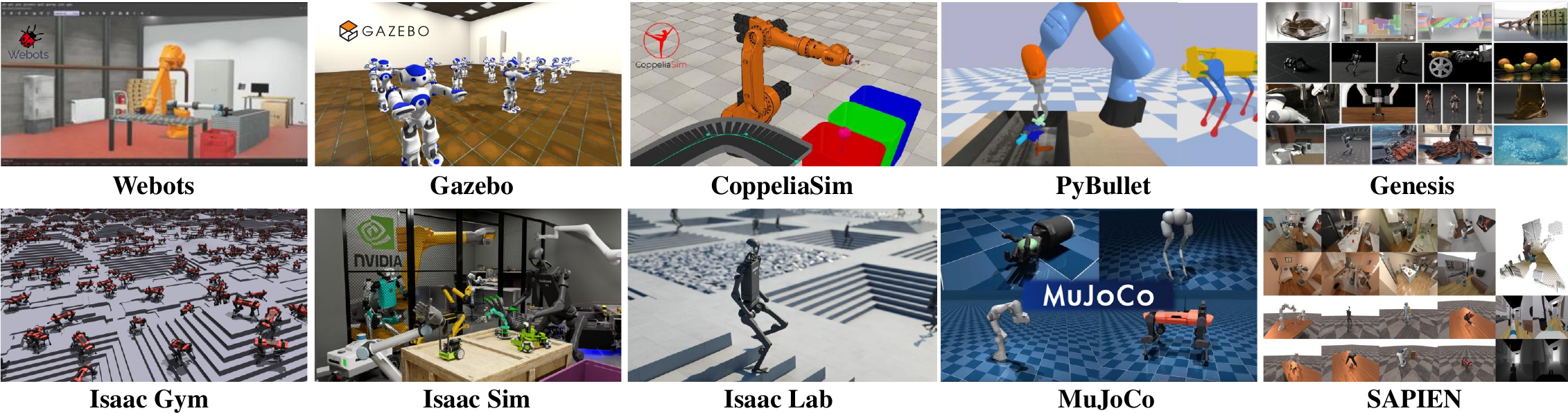}
    \caption{Mainstream Simulators for robotic research.}
    \label{fig:simulators}
\end{figure*}
% \xiao{需要给出一个表,阐述一下仿真器各个的优点和缺点,避免对每个仿真器逐一大篇幅罗列,难以get到核心要点} 
% 强调当前的功能特点,去除一些冗余部分
% 考虑对当前主流仿真器的介绍单开一节

As the traditional robots evolve toward embodied intelligence, their application demands are gradually shifting from structured industrial settings to open and dynamic human living environments. However, algorithmic research on intelligent robots continues to face significant data bottlenecks, which severely limit the generalization capability and applicability of related algorithms. 

Specifically, current data collection for intelligent robots still faces the following challenges: 
1) \textit{\textbf{Cost and Safety problem}}: Data collection strategies based on teleoperation~\cite{fu2024mobile, cheng2024open} or universal manipulation interfaces (UMI)~\cite{chi2024universal} impose hardware requirements and demand a certain level of operator proficiency, bringing financial problems to the data collection process. Moreover, the physical experiments also raise the safety problem related to hazardous and dangerous scenarios or robotic actions.
2) \textit{\textbf{Control and Repeatability problem}}: The data collection process is influenced by numerous factors such as lighting, background, and sensor noise, which affect data quality. To control all these factors, the collection scenario must be carefully designed. Additionally, because the scene cannot be fully controlled, it becomes challenging to replicate experiments with identical settings for verification.

% 1) Insufficient data volume: The acquisition of real-world data is constrained by hardware costs and experimental risks during collection (e.g., potential mechanical damage from falls), making it difficult to obtain large-scale, multimodal datasets in real-world scenarios.
% 2) Cumbersome and expensive data collection process: Data collection strategies based on teleoperation~\cite{fu2024mobile, cheng2024open} or universal manipulation interfaces (UMI)~\cite{chi2024universal} impose hardware requirements and demand a certain level of operator proficiency, making the overall process cumbersome. Moreover, the physical experiments raise the financial problem related to system development and data collection. 
% 3) Difficulties in collecting long-tail scenarios: In addition, edge cases encountered by humanoid robots in real-world environments are difficult to anticipate and tackle during the data collection process.
% 随着人形机器人向具身智能方向演进,其应用需求已逐步呈现从结构化工业环境扩展至开放动态的人类生活空间趋势。然而,人形机器人算法研究仍绵连较大的数据瓶颈,严重制约人形机器人相关算法的泛化能力与应用场景。具体而言,人形机器人数据目前仍面临以下挑战：1）\textbf{数据量不足：}真实世界数据获取受限于硬件成本以及采集过程中的实验风险（如跌倒可能造成机械损伤）,难以大规模地获取真实场景数据；2）\textbf{数据采集流程繁琐：}基于遥操作~\cite{fu2024mobile, cheng2024open}或者通用操作接口（UMI）~\cite{chi2024universal}的采集策略,不仅有设备上的要求,还要求采集者具备一定的熟练度,数据采集的流程较为繁琐；3）\textbf{长尾场景难以收集：}；除此之外,针对人形机器人在真实场景中面临的边界情况,在数据采集流程中很难进行预先设计。

To address the aforementioned bottlenecks, simulation-to-reality (Sim2Real) transfer has emerged as a key approach to overcome data limitations. This paradigm constructs high-fidelity virtual environments to generate multimodal synthetic data, offering several key advantages: 
1) \textit{\textbf{Significantly improved data generation efficiency}}: The use of GPU-accelerated physics engines and distributed rendering in simulators enables the rapid generation of large volumes of simulated data, which makes the data collection process cost-effective and safe;
2) \textit{\textbf{Automated annotation and controllable semantic labeling}}: Simulators can directly output pixel-level ground truth, including semantic segmentation maps, depth maps, and 6D object poses, allowing precise control of the simulated environments. Furthermore, simulators also provide an efficient, safe, and reproducible testing ground for debugging, validating, and optimizing robot perception, control, and planning algorithms, enhancing their robustness and reliability during actual deployment. %long-tail conditions can be systematically introduced through rule-based configurations, enabling the effective generation of rare scenarios.
% 针对上述瓶颈,仿真-真实迁移（Sim2Real）成为突破数据壁垒的核心路径。该范式通过构建高保真虚拟环境,生成多模态合成数据,其优势在于：1）\textbf{大幅提升数据生成效率}：通过仿真器基于GPU加速的物理引擎以及分布式渲染等功能,能够快速产生大批量的仿真数据；2）\textbf{自动化标注与可控的语义标签}：仿真器可直接输出像素级真值,提供包括语义分割图、深度图、物体6D姿态等信息在内的真值,进一步地,还能通过规则等方式引入长尾条件,有效地生成长尾场景。

% \ZZH{Besides the above advantages, } 
This chapter focuses on mainstream simulators in contemporary robotics research. We first introduce the widely adopted simulation platforms, followed by an introduction of their diverse capabilities, encompassing physical simulation properties, rendering capabilities, and sensor/joint component support abilities. Finally, we summarize the current development of robotic simulators and discuss prospective research direction.
% , particularly in terms of physical modeling and training efficiency. It examines the capabilities of mainstream simulators, including physical simulation fidelity, rendering capabilities, and sensor support abilities.
% 本章节重点探讨适用于人形机器人任务中的不同仿真器在物理建模与训练效率上的优势,针对主流仿真器的不同能力（物理仿真能力,真实感渲染能力,传感器支持能力,额外功能支持等）进行探讨,最后讨论仿真器的研究现状与未来方向。

\subsection{Mainstream Simulators}
~\label{subset:mainstream_simulators}

\textbf{Webots} \cite{Webots04}, introduced by Cyberbotics Ltd. in 1998, provides an integrated framework for robot modeling, programming, and simulation, and has been widely adopted in education, industry, and academia. It was open-sourced in 2018, significantly enhancing its accessibility for research and education worldwide. Webots offers a rich set of prebuilt robot models, supports various sensor modalities, is cross-platform, and provides multi-language APIs, making it a versatile and user-friendly tool. However, it lacks support for deformable bodies and fluid dynamics, and shows limitations in advanced physics and large-scale learning tasks.

\textbf{Gazebo} \cite{1389727}, developed in 2002 by Andrew Howard and Nate Koenig at the University of Southern California, is a widely adopted open-source simulator known for its extensibility and integration with robotic middleware. In addition to supporting a broad range of sensors and robot models, 
Gazebo features a modular plugin system that enables users to customize simulation components, including models, controllers and sensors.
The system also allows users to interface with various physics and rendering engines.
However, it shares similar limitations with Webots in terms of advanced physical interaction, lacking support for suction-based grasping, deformable objects, and fluid dynamics.
Both Webots and Gazebo are tightly integrated with the Robot Operating System (ROS), enabling seamless testing and validation of robotic algorithms within the ROS ecosystem. Nonetheless, they do not offer native support for parallel execution or reinforcement learning workflows.
% 其中,Webots和Gazebo是相对早期的仿真平台。机器人模拟器Webots于1998年由Cyberbotics Ltd.推出,提供包括建模、程控、仿真等完整的机器人模拟框架,被广泛应用于教育、工业、学术等用途。为了促进教育和研究的普及,Webots于2018年12月转为开源项目,采用Apache 2.0许可证,向全球用户免费开放。Webots内置多种机器人模型,支持多种传感器模拟,跨平台兼容,支持多种语言,是一个功能完备的机器人仿真平台。Gazebo于2002年由南加州大学的Andrew Howard和Nate Koenig开发。除了多种传感器、机器人模型库等基础仿真功能,Gazebo还支持以插件的方式扩展,使得用户不仅可自定义仿真模型、控制器和传感,还可以集成不同物理引擎与渲染引擎等后端。此外,Webots与ROS（Robot Operating System）深度集成,方便在仿真中测试和验证ROS生态系统中的算法和工具。

% simulator演进
\textbf{MuJoCo} (Multi-Joint dynamics with Contact)~\cite{todorov2012mujoco} is a physics engine developed in 2012 by Professor Emo Todorov’s group at the University of Washington, and was later acquired by Google DeepMind in 2021. Designed specifically for simulating contact-rich dynamics in articulated systems, MuJoCo offers high-precision physical computation, an optimized generalized-coordinate formulation, and native support for biomechanical modeling.
MuJoCo formulates contact constraints as convex optimization problems. This approach enables accurate simulation of complex interactions. These interactions include soft contacts, frictional behavior and tendon-actuated motion. At the same time, it mitigates common limitations of traditional engines, such as interpenetration and instability. It maintains both numerical accuracy and computational efficiency even under large simulation time steps.
The engine supports a lightweight XML-based modeling format (MJCF/URDF) and provides multi-language APIs (Python, C++), enabling rapid model construction and iteration. These features have led to its widespread adoption in robotics and reinforcement learning research.
Overall, MuJoCo excels in contact dynamics and RL applications with ROS integration support, but is limited in rendering capabilities due to its OpenGL backend and lacks support for fluid, discrete element (DEM), or LiDAR simulation.

With the rapid advancement of robotics and artificial intelligence, the demand for efficient simulation tools has increased. In response, Erwin Coumans, the original developer of the Bullet physics engine, introduced \textbf{PyBullet} in 2017\cite{coumans2019}. PyBullet wraps the powerful capabilities of Bullet in a Python interface, aiming to provide researchers and developers with a lightweight and easy-to-integrate simulation platform.
Although PyBullet falls slightly short of some mainstream simulators in terms of simulation fidelity and feature richness, it has gained widespread adoption in academia due to its open-source nature, lightweight design, user-friendly Python interface, and ease of use. Over time, it has fostered a large and active user community.
% \ZZH{PyBullet is valued for being lightweight with fast simulation speed, supporting multi-instance simulation, reinforcement learning, and inverse kinematics/dynamics. Its limitations include restricted support for deformable objects, OpenGL-based rendering with limited rendering capabilities, no support for IMU simulation, and unavailable native ROS support.}
% 随着机器人学和人工智能领域发展,对高效仿真工具的需求增加,Bullet引擎的作者Coumans在2017年推出了PyBullet,将Bullet的强大功能封装为Python接口,旨在为研究人员和开发者提供一个轻量级、易于集成的仿真平台。尽管在功能丰富性和仿真保真度方面,PyBullet相较于部分主流仿真平台略有不足,但凭借其开源、轻量、易用的特点,以及完善的Python接口支持,PyBullet在学术界得到了广泛应用,并发展出了一个规模庞大且活跃的用户社区。

\textbf{CoppeliaSim}\cite{coppeliaSim}, formerly known as V-REP (Virtual Robot Experimentation Platform), was initially released around 2010 by Toshiba Corporation as a general-purpose robot simulation software aimed at developers, researchers, and educators. Since 2019, it has been further developed and maintained by the Swiss company Coppelia Robotics.
The core strength of CoppeliaSim lies in its distributed control architecture, which allows Python and Lua scripts or C/C++ plugins to function as individual synchronous controllers. Additional asynchronous controllers can be executed via various middleware solutions (e.g., ROS, remote APIs) and support integration with C/C++, Python, Java, and MATLAB across separate processes, threads, or even machines. The educational edition of CoppeliaSim is open-source and freely available, making it widely adopted in academic and educational settings. 
% Coppeliasim前身是于2010年左右由东芝公司首次发布的机器人仿真软件V-REP（Virtual Robot Experimentation Platform）,主要面向机器人开发者、研究人员以及教学领域。2019年起被瑞士公司Coppelia Robotics升级并维护。Coppeliasim的核心在于它是围绕分布式控制架构构建的,该架构具有Python和Lua脚本,或者C/C++插件充当单独的同步控制器。额外的异步控制器可以通过各种中间件解决方案（ROS、远程 API等）和C/C++、Python、Java和Matlab等编程语言在另一个进程、线程或机器中执行。Coppeliasim的教育版是开源免费的,受到学界与教育界广泛使用。

% 其中Nvidia推出了系列工作
With the growing convergence of robotics, reinforcement learning, and photorealistic simulation, the demand for scalable, GPU-accelerated simulation platforms has surged. The NVIDIA Isaac series addresses this need by establishing a closed-loop technological ecosystem for robotics development and embodied AI research. Its evolution reflects a shift from isolated acceleration tools to a comprehensive, full-stack simulation infrastructure.
\textbf{Isaac Gym}~\cite{makoviychuk2021isaac}, introduced in 2021, pioneered large-scale GPU-accelerated physics simulation by enabling the parallel training of thousands of environments simultaneously. Built upon NVIDIA’s PhysX engine, it significantly improves sample efficiency in tasks such as locomotion control and policy learning for legged robots.
However, Isaac Gym is limited in rendering fidelity due to its lack of ray tracing support, and does not offer fluid or LiDAR simulation capabilities.

\textbf{Isaac Sim}~\cite{nvidia_isaac_sim} is subsequently introduced by NVIDIA with integration of the Omniverse platform~\cite{nvidia_omniverse}, a full-featured digital twin simulator built on Omniverse. It incorporates the PhysX 5 physics engine and RTX-based real-time ray tracing, enabling high-fidelity LiDAR simulation with millimeter-level precision. By adopting the USD (Universal Scene Description)~\cite{aousd} standard, Isaac Sim supports physically accurate simulation of multimodal sensors, including RGB-D cameras and IMUs.
The 2025 release of Isaac Sim 5.0 further improved rigid-body momentum conservation, added joint visualization tools, and introduced simulation data analytics modules, achieving robotic arm grasping precision within 0.1 mm.
% Despite its advanced capabilities, Isaac Sim remains closed-source, limiting customizability and low-level modification.

\textbf{Isaac Lab}~\cite{isaaclab_docs} is a modular reinforcement learning framework built on top of Isaac Sim, designed to streamline and optimize the robot learning pipeline. It employs a tiled rendering technique to efficiently process multi-camera inputs, improving training throughput by approximately 1.2×. Isaac Lab supports both imitation learning and reinforcement learning paradigms, enabling rapid policy construction from demonstration datasets in HDF5 format.
Isaac Gym's parallel computing capabilities are now deeply integrated into Isaac Sim's underlying architecture. Built upon Isaac Sim, Isaac Lab employs modular design to abstract simulation environments into configurable task units.
% 这句话读起来有点怪了,拆成两句话了
% With the parallel computing capabilities of Isaac Gym now deeply embedded within the architecture of Isaac Sim, Isaac Lab abstracts simulation environments into configurable task units through a modular design. 
This evolving software stack is advancing toward support for heterogeneous physics engines (including both rigid and soft-body dynamics), sim-to-real transfer, and generative AI-driven scene synthesis (e.g., the Cosmos world model), thereby providing a unified infrastructure for embodied intelligence research—from low-level physics to high-level behavior learning. However, it inherits the high hardware requirements of Isaac Sim, which may limit accessibility for resource-constrained users.
Both Isaac Sim and Isaac Lab have released early developer preview versions and are expected to be gradually open-sourced.
% NVIDIA Isaac系列仿真器构成机器人开发与强化学习研究的技术生态闭环,其演进路径反映了从专用加速工具向全栈式仿真平台的升级过程。
% Isaac Gym~\cite{makoviychuk2021isaac}是由NVIDIA于2021年提出的高性能物理仿真器,开创了基于GPU加速的并行物理仿真范式,通过PhysX引擎支持数千个环境同步训练,在四足机器人步态优化等任务中实现训练效率的显著提升。
% 随着Omniverse平台的技术整合,NVIDIA基于Omniverse平台构建的全功能数字孪生平台Isaac Sim\cite{nvidia_isaac_sim},集成PhysX 5物理引擎与RTX实时光追技术,可模拟毫米级精度的激光雷达点云数据,并通过USD场景格式实现多模态传感器（RGB-D摄像头、IMU等）的物理级仿真。2025年发布的Isaac Sim 4.5版本进一步强化了刚体动量守恒算法,新增关节可视化工具和仿真数据统计模块,使得工业机械臂抓取误差控制在0.1mm级别。
% Isaac Lab仿真器~\cite{isaaclab_docs}是由NVIDIA基于Isaac Sim进一步推出的模块化强化学习框架,专注于机器人学习流程优化,其平铺渲染技术将多摄像头输入合并处理,训练吞吐量提升1.2倍,同时整合模仿学习与强化学习范式,支持通过HDF5格式示教数据快速构建行为策略。
% 如今,Isaac Gym的并行计算能力已深度融入Isaac Sim的底层架构,而构建于Isaac Sim之上的Isaac Lab通过模块化设计将仿真环境抽象为可配置任务单元,当前该技术栈正朝着物理引擎异构化（融合刚体/柔体动力学）、仿真-现实数据对齐以及生成式AI驱动场景合成（如Cosmos世界模型）方向演进,为具身智能研究提供从底层物理仿真到高层策略学习的完整基础设施支撑。

% 最近的
\textbf{SAPIEN} (SimulAted Part-based Interactive ENvironment), introduced in 2020 by researchers from the University of California, San Diego and collaborating institutions~\cite{xiang2020sapien}, is a simulation platform designed for physically realistic modeling of complex, part-level interactive objects, extending beyond traditional rigid-body dynamics. To support research in articulated object manipulation, the authors released PartNet-Mobility\cite{xiang2020sapien}, a comprehensive dataset featuring motion-annotated, articulated 3D objects.
Building on the SAPIEN engine, the research team later introduced the ManiSkill\cite{gu2023maniskill2} and ManiSkill3\cite{taomaniskill3} benchmarks, which provide diverse manipulation tasks, high-quality demonstrations, and efficient, parallelized data collection pipelines. This ecosystem has become a widely used benchmark suite for evaluating manipulation policies and embodied intelligence algorithms in realistic physics-based environments.
SAPIEN offers interactive visualization, supports RGB and IMU sensors, and is tightly coupled with the ManiSkill benchmark. However, it lacks support for soft-body and fluid dynamics, ray tracing, LiDAR, GPS, and ROS integration, which limits its applicability in broader robotic simulation scenarios.
% SAPIEN（SimulAted Part-based Interactive ENvironment）由加州大学圣地亚哥分校（UC San Diego）和其他机构的研究人员开发,最早发表于2020年,在论文《SAPIEN: A SimulAted Part-based Interactive ENvironment》中提出。SAPIEN的核心优势在于其在刚体的支持之上还提供对部件级复杂可交互物体的仿真支持（同时发布PartNet-Mobility,一个带有运动注释的铰链物体数据集）。在SAPIEN的基础上,其团队还进一步提出了ManiSkill、ManiSkill3等基准,提供丰富的任务、演示,且支持并行高效的数据采集。

\textbf{Genesis}~\cite{Genesis}, released in 2024, is a general-purpose physical simulation platform developed by a global consortium of researchers. Its core objective is to unify a wide array of physics solvers—including rigid body dynamics, Material Point Method (MPM), Smoothed Particle Hydrodynamics (SPH), Finite Element Method (FEM), Position-Based Dynamics (PBD), and stable fluid solvers—within a high-fidelity framework capable of capturing complex physical phenomena with maximal realism.
A key innovation in Genesis is its generative data engine, which enables users to specify simulation scenarios and generate multimodal datasets from natural language prompts. Built with differentiability as a fundamental design principle, Genesis is well-suited for applications in embodied intelligence, physical reasoning, and differentiable simulation.
According to public benchmarks, Genesis demonstrates a 2.70× to 11.79× throughput advantage over Isaac Gym across batch sizes ranging from 512 to 32,768 environments.
% Genesis significantly outperforms existing platforms such as Isaac Gym and MJX in simulation throughput and scalability. 
The platform is currently undergoing staged open-source release.
Despite its strengths, Genesis does not yet support LiDAR or GPS simulation, nor does it provide ROS integration.
% Genesis是由世界各地高校多人组成的团队开发,于2024年发布的一个通用的物理仿真平台,旨在将不同的物理解算器（包括刚体、MPM、SPH、FEM、PBD、稳定流体等等）统一到一个框架中,以最高保真度重建物理世界。此外其核心特色在于提供一个生成式数据引擎,可将用户提示的自然语言描述转换为各种数据模式。为了更好地支持具身智能、物理智能等应用,Genesis被设计为完全可微分。另外,其公开的报告显示Genesis在仿真效率上远超Isaac Gym、MJX等其他主流仿真平台。目前,Genesis的各种功能在陆续开放与开源中。

In addition, NVIDIA \textbf{Newton}\cite{huang2025newton} is an open-source physics engine jointly developed by NVIDIA, Google DeepMind, and Disney Research in 2025. Targeting at high-fidelity simulation and robot learning, Newton provides a full-stack framework spanning from basic physical modeling to complex multiphysics interactions. Built on top of the NVIDIA Warp framework, it achieves over 70$\times$ simulation speedup via GPU acceleration and supports rigid/soft-body dynamics, contact and friction modeling, as well as custom solver integration.
Newton is designed to be deeply compatible with existing robot learning platforms such as MuJoCo Playground and Isaac Lab, allowing seamless reuse of existing robot models and training pipelines. With its differentiable physics engine, Newton enables backpropagation of gradients through simulation, offering a solid mathematical foundation for learning-based control optimization. Additionally, its OpenUSD-based scene construction capability aligns physical laws with virtual environments at a fine-grained level. Although still in its early development stage, the Newton ecosystem is rapidly evolving and aims to bridge the sim-to-real gap in domains such as industrial manipulation and humanoid motion planning.
% NVIDIA Newton~\cite{huang2025newton}是由NVIDIA联合Google DeepMind与Disney Research于2025年提出的开源物理引擎,该仿真器主要针对机器人学习与高保真仿真场景,并提供从基础物理模拟到复杂多物理场交互的全栈能力。基于NVIDIA Warp框架构建,Newton通过GPU加速实现了70倍以上的仿真效率提升,支持刚体/柔体动力学、接触摩擦建模及自定义求解器集成,其与MuJoCo Playground、Isaac Lab的深度兼容性,使得开发者可直接复用现有机器人模型与训练管线。通过可微分物理特性,Newton允许梯度在仿真中反向传播,为优化机器人控制策略提供数学基础,而基于OpenUSD的场景构建能力,则实现了物理规则与虚拟环境的精准对齐,该项目的生态尚在发展中,致力于推动工业机械臂抓取、人形机器人运动规划等领域的仿真-现实差距缩小。

% 人形机器人背景介绍
% % 人形机器人的数据困境：1.人工构建/捕获的数据集规模不支持scaling up；2.生成式方法难以满足
% 然而,当前人形机器人发展仍严重受到数据的制约。具体而言,人形机器人目前仍面临以下问题：1. ；2.

% 基于仿真器进行数据生成的优势：1.物理真实、视觉真实；2.能够直接在仿真环境进行强化学习的训练

% 市面主流仿真平台的介绍（Isaac Sim/Gym/Lab, Newton, Mujoco, PyBullet,Gazebo,CoppeliaSim,Genesis）

\begin{table*}[]
	\centering
    \resizebox{\textwidth}{!}{
	\begin{tabular}{lccccccccc}
 \toprule
     \rowcolor{gray!10}
	Simulator & Physics Engine & Suction & \makecell{Random\\external forces} & \makecell{Deformable\\objects} & \makecell{Soft-body\\contacts} & \makecell{Fluid\\mechanism} & \makecell{DEM\\simulation} & \makecell{Differentiable\\physics}\\
 \midrule
	Webots & ODE(default) & \textcolor{darkgreen}{\checkmark} & \textcolor{darkgreen}{\checkmark} & \textcolor{red}{\ding{55}} & \textcolor{darkgreen}{\checkmark} & \textcolor{darkgreen}{\checkmark} & \textcolor{red}{\ding{55}}  & \textcolor{red}{\ding{55}}   \\
     \rowcolor{gray!10}
	Gazebo & DART(default) & \textcolor{darkgreen}{\checkmark} & \textcolor{darkgreen}{\checkmark} & \textcolor{red}{\ding{55}} & \textcolor{darkgreen}{\checkmark} & \textcolor{darkgreen}{\checkmark} & \textcolor{darkgreen}{\checkmark} & \textcolor{red}{\ding{55}}\\
	MuJoCo & MuJoCo & \textcolor{darkgreen}{\checkmark} & \textcolor{darkgreen}{\checkmark} & \textcolor{darkgreen}{\checkmark} & \textcolor{darkgreen}{\checkmark} & \textcolor{red}{\ding{55}} & \textcolor{red}{\ding{55}} & \textcolor{darkgreen}{\checkmark} \\
     \rowcolor{gray!10}
	CoppeliaSim & Bullet, ODE, Vortex, Newton & \textcolor{darkgreen}{\checkmark} & \textcolor{darkgreen}{\checkmark} & \textcolor{red}{\ding{55}} & \textcolor{darkgreen}{\checkmark} & \textcolor{red}{\ding{55}} & \textcolor{red}{\ding{55}} & \textcolor{red}{\ding{55}} \\
	PyBullet & Bullet & \textcolor{red}{\ding{55}} & \textcolor{darkgreen}{\checkmark} & \textcolor{darkgreen}{\checkmark} & \textcolor{darkgreen}{\checkmark} & \textcolor{red}{\ding{55}} & \textcolor{red}{\ding{55}} & \textcolor{darkgreen}{\checkmark}  \\
     \rowcolor{gray!10}
	Isaac Gym & PhysX, FleX(GPU) & \textcolor{red}{\ding{55}} & \textcolor{darkgreen}{\checkmark} & \textcolor{darkgreen}{\checkmark} & \textcolor{darkgreen}{\checkmark} & \textcolor{red}{\ding{55}} & \textcolor{red}{\ding{55}} & \textcolor{red}{\ding{55}} \\
	Isaac Sim & PhysX(GPU) & \textcolor{darkgreen}{\checkmark} & \textcolor{darkgreen}{\checkmark} & \textcolor{darkgreen}{\checkmark} & \textcolor{darkgreen}{\checkmark} & \textcolor{darkgreen}{\checkmark} & \textcolor{red}{\ding{55}} & \textcolor{red}{\ding{55}}\\
     \rowcolor{gray!10}
	Isaac Lab & PhysX(GPU) & \textcolor{red}{\ding{55}} & \textcolor{darkgreen}{\checkmark} & \textcolor{darkgreen}{\checkmark} & \textcolor{darkgreen}{\checkmark} & \textcolor{red}{\ding{55}} & \textcolor{red}{\ding{55}} & \textcolor{red}{\ding{55}} \\
	SAPIEN & PhysX & \textcolor{red}{\ding{55}} & \textcolor{darkgreen}{\checkmark} & \textcolor{red}{\ding{55}} & \textcolor{darkgreen}{\checkmark} & \textcolor{red}{\ding{55}} & \textcolor{red}{\ding{55}} & \textcolor{red}{\ding{55}} \\
     \rowcolor{gray!10}
	Genesis & Custom-designed & \textcolor{red}{\ding{55}} & \textcolor{darkgreen}{\checkmark} & \textcolor{darkgreen}{\checkmark} & \textcolor{darkgreen}{\checkmark} & \textcolor{darkgreen}{\checkmark} & \textcolor{red}{\ding{55}} & \textcolor{darkgreen}{\checkmark}  \\
     \bottomrule
	\end{tabular}}
	\caption{Comparison of physical simulation across different simulators. \textcolor{darkgreen}{\checkmark}: support, \textcolor{red}{\ding{55}}: lack of support.}
	\label{tab:simulator_phy}
\end{table*}

\subsection{Physical Properties of Simulators}
\label{subsec:simulator_properties}

% 背景介绍
Intelligent robots are required to perform human-like behaviors such as grasping, walking, and collaboration in complex and dynamic environments. In this context, the physical modeling capabilities of the simulator directly determine the realism of the generated data and the effectiveness of policy transfer. High-fidelity physical property simulation not only enhances the realism of environmental interactions (e.g., soft body deformation, center-of-mass shifts caused by fluid dynamics), but also improves algorithm generalization by introducing stochastic perturbations that prevent policy overfitting to the simulated environment. This, in turn, broadens the applicability of humanoid robot algorithms across a wider range of scenarios. This section provides a comparative analysis of the simulators discussed above, focusing on their capabilities in simulating various physical properties. Following the categorization in~\cite{collins2021review}, this subsection selects physical attributes that are especially important for humanoid robots and world model, and compares how different simulators support them. Table~\ref{tab:simulator_phy} summarizes the support of different simulators for various types of physical simulation.
% 人形机器人需要在复杂、动态的环境中实现如抓取、行走、协作等类人的行为。在此过程中,仿真器的物理建模能力直接决定了生成数据的真实性以及训练策略迁移的有效性。高精度的物理属性仿真不仅能够提升环境交互的真实性（如柔性物体形变,流体带来的物体重心改变等）；还能够通过增加随机扰动,避免策略过拟合到仿真场景,增强算法的泛化能力,从而进一步拓展人形机器人算法的应用场景边界。本小节重点对比上述各仿真器对不同物理属性的仿真能力。

\subsubsection{Suction}

In robotic simulation, suction modeling primarily refers to the non-rigid attachment behavior at contact interfaces, such as simulating vacuum suction for object grasping. This function is widely used in industrial automation and warehouse picking.

Currently, mainstream robotic simulation platforms vary in implementation of suction effects. MuJoCo relies on user-defined logic to simulate suction by detecting contact and applying external forces or creating virtual links, offering limited accuracy and control. Gazebo implements suction via plugins that dynamically create joints based on contact, enabling more flexible control of mechanical behavior. In contrast, Webots, CoppeliaSim and Isaac Sim provide native module support for suction, offering greater ease of use.
% Mainstream robotic simulation platforms exhibit distinct approaches to implementing suction effects.
% Webots provides a native VacuumGripper node that dynamically creates and breaks links with objects to simulate the suction process.
% In MuJoCo, users can implement suction effects through custom logic by detecting contact and applying constraints or external forces.
% CoppeliaSim features a built-in suction pad component that uses script-based control to establish virtual connections with target objects.
% Isaac Sim employs the Surface Gripper extension to create joint-based connections between the gripper and target object, enabling suction and release.
% Gazebo can simulate suction and release processes by detecting contacts and dynamically creating joints via plugins. 
% The Genesis simulation platform currently does not provide native support for suction effects. However, users can manually control the object's position to follow the end-effector of the suction gripper, thereby simulating suction behavior. The developers officially state that the related feature is currently under development.
% 吸附效应仿真主要是指仿真器模拟真空吸附等接触面的非刚性附着行为,其可应用于模拟使用真空吸盘抓取光滑表面物体（玻璃、金属）或攀爬垂直墙面等任务。
% 在多个机器人仿真平台中,对吸附效应（suction）的支持程度各不相同。Isaac Sim提供了名为Surface Gripper的扩展,用于模拟吸附器的行为。该功能通过在接触时创建关节连接,实现吸附效果。然而,该功能目前不支持与粒子系统（如布料）交互的吸附。Isaac Lab作为基于Isaac Sim的框架,继承了其吸附器仿真能力。MuJoCo本身不提供内置的吸附器仿真功能,但有开发者通过自定义约束或控制逻辑来模拟吸附器的行为。

% 介绍每个simulator仿真器吸附具体怎么做的,相同的就合并同类项,不同的就单独拎出来讲

\subsubsection{Random external forces}

Random external forces aim to simulate uncertainties in the environment, such as object collisions, wind forces, and so on. Applying random external forces to humanoid robots can better enhance their balance capabilities and resistance to disturbances, thus preventing overfitting of training strategies in stable environments. 

Most mainstream robot simulation platforms support the simulation of random external forces, though their implementations vary. MuJoCo, PyBullet, SAPIEN, CoppeliaSim, and Isaac Gym primarily rely on user-defined methods through scripting or APIs to apply such perturbations. Notably, Isaac Gym offers parallel control interfaces, enabling efficient addition of random disturbances in large-scale scenarios. In contrast,Webots, Gazebo, Isaac Sim, and Isaac Lab provide well-developed interfaces that allow direct application of random forces or disturbances via random velocities. 

\subsubsection{Deformable objects}

Deformable objects refer to materials that undergo shape changes under external forces and are widely used to simulate the physical behavior of flexible materials such as cloth, ropes, and soft robots.
Support for deformable object simulation across mainstream robotic simulators ranges from basic to high fidelity. Platforms like MuJoCo and PyBullet provide foundational soft body capabilities for simple deformable entities such as cloth or elastic materials. In contrast, Isaac Gym, Isaac Sim, and Isaac Lab offer more advanced solutions.
They leverage GPU acceleration or PhysX-based\cite{nvidia_physx_9_21_0713} finite element methods for finer control and higher realism. Genesis further extends these capabilities by integrating state-of-the-art physics solvers, enabling detailed and high-precision simulation of complex deformable materials.
% When simulating deformable objects, simulators primarily model the nonlinear deformation of materials such as fabrics, rubber, and biological tissues. Humanoid robots with deformable object manipulation capabilities are expected to be applied in medical scenarios such as surgeries, as well as in household scenarios like clothing folding.
% 仿真器在对可形变物体进行仿真时,主要模拟布料、橡胶、生物软组织等材料的非线性形变。具备可形变物体操作能力的人形机器人,预期能够应用于医疗手术等医用场景以及衣物折叠等家用场景。
% 在多个机器人仿真平台中,对可形变物体的模拟支持程度各不相同。Isaac Gym主要面向刚体动力学模拟,其对可形变物体的支持有限,例如不支持软体物体、流体或可形变物体的模拟。Isaac Sim和Isaac Lab提供了对可形变物体的支持,利用NVIDIA PhysX的 GPU 加速有限元方法（FEM）进行软体物体的模拟。Isaac Lab 允许用户设置节点位置、速度以及部分节点的运动目标,实现对软体物体的控制。MuJoCo从3.0版本开始引入可变形物体支持,通过新的模型元素flex实现,具体而言,flex是一组由质量点和无质量的可伸缩元素（如胶囊、三角形或四面体）组成的结构,可以模拟可变形几何元素之间的碰撞。

\subsubsection{Soft-body contacts}

Unlike deformable objects, Soft-body contacts refer to the simulation of interactions between soft materials—such as cloth, rubber and other objects. These simulations involve complex deformation responses and force transmission processes, and are widely used in the study of manipulation, collision, and contact mechanics of deformable materials.

Current mainstream simulators support soft-body contact simulation at two main levels: basic support and high-precision modeling. Webots, Gazebo, MuJoCo, CoppeliaSim, and PyBullet offer basic simulation capabilities through soft-contact parameters, simplified contact models, or articulated body structures, which are suitable for general-purpose applications. In contrast, Isaac Gym, Isaac Sim, Isaac Lab, and Genesis provide more advanced and accurate simulations using GPU acceleration or methods such as finite element modeling (FEM), making them better suited for complex physical interaction scenarios.
% 与可形变物体不同,仿真器在对柔性物体接触仿真过程中,重点关注模拟柔性材料（如硅胶、泡沫）在受力后的局部形变特性。进行柔性物体接触力仿真,能够进一步拓宽精密易碎品抓取。
% Isaac Sim和Isaac Lab利用NVIDIA PhysX引擎的有限元方法（FEM）来模拟软体物体的变形和接触行为,具体而言,柔性物体通过双四面体网格进行建模,一个用于模拟变形,另一个用于碰撞检测,这使得在与刚体或其他软体物体接触时,能够准确地模拟局部形变和接触力分布。此外,Isaac Sim的Contact Sensor扩展允许用户在物体表面定义接触区域,并获取接触力和接触点等详细信息,进一步增强了对柔性材料接触行为的模拟能力。MuJoCo基于模型元素flex,构建了接触模型将接触力根据接触点的权重分配到相关的质量点上,从而实现对柔性材料接触行为的高保真模拟。

\subsubsection{Fluid mechanism}

Fluid mechanism simulation refers to the computational modeling of the motion and interaction of fluids such as liquids and gases. 
% It is based on physical laws governing the conservation of mass, momentum, and energy, and typically involves solving the Navier–Stokes equations using numerical methods. 
This type of simulation is widely used in engineering and robotics, enabling researchers and engineers to predict and analyze fluid behavior under various conditions, optimize system designs, improve efficiency, and reduce costs.

Mainstream simulators differ significantly in their support for fluid mechanisms. Webots and Gazebo offer basic fluid simulation capabilities, suitable for modeling simple interactions such as buoyancy and drag, though with limited accuracy. Isaac Sim, built on Omniverse and PhysX, supports more complex fluid behaviors through particle-based methods. Genesis integrates advanced physical solvers, offering native support for high-fidelity fluid simulation and ranking among the most comprehensive and accurate platforms currently available. Other mainstream simulators currently lack native support for fluid mechanisms, though limited community-driven extensions exist—for example, in Isaac Lab.
% Common numerical methods include the finite volume method (FVM), finite difference method (FDM), and finite element method (FEM).
% 流体仿真主要是指模拟液体流动、气体扩散等连续介质力学行为,核心在于数值求解Navier–Stokes方程。这些方程描述了流体的质量、动量和能量守恒,是流体力学的基础。然而,由于其非线性特性和湍流的复杂性,三维不可压缩流动的精确解仍是数学界的开放问题之一。因此,现代流体仿真器通常采用数值方法近似求解Navier–Stokes方程,以实现高效且稳定的仿真。常见的数值方法包括有限体积法（FVM）、有限差分法（FDM）、有限元法（FEM）等。这些方法通过将连续的流体域离散化为有限的计算单元,结合适当的边界条件和求解算法,能够在保证计算效率的同时,提供高质量的流体仿真结果。
% Isaac Sim利用NVIDIA PhysX引擎的粒子系统扩展,采用离散粒子方法（如SPH或FEM）来模拟流体行为。这些粒子系统允许用户创建水流、喷泉等流体效果,并支持与刚体的双向耦合交互。Isaac Lab作为基于Isaac Sim的框架,继承了其流体仿真能力,但目前尚未提供专门的API来简化流体模拟的集成。MuJoCo则提供了两个现象学模型,分别用于模拟飞行和游泳等行为,这些模型是无状态的,能够捕捉刚体在流体介质中运动的主要特征,但不涉及对流体动力学的详细模拟。

\begin{table*}[]
    \centering
    \begin{tabular}{l|cccc}
    \rowcolor{gray!10}
    Simulator & Rendering Engine & Ray Tracing & Physically-Based Rendering & Scalable Parallel Rendering\\
    \hline
    Webots & WREN (OpenGL-based) & \textcolor{red}{\ding{55}} & \textcolor[RGB]{0,100,0}{\checkmark} & \textcolor{red}{\ding{55}}\\
    \rowcolor{gray!10}
    Gazebo & Ogre (OpenGL-based) & \textcolor{red}{\ding{55}} & \textcolor[RGB]{0,100,0}{\checkmark} & \textcolor{red}{\ding{55}}\\
    Mujoco & OpenGL-based & \textcolor{red}{\ding{55}} & \textcolor{red}{\ding{55}} & \textcolor{red}{\ding{55}} \\
    \rowcolor{gray!10}
    CoppeliaSim & OpenGL-based & \textcolor{red}{\ding{55}} & \textcolor{red}{\ding{55}} & \textcolor{red}{\ding{55}}\\
    PyBullet & \makecell{OpenGL-based (GPU)\\TinyRender (CPU)} & \textcolor{red}{\ding{55}} & \textcolor{red}{\ding{55}} & \textcolor{red}{\ding{55}}\\
    \rowcolor{gray!10}
    Isaac Gym & Vulkan-based & \textcolor{red}{\ding{55}} & \textcolor{red}{\ding{55}} & \textcolor[RGB]{0,100,0}{\checkmark}\\
    Isaac Sim & Omniverse RTX Renderer  & \textcolor[RGB]{0,100,0}{\checkmark} & \textcolor[RGB]{0,100,0}{\checkmark} & \textcolor[RGB]{0,100,0}{\checkmark}\\
    \rowcolor{gray!10}
    Isaac Lab & Omniverse RTX Renderer  & \textcolor[RGB]{0,100,0}{\checkmark} & \textcolor[RGB]{0,100,0}{\checkmark} & \textcolor[RGB]{0,100,0}{\checkmark}\\
    SAPIEN &\makecell{SapienRenderer \\ (Vulkan-based)} & \textcolor[RGB]{0,100,0}{\checkmark} & \textcolor[RGB]{0,100,0}{\checkmark} & \textcolor[RGB]{0,100,0}{\checkmark}\\
    \rowcolor{gray!10}
    Genesis & PyRender+LuisaRender & \textcolor[RGB]{0,100,0}{\checkmark} & \textcolor[RGB]{0,100,0}{\checkmark} & \textcolor[RGB]{0,100,0}{\checkmark}\\
    \bottomrule
    
    \end{tabular}
    \caption{Rendering Features Across Various Simulators. \textcolor[RGB]{0,100,0}{\checkmark}: support, \textcolor{red}{\ding{55}}: lack of support.}
    \label{tab:simulator_rendering}
\end{table*}

\subsubsection{DEM(Discrete Element Method) simulation}

The Discrete Element Method (DEM) is a numerical simulation technique that models objects as assemblies of rigid particles, simulating interactions such as contact, collision, and friction between particles. It is widely used to simulate the physical behavior of granular materials and powders. DEM accurately captures the microscopic mechanical properties of particle systems but involves high computational cost, often mitigated by parallel computing or GPU acceleration. In robotic simulations, DEM can be employed to model interactions between robots and granular materials, such as grasping and manipulation.

However, current mainstream simulators do not natively support DEM. Although platforms such as MuJoCo and the Isaac series can simulate granular materials, their contact models are not specifically optimized for the microscopic interactions of particulate matter. Notably, Gazebo supports functionality extension through plugins, and NASA’s OceanWATERS project ~\cite{catanoso2020analysis} integrates DEM simulation with Gazebo to enable indirect simulation of granular material behavior. This approach primarily supports force feedback in tasks like excavation rather than real-time particle-level simulation.
% 离散元仿真主要针对颗粒物质（如砂石、粉末）的离散化个体运动与群体碰撞仿真。其主要应用于土木与岩土工程、矿业与冶金、化工与制药、能源与石油工程以及材料科学等领域。随着计算能力的提升,DEM技术也被应用于海洋工程、环境科学等新兴领域,助力复杂系统的建模与分析。

\subsubsection{Differentiable physics}

In the field of simulation, differentiable physics refers to a simulator’s ability to compute gradients of physical states with respect to input parameters, such as control signals, object poses, and physical properties. This capability enables end-to-end optimization and learning, allowing seamless integration with machine learning models, particularly reinforcement learning and optimization algorithms, to achieve efficient self-learning and task performance improvement.

Several simulation platforms have recently made rapid advancements in supporting differentiable physics. The XLA branch of MuJoCo (MuJoCo XLA) enables differentiable simulation via JAX, allowing gradient computation and optimization tasks. PyBullet offers a differentiable interface through its subproject, the Tiny Differentiable Simulator, which is suitable for gradient-based learning and optimization. Genesis incorporates differentiability from the ground up, having already implemented differentiable physics in its MPM solver and planning to extend it to rigid body and articulated system solvers. Overall, these platforms are increasingly integrating high-fidelity physical modeling with automatic differentiation, accelerating the deployment of differentiable simulation in embodied intelligence and robot learning.
% 物理可微的特性是指在对物理量进行仿真的过程中,构建物理量的梯度传递链路,使其支持基于反向传播的联合优化,从而能够在网络训练的过程中,直接学习物理量的分布,而避免使用强化学习的方法间接获得环境的信息。
% Isaac Sim构建于NVIDIA Omniverse平台之上,支持高保真渲染和复杂环境模拟,包括软体物体、流体、可形变物体等,其渲染能力和物理模拟功能使其适用于复杂的机器人仿真任务。Isaac Lab作为基于Isaac Sim的机器人学习框架,继承了其对物理可微的支持能力,它提供了高层次的API,方便用户进行机器人学习任务的开发和测试。值得一提的是,新推出的开源物理引擎NVIDIA Newton专注于高性能的物理仿真,也宣称其对于物理可微提供良好的支持。

\begin{table*}[]
	\centering
	\begin{tabular}{l|ccc|cc}
         \centering
        \tabularnewline \rowcolor{gray!10}
		 & \multicolumn{3}{c|}{Sensor} & \multicolumn{2}{c}{Joint type} \\ 
        \rowcolor{gray!10}
		Simulator& \makecell{IMU/Force contact/\\RGB Camera}&LiDAR& GPS & \makecell{Floating/Fixed/Hinge\\Spherical/Prismatic}  & Helical \\ 
  \hline
		Webots &  \textcolor[RGB]{0,100,0}{\checkmark}  &  \textcolor[RGB]{0,100,0}{\checkmark}  & \textcolor[RGB]{0,100,0}{\checkmark} & \textcolor[RGB]{0,100,0}{\checkmark} & \textcolor{red}{\ding{55}} \\
        \rowcolor{gray!10}
        Gazebo &   \textcolor[RGB]{0,100,0}{\checkmark}  &  \textcolor[RGB]{0,100,0}{\checkmark}  & \textcolor[RGB]{0,100,0}{\checkmark} & \textcolor[RGB]{0,100,0}{\checkmark} & \textcolor[RGB]{0,100,0}{\checkmark} \\
        Mujoco &   \textcolor[RGB]{0,100,0}{\checkmark}  &  \textcolor[RGB]{0,100,0}{\checkmark}  & \textcolor{red}{\ding{55}} & \textcolor[RGB]{0,100,0}{\checkmark} & \textcolor{red}{\ding{55}} \\
        \rowcolor{gray!10}
        CoppeliaSim &   \textcolor[RGB]{0,100,0}{\checkmark}  &  \textcolor[RGB]{0,100,0}{\checkmark}  & \textcolor[RGB]{0,100,0}{\checkmark} & \textcolor[RGB]{0,100,0}{\checkmark} & \textcolor[RGB]{0,100,0}{\checkmark} \\
        PyBullet &   \textcolor[RGB]{0,100,0}{\checkmark}  &  \textcolor[RGB]{0,100,0}{\checkmark}  & \textcolor{red}{\ding{55}} & \textcolor[RGB]{0,100,0}{\checkmark} & \textcolor{red}{\ding{55}} \\
        \rowcolor{gray!10}
        Isaac Gym &   \textcolor[RGB]{0,100,0}{\checkmark}  &  \textcolor{red}{\ding{55}}  & \textcolor[RGB]{0,100,0}{\checkmark} & \textcolor[RGB]{0,100,0}{\checkmark} & \textcolor{red}{\ding{55}} \\
        Isaac Sim & \textcolor[RGB]{0,100,0}{\checkmark}  &  \textcolor[RGB]{0,100,0}{\checkmark}  & \textcolor[RGB]{0,100,0}{\checkmark} & \textcolor[RGB]{0,100,0}{\checkmark} & \textcolor{red}{\ding{55}} \\
        \rowcolor{gray!10}
        Isaac Lab &   \textcolor[RGB]{0,100,0}{\checkmark}  &  \textcolor[RGB]{0,100,0}{\checkmark}  & \textcolor[RGB]{0,100,0}{\checkmark} & \textcolor[RGB]{0,100,0}{\checkmark} & \textcolor{red}{\ding{55}} \\
        SAPIEN &   \textcolor[RGB]{0,100,0}{\checkmark}  &  \textcolor{red}{\ding{55}}  & \textcolor{red}{\ding{55}} & \textcolor[RGB]{0,100,0}{\checkmark} & \textcolor{red}{\ding{55}} \\
        \rowcolor{gray!10}
        Genesis &   \textcolor[RGB]{0,100,0}{\checkmark}  &  \textcolor[RGB]{0,100,0}{\checkmark}  & \textcolor[RGB]{0,100,0}{\checkmark} & \textcolor[RGB]{0,100,0}{\checkmark} & \textcolor{red}{\ding{55}} \\
        \bottomrule
	\end{tabular}
	\caption{Sensor and Joint Component Support Across Various Simulators. \textcolor[RGB]{0,100,0}{\checkmark}: support, \textcolor{red}{\ding{55}}: lack of support.}
	\label{tab:sensorandjoint}
\end{table*}

\subsection{Rendering Capabilities}
~\label{sec:rendering_capabilities}

Simulation rendering capabilities are pivotal in modern robotics research and development. They not only provide researchers with an efficient virtual experimentation environment but also ensure that the appearance and behavior of the robot are more realistic and accurate in predictions. High-fidelity rendering plays a crucial role in diminishing the simulation-to-reality (sim-to-real) gap. This is essential for various stages of robot development, including design, validation, and optimization, and it significantly bolsters the robustness and reliability of perception, control, and Simultaneous Localization and Mapping (SLAM) algorithms during their transition to real-world deployment.

This subsection presents a comparative analysis of the rendering capabilities of the aforementioned simulators, focusing on four key technological aspects: the underlying rendering engines, support for ray tracing, implementation of Physically Based Rendering (PBR), and capabilities for parallel rendering. The results are shown in Table ~\ref{tab:simulator_rendering}.

\subsubsection{Rendering Engine}

A rendering engine is the core software used to create 2D images from 3D scene descriptions. Its tasks include processing geometric data, applying textures, calculating illumination from various light sources, and executing shading models to determine the final appearance of surfaces.

OpenGL\cite{opengl}, a long-standing cross-platform graphics API, is widely adopted.
Webots employs its proprietary Webots Rendering Engine (WREN), architected upon OpenGL 3.3. WREN is specifically optimized for GPU hardware and tailored to the Webots simulation platform, incorporating features such as anti-aliasing, resolution enhancement, ambient occlusion, and a full PBR pipeline.
MuJoCo features an integrated renderer implemented using fixed-function OpenGL, prioritizing efficiency by preloading GPU resources but limited to basic visual effects.
CoppeliaSim leverages a built-in OpenGL-based rendering engine for real-time rasterization.  
PyBullet offers basic rendering via OpenGL on the GPU and includes TinyRender, a CPU-based software renderer, for headless or non-GPU scenarios.

As OpenGL's capabilities have gradually fallen short of real-world demands, subsequent rendering engines have adopted more versatile graphics APIs to better leverage GPU resources and achieve higher rendering performance. Isaac Gym utilizes Vulkan\cite{vulkan} for its lightweight GUI visualization. This Vulkan-powered viewer does not aim for high-fidelity rendering but provides efficient, real-time visual feedback for debugging and simulation workflows. The subsequent NVIDIA Isaac suite presents a more advanced approach. Isaac Sim and Isaac Lab are deeply integrated with the NVIDIA Omniverse ecosystem. They employ the Omniverse RTX renderer with the Hydra rendering delegate\cite{HydraRenderer}, leveraging NVIDIA's RTX technology for high-fidelity graphics. SAPIEN employs a custom, high-performance rendering engine built upon the Vulkan graphics API, known as SapienRenderer. Genesis utilizes a native rendering pipeline, integrating cutting-edge frameworks such as PyRender\cite{pyrender} and LuisaRender\cite{LuisaRender}.

\subsubsection{Ray Tracing}

Ray tracing is a rendering technique that simulates the physical behavior of light by tracing paths of light rays as they interact with surfaces in a 3D scene. Its primary advantage over traditional rasterization in robotics is the ability to produce highly accurate shadows, reflections, refractions, global illumination, and, critically, more physically realistic simulation of sensors like LiDAR and depth cameras. Rasterization often struggles with non-linear optical sensors and complex light interactions.

Among the reviewed simulators, Webots, MuJoCo, and PyBullet do not offer native real-time ray tracing, relying on rasterization for their primary visual output. CoppeliaSim integrates the POV-Ray tracer, enabling high-quality static image generation with ray tracing, but it does not support real-time ray tracing for dynamic simulations. Rendering in Isaac Gym is also relatively basic, and does not support either ray tracing or the more sophisticated synthetic data sensors provided in Omniverse. 
In contrast, simulators built for high-fidelity visual output increasingly incorporate ray tracing. Isaac Sim and Isaac Lab, through the Omniverse RTX renderer, provide robust real-time ray tracing, enabling effects like global illumination, reflections, and refractions. SAPIEN also offers significant ray tracing support; its SapienRenderer (Vulkan-based) supports both rasterization and ray tracing pipelines, selectable via shader packs. Gazebo, while primarily rasterization-based, shows a path towards ray tracing through experimental NVIDIA OptiX\cite{nvidia_optix} support in its gz-rendering library. Genesis, aiming for photorealism, utilizes a state-of-the-art high-performance renderer ``LuisaRender".

\subsubsection{Physically-based Rendering}

PBR is an approach that models how light interacts with materials based on their physical properties, such as roughness and metallicity. This results in more realistic and consistent visuals across different lighting conditions, leading to better material definition and improved visual fidelity, which is crucial for training vision-based robotic learning methods.

Webots' WREN engine implements a PBR pipeline, incorporating advanced lighting models and non-color textures for realism comparable to modern game engines.
Modern Gazebo, through Ignition Rendering, also supports PBR, and its Ogre backend has PBR capabilities.
In contrast, MuJoCo, CoppeliaSim and PyBullet, with their focus on basic visual effects and rudimentary rendering respectively, lack PBR support.
Isaac Gym, using a basic Vulkan-based pipeline, also does not implement PBR. 

High-fidelity simulators embrace PBR. Isaac Sim and Isaac Lab, built on NVIDIA Omniverse, inherently support PBR through the Omniverse RTX renderer and leverage the Material Definition Language (MDL) for defining realistic, physically-based materials. SAPIEN and Genesis also support PBR. Their advanced rendering capabilities, including ray tracing and accurate material property representation, are integral to PBR systems. 

\subsubsection{Parallel Rendering}

Parallel rendering in robotics simulation encompasses rendering multiple independent simulation environments simultaneously for large-scale RL or data collection. Its utility lies in significantly speeding up RL agent training, enabling efficient generation of large synthetic datasets.

Simulators like Isaac Gym, Isaac Sim/Lab, SAPIEN (especially with ManiSkill), and Genesis are designed with strong parallel rendering capabilities as a core architectural feature. Isaac Gym can simulate and render thousands of environments in parallel on a single GPU, including camera sensors. Isaac Sim and Isaac Lab extend this to multi-GPU training, where each process can run on a dedicated GPU with its own Isaac Sim instance, and Isaac Sim itself supports multi-GPU rendering for complex scenes or multiple cameras.
SAPIEN, particularly with ManiSkill3, offers a GPU-parallelized visual data collection system achieving very high FPS (e.g., 30,000+ FPS with rendering on a high-end GPU) and supports heterogeneous simulation where each parallel environment can differ.

Older or more general-purpose simulators like Webots, Gazebo, MuJoCo, CoppeliaSim, and PyBullet may support running multiple instances or have some form of physics parallelization but generally lack the integrated, high-throughput parallel visual rendering pipelines of the specialized platforms. Webots allows multiple instances (e.g., via Docker) and disabling rendering for speed, but parallel visual output is not a primary strength. Gazebo parallelizes its physics engine but is not designed for massive parallel visual output for RL. MuJoCo can run parallel simulation threads with basic or offscreen rendering per instance. CoppeliaSim allows multiple scenes but renders only the active one, not suited for parallel data collection. PyBullet supports multiple physics servers and headless image capture via a CPU renderer, with some multi-threading in Bullet physics, but lacks high-throughput GPU parallel visual rendering.

\begin{figure}
    \centering
    \includegraphics[width=\linewidth]{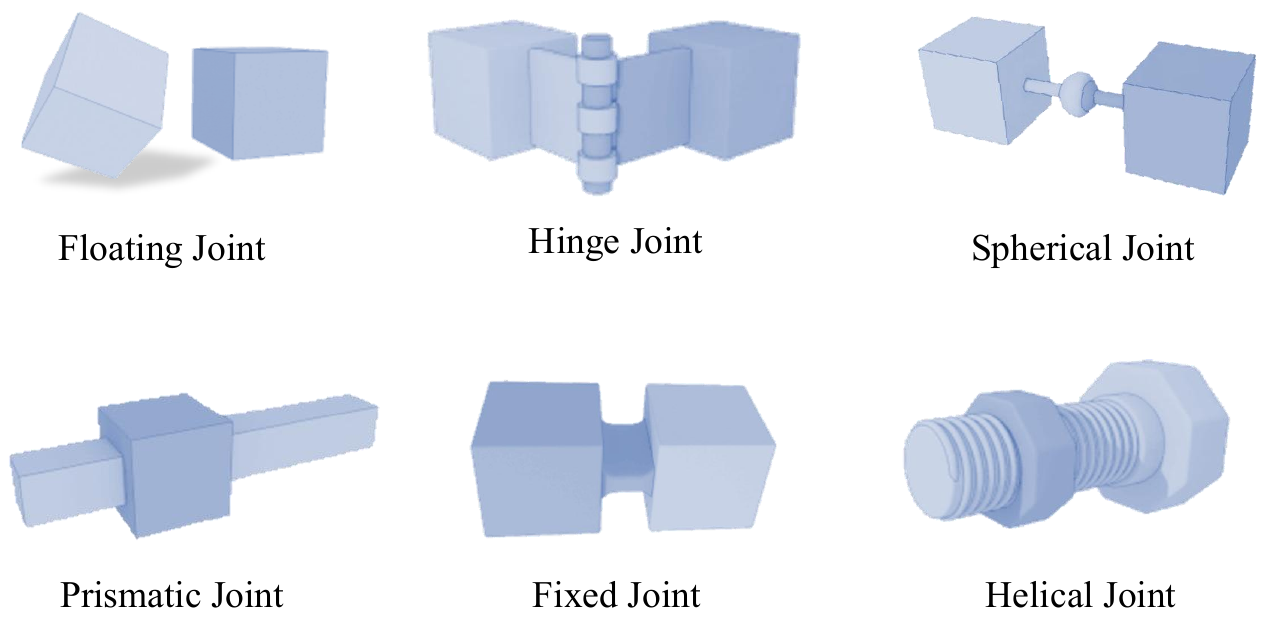}
    \caption{Main joint types in simulators.\cite{ls_group_kinematics}}
    \label{fig:joint_types}
\end{figure}

\subsection{Sensor and Joint Component Types}
~\label{sec:sensor_and_joint}
% 传感器支持
Sensors are essential components in robot perception, providing the foundation for motion control and environmental understanding by capturing multimodal information such as vision, force, and inertia. In simulation platforms, realistic and reliable sensor models offer an efficient, safe, and repeatable testing basis for debugging, validating, and collecting data for robot perception algorithms. We surveyed the support for various sensors across mainstream simulation platforms. Specifically, the sensors considered include inertial measurement unit (IMU), force contact, RGB camera, LiDAR, and GPS.

As shown in Table~\ref{tab:sensorandjoint}, most mainstream simulation platforms provide standard support for commonly used sensors, including vision (RGB), IMU, and contact force sensing. 
For example, Isaac Sim and Isaac Lab offer comprehensive high-fidelity sensor simulation. Isaac Gym supports IMU state and contact information output, but its vision capabilities are relatively limited in terms of runtime efficiency and parallel processing. In practice, it is often used in conjunction with Isaac Sim or Omniverse to enable high-quality visual sensing and processing. Genesis provides comprehensive multimodal sensor support, making it suitable for complex perception tasks. 
Beyond these, support for other sensors varies across platforms. Specifically, Isaac Gym and SAPIEN do not provide native support for LiDAR sensors, while MuJoCo, PyBullet, and SAPIEN lack support for GPS.
% current robot simulation platforms typically simulate a range of sensors, including vision(RGB), LiDAR, IMU(inertial measurement unit), and contact force, using techniques such as ray casting, physical feedback, and noise modeling. These simulations not only replicate the physical characteristics of sensors but also introduce real-world noise and delays, enabling perception algorithms to receive inputs closely resembling real-world data, thus improving algorithm generalization and deployment efficiency.
% 传感器是机器人感知外部世界的重要组件,通过采集视觉、力觉、惯性等多模态信息为运动控制和环境理解提供基础；在仿真平台中,真实可靠的传感器模型为机器人感知算法的调试、验证与数据采集提供了高效、安全且可重复的测试基础。
% 本文将各仿真平台对于不同传感器的支持程度总结在表\ref{tab:sensorandjoint}中,当前的机器人仿真平台通常通过射线投射、物理反馈和噪声建模等手段,模拟视觉、激光雷达、惯性测量和接触力等多种传感器。这些模拟不仅再现了传感器的物理特性,还引入了真实世界中的噪声和延迟,使得感知算法在虚拟环境中能获得接近实物的输入,从而提升了算法的泛化能力和部署效率。

Besides sensor modeling, accurately simulating various joint types (such as floating, hinge, spherical, prismatic, fixed and helical) is crucial for reproducing the robot's structure and motion characteristics in simulation platforms. The combination of these joints determines the robot's degrees of freedom (DOF) and flexibility, which in turn affects its performance in complex tasks. By precisely modeling these joints in the simulation, developers can test and optimize robot designs and control strategies in a virtual environment, thereby improving system performance and reliability. 

As shown in Fig.~\ref{fig:joint_types}, the simulation of joint components in these platforms is typically based on motion constraint mechanisms provided by physics engines, with the DOF between two rigid bodies defined to achieve the joint's movement. 
Floating joints enable free movement and rotation in all directions. Fixed joints completely constrain the relative motion of two rigid bodies, hinge (revolute) joints allow rotation around a specific axis,  prismatic joints allow linear motion along a specific axis, and spherical joints provide three rotational degrees of freedom. Simulation platforms often allow users to set parameters such as range limits, damping, and spring stiffness for joints to simulate the physical characteristics of actual mechanical systems.
Some platforms also support more complex joint types, such as helical joints (which enable coupled rotational and translational motion). A helical joint allows two connected rigid bodies to perform synchronized rotation and linear movement along the same axis, similar to the advancement of a screw within a nut, with a single DOF and a helical motion trajectory.  
Table~\ref{tab:sensorandjoint} also summarizes the support for various joint types across mainstream simulation platforms. Most simulators support common joint types such as floating, fixed, hinge, spherical, and prismatic. However, helical joints are less commonly supported and are natively implemented only in Gazebo and CoppeliaSim.

\subsection{Discussions and Future Perspectives}
~\label{subsec:discussion}
Simulators are computational models designed to replicate real-world processes or systems, and they have become indispensable tools across various scientific and engineering disciplines. Their primary purpose is to provide a controlled environment for experimentation, analysis, and prediction without the need for costly or risky physical trials. Despite their widespread use, simulators possess both significant advantages and notable limitations, which highlight the need for more advanced modeling approaches such as world models.

\subsubsection{Advantages of Simulators}

Simulators offer several key benefits that make them valuable in research and practical applications:

\textbf{Cost-effectiveness}: By reducing or eliminating the need for physical experiments, simulators can significantly lower the financial burden associated with testing and development.

\textbf{Safety}: They enable the simulation of hazardous or dangerous scenarios (e.g., nuclear reactor failures or extreme weather conditions) without posing real-world risks.

\textbf{Control}: Simulators provide precise control over variables and experimental conditions, allowing researchers to isolate specific factors and study their effects.

\textbf{Repeatability}: Experiments conducted in simulators can be repeated exactly, ensuring consistency and facilitating the verification of results.

\subsubsection{Challenges of Simulators}

However, simulators are not without their challenges, and these limitations often constrain their effectiveness:

\textbf{Accuracy}: Simulators may fail to fully capture the complexity of real-world systems, leading to inaccuracies in their predictions. Simplifications or approximations made during model design can result in discrepancies between simulated and actual outcomes.

\textbf{Complexity}: Real-world systems are often highly intricate, with numerous interacting components and variables. Creating a simulator that accurately reflects this complexity is computationally demanding and, in some cases, infeasible.

\textbf{Data dependency}: Effective simulators typically require large amounts of high-quality data to calibrate and validate their models. In scenarios where data is scarce or difficult to obtain, the performance of simulators can be severely compromised.

\textbf{Overfitting}: There is a risk that simulators may become overly tailored to specific scenarios or datasets, reducing their ability to generalize to new or unseen conditions. This limits their applicability in dynamic or evolving environments.

\subsubsection{Future Perspectives}

Limitations in accuracy, complexity, and data dependency highlight the need for more sophisticated, adaptable modeling frameworks.
This has motivated the development of world models, which aim to provide a more comprehensive and flexible approach to understanding and predicting real-world dynamics. Unlike traditional simulators, world models leverage advances in machine learning and artificial intelligence to create representations that can adapt to new data, handle complex systems more efficiently, and reduce reliance on extensive datasets. As such, the study of world models represents a natural evolution in the pursuit of more robust and versatile tools for modeling real-world phenomena.

\section{World Models}
\label{sec:world_models}

\vspace{1mm}
\begin{quote}
    ``World models are \textbf{generative} AI models that understand the \textbf{dynamics} of the real world, including physics and spatial properties.''
    \begin{flushright}
        -- \emph{NVIDIA's World Foundation Models}~\cite{nvidiaWorldModels}
    \end{flushright}
\end{quote}
\vspace{1mm}

\begin{figure}
\centering
\begin{subfigure}[b]{0.48\textwidth}
   \includegraphics[width=1\linewidth]{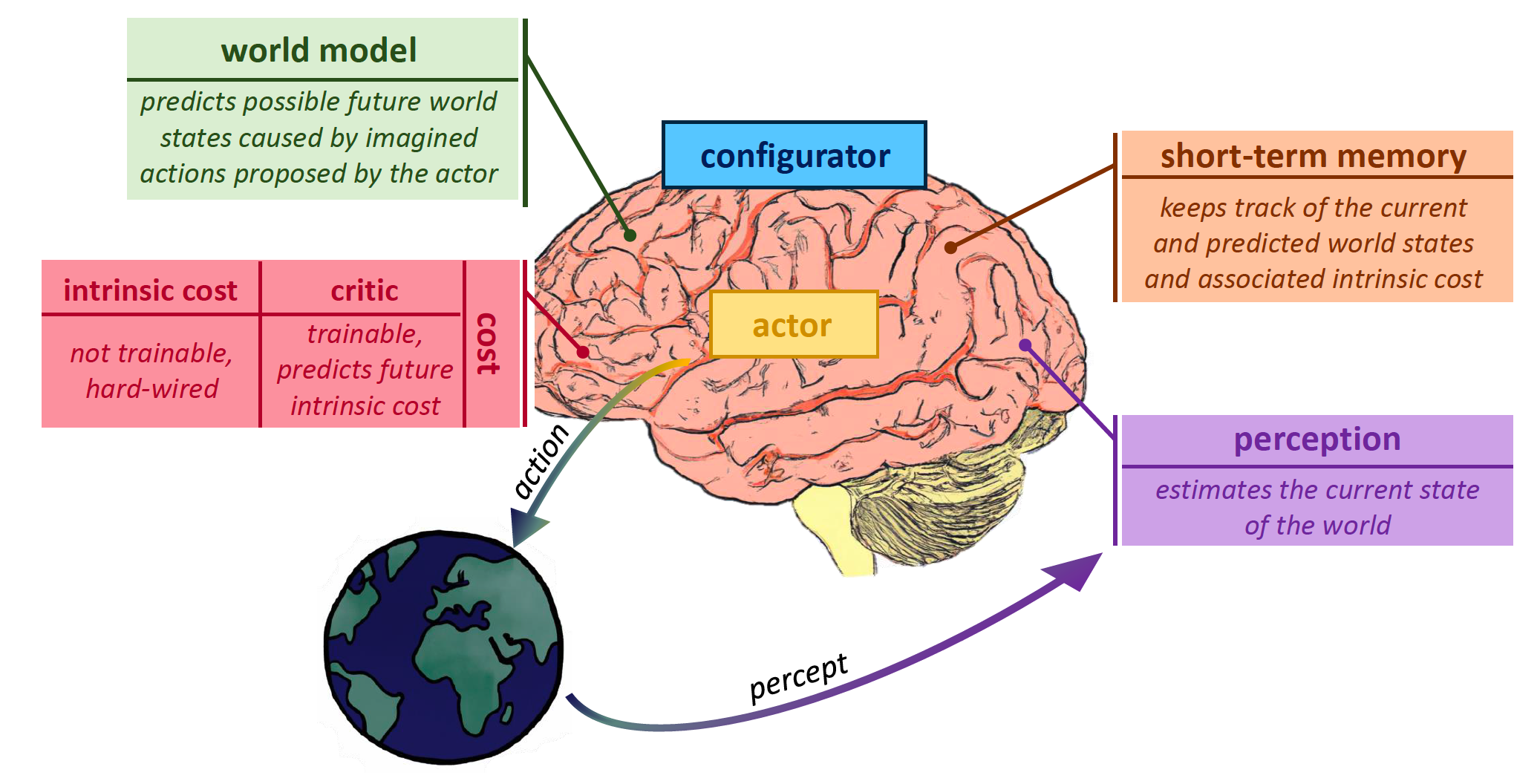}
   \caption{World model in the AI system proposed by Lecun~\cite{dawid2024introduction}.}
   \label{fig:Ng1} 
\end{subfigure}

\begin{subfigure}[b]{0.45\textwidth}
   \includegraphics[width=1\linewidth]{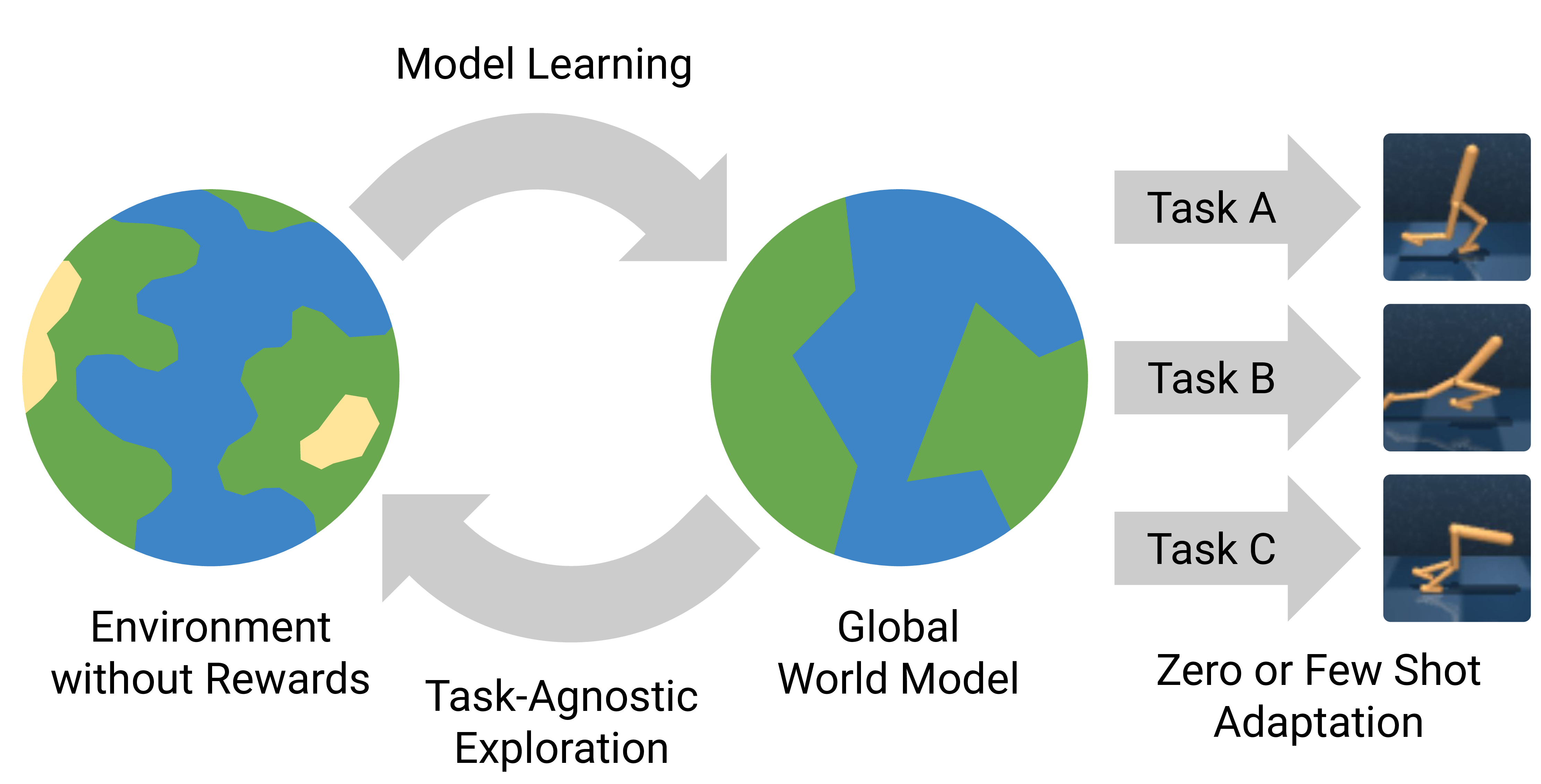}
   \caption{Learn World model in a self-supervised manner~\cite{sekar2020planning}.}
   \label{fig:Ng2}
\end{subfigure}

%\caption{World model }
\caption{Illustration of the role and training of world models in AI systems.}
\end{figure}

%\ZQR{早期world model工作+因为lecun说 人是inner world model，人只接受双目的视频（视频资源也是最多的）+视频世界模型}

\begin{figure*}
    \centering
    \includegraphics[width=\linewidth]{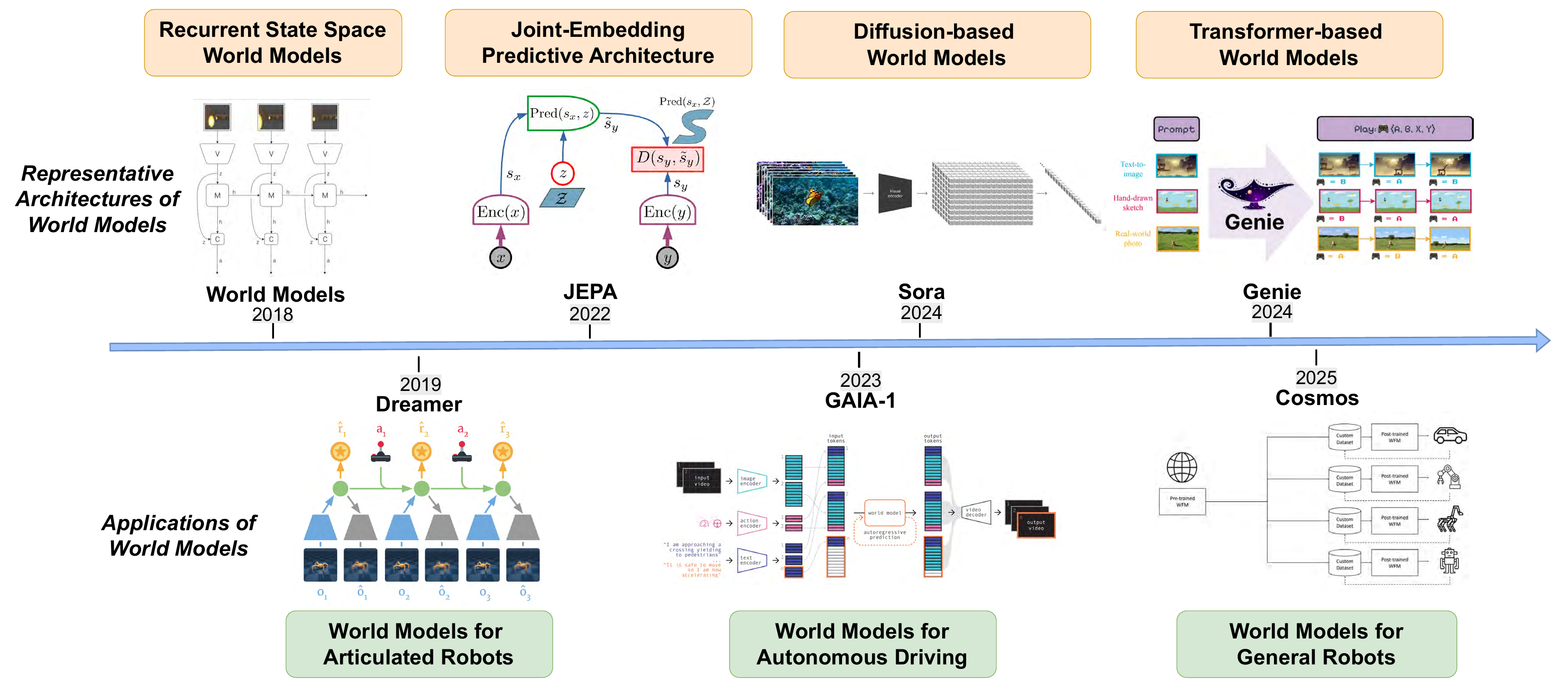}
    \caption{Representative architectures and applications of world models.}
    \label{fig:wm_evolution}
\end{figure*}

%\xiao{一开始，需要直接给出world model的定义}
Inspired by the human brain's ability to form internal representations of the world, world models has emerged as a pivotal framework in artificial intelligence. These models enable agents to predict future states and plan actions, mimicking the cognitive processes that allow humans to navigate and interact with their environment. In 2018, David Ha and Jürgen Schmidhuber introduced the idea of world models\cite{Ha2018}, demonstrating that an AI could learn a compressed, generative model of its environment and use it to simulate experiences, thereby facilitating reinforcement learning without direct interaction with the real world.

As the field progressed, advancements in video generation models have significantly enhanced the capabilities of world models. Since early 2024, video generation models, such as Sora\cite{sora} and Kling\cite{kling} have garnered considerable attention from both academia and industry due to their high-fidelity video synthesis and realistic modeling of the physical world. 
The technical report on Sora, titled "Video Generation Models as World Simulators" \cite{sora}, emphasizes the potential of employing video generation models as robust engines for simulating the physical world. The Navigation World Model (NWM)\cite{bar2024navigation} employed a Conditional Diffusion Transformer (CDiT) to predict future visual observations based on past experiences and navigation actions, which enables agents to plan navigation trajectories by simulating potential paths and evaluating their outcomes.

Yann LeCun also emphasized the importance of video-based world models, stating that humans develop internal models of the world through visual experiences, particularly through binocular vision. He argues that for artificial intelligence to achieve human-level cognition, it must learn in a manner similar to humans, primarily through visual perception. This perspective underscores the significance of integrating video data into world models to capture the richness of spatial and temporal information. He further proposed the Video Joint Embedding Predictive Architecture (V-JEPA) model\cite{drozdov2024video}, which aims to learn abstract representations of videos by predicting missing parts of videos, thus providing new ideas for building a more powerful visual world model. 

Building upon these insights, recent developments in video generation models aim to create more sophisticated world models that can represent and understand dynamic environments. By leveraging large-scale video datasets and advanced neural architectures, these models strive to replicate the way humans perceive and interact with the world, paving the way for more advanced and adaptable artificial intelligence systems.

This section begins by reviewing the architectural evolution of video generation models, tracing their technological advancements to establish a foundation for subsequent discussions on their role as world models. 
It then highlights key developments in the application of generative models as world models across various domains, including their function as controllable data generators and their use in model-based reinforcement learning (model-based RL) for dynamic and reward modeling.

\subsection{Representative Architectures of World Models}
~\label{subsec:architecure_of_WM}
To effectively capture the dynamics of complex environments, world models have evolved into a diverse set of architectural paradigms, each reflecting a different perspective on how to represent and predict the world. 
From early \textit{compact latent dynamics models} to recent powerful \textit{generative architectures}, these models differ in how they encode states, handle temporal dependencies, and model future observations. 
Some approaches prioritize \textit{efficient state abstraction and predictive learning in latent space}, while others focus on \textit{high-fidelity generation of future sensory inputs such as video or 3D scenes}. Meanwhile, advances in sequence modeling, self-supervised learning, and generative modeling, particularly with transformers and diffusion models, have profoundly shaped the design of modern world models. In this section, we review representative architectural families that have been widely adopted or recently proposed, highlighting their core modeling principles.
An overview of the architecture evolution is shown in Fig.~\ref{fig:wm_evolution}.

{\bf Recurrent State Space Model.} Recurrent State Space Models (RSSMs) are among the earliest and most widely adopted architectures for learning world models. The core idea is to use a compact latent space to encode the evolving state of the environment, and to model its temporal dynamics using a recurrent structure. This design enables long-horizon prediction and decision-making by simulating possible futures in the latent space, rather than directly predicting raw observations. The RSSM framework was popularized by the Dreamer series~\cite{hafner2018planet, dreamerv1,dreamerv2,hafner2023dreamerv3,wu2023daydreamer}, which demonstrated that planning and reinforcement learning can be effectively conducted within the learned latent dynamics. The recurrent modeling of latent transitions remains a central mechanism for learning predictive and decision-aware representations in continuous environments.

{\bf Joint-Embedding Predictive Architecture.}
Like RSSMs, Joint-Embedding Predictive Architectures (JEPAs) also model the world in an abstract latent space, but differ in learning objective. Instead of reconstructing visual observations, JEPA models are trained to predict abstract-level representations of missing content in a purely self-supervised manner.
Originally proposed by Yann Lecun as part of a broader framework for autonomous machine intelligence~\cite{lecun2022wm}, JEPA eliminates the need for explicit generative decoders by framing prediction as a representation matching problem. I-JEPA (Image JEPA)~\cite{ijepa} and V-JEPA (Video JEPA)~\cite{vjepa, assran2025vjepa2} instantiate this idea in static and temporal domains respectively. I-JEPA learns to predict masked image regions directly in latent space, while V-JEPA extends this to spatiotemporal structures, enabling representation learning from videos without frame-level reconstruction. This architecture is particularly appealing for large-scale pretraining, where semantic abstraction and data efficiency are prioritized over exact observation synthesis.

{\bf Transformer-based State Space Models.}
As world models scale to more complex environments and longer horizons, the limitations of recurrent architectures in modeling long-range dependencies have become more evident. Transformer-based models address this by replacing RNNs with attention-based sequence modeling, offering greater expressiveness and parallelism. In latent dynamics modeling, this shift is exemplified by works such as TransDreamer~\cite{chen2022transdreamer}, TWM~\cite{TWM}, and Google Deepmind's Genie~\cite{Bruce2024GenieGI}, where Transformers are used to model transitions in latent state spaces over extended temporal windows. These models retain the latent-space prediction philosophy of RSSMs, but benefit from the Transformer’s ability to capture global context across time, improving long-horizon consistency and planning performance.

\begin{figure*}[ht!]
    \vspace{-5mm}
    \centering
    \includegraphics[width=\linewidth]{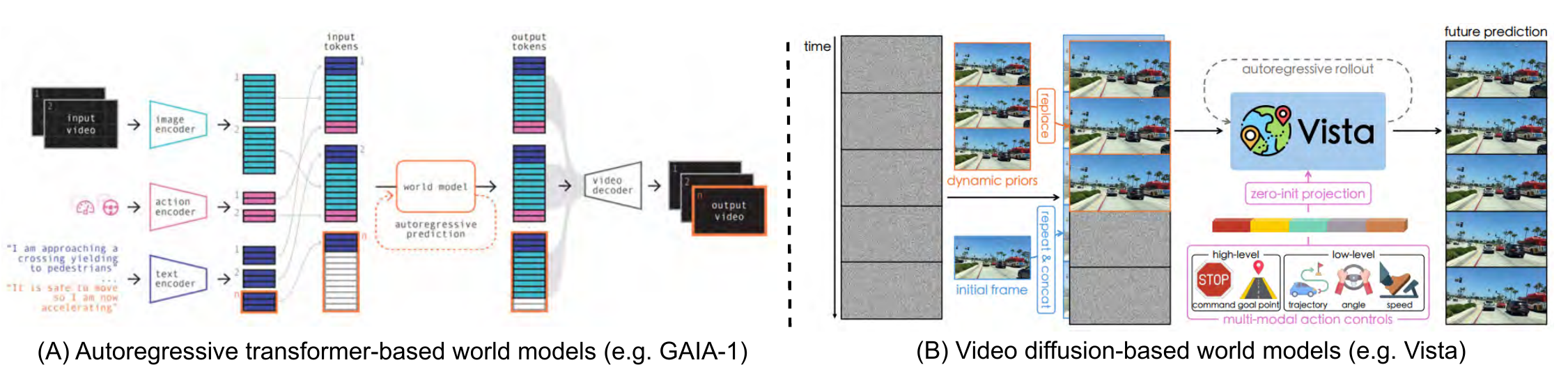}
    \caption{Comparison of autoregressive transformer-based world models and video diffusion-based world models. }
    \label{fig:wm_cmp}
\end{figure*}

{\bf Autoregressive Generative World Models.}
Inspired by recent breakthrough in natural language processing, autoregressive generative models treat world modeling as a sequential prediction task over tokenized visual observations. These models leverage Transformer architectures to generate future observations conditioned on past context, often integrating multimodal inputs such as actions or language. Early video generation frameworks like CogVideo~\cite{cogvideo}, NUWA~\cite{nuwa}, and VideoPoet~\cite{kondratyuk2023videopoet} laid the groundwork by demonstrating scalable autoregressive synthesis of realistic videos. Building on this, recent generative world models for autonomous driving and 3D scene generation, such as GAIA-1~\cite{hu2023gaia}, OccWorld~\cite{zheng2024occworld}, incorporate autoregressive Transformers to model complex environment dynamics with multimodal inputs and outputs. 
However, discrete token quantization frequently results in the loss of high-frequency details, adversely affecting the visual quality of generated videos.

{\bf Diffusion-based Generative World Models.}
In recent years, diffusion models have rapidly become a cornerstone of video generation, offering stable training and superior fidelity in synthesizing temporally consistent visual sequences. By iteratively denoising from noise, these models capture complex data distributions with high expressiveness. To reduce computational cost, recent works shift from pixel-space diffusion (e.g., VDM~\cite{vdm}) to latent-space modeling using pretrained autoencoders, enabling efficient cascaded generation of long and high-resolution videos (e.g., Imagen Video~\cite{imagenvideo}, VideoLDM~\cite{videoldm}, SVD~\cite{svd}). With the emergence of large-scale pretrained models such as OpenAI's Sora~\cite{sora} and Google Deepmind's Veo3~\cite{veo3}, diffusion has demonstrated not only visual realism but also the ability to model 3D structure and physical dynamics—qualities aligning closely with the goals of world models.

As a result, diffusion-based architectures are now increasingly adopted in generative world models, where the goal is to simulate future observations grounded in realistic and controllable environments. Models like DriveDreamer~\cite{DriveDreamer2023}, Vista~\cite{Gao2024VistaAG} and GAIA-2~\cite{russell2025gaia2} apply diffusion model to generate video or 3D scenes conditioned on actions or other modalities. Compared to autoregressive models, diffusion models offer stronger spatiotemporal coherence and higher visual quality, making them a compelling foundation for building high-fidelity, predictive neural simulators of the world. 

Despite substantial progress in video diffusion-based world models, challenges persist in sampling speed, long-duration generation, and the ability to model causal temporal dynamics. Consequently, emerging research such as Vid2World~\cite{huang2025vid2world} and Epona~\cite{zhang2025epona} is investigating the integration of diffusion models’ visual expressiveness with the temporal modeling strengths of autoregressive models, aiming to achieve realistic and interactive video-based world modeling.

\subsection{Core Roles of World Models} \label{sec:wm_role}
Beyond architectural designs, it is equally important to understand the roles world models play in intelligent systems. As general-purpose representations of the environment, world models serve as crucial enablers across various domains. Their capacity to abstract and predict environmental dynamics allows them to support downstream applications far beyond simple reconstruction tasks. In this section, we highlight three core roles that world models fulfill: 1) as \textit{neural simulators} that generate controllable, high-fidelity synthetic experiences. 2) as \textit{dynamic models} that enable planning and decision-making in model-based reinforcement learning; and 3) as \textit{reward models} that help extract meaningful training signals in the absence of dense or well-defined rewards. These perspectives not only reflect the practical utility of world models but also guide future research directions in leveraging world modeling for intelligent agents.

% \vspace{1mm}
% \begin{quote}
%     ``Our results suggest that scaling video generation models is a promising path
%     towards building general purpose \textbf{simulators} of the physical \textbf{world}.''
%     \begin{flushright}
%         -- \emph{\textbf{OpenAI}'s Sora Technical Report}~\cite{sora}
%     \end{flushright}
% \end{quote}
% \vspace{1mm}

\subsubsection{World Models as Neural Simulator}

The rise of generative world models has unlocked the potential to simulate complex, controllable environments across vision and action domains. These models synthesize temporally coherent and semantically grounded videos conditioned on diverse inputs, such as text, images, and trajectories, enabling scalable training and evaluation for autonomous driving, robotics, and virtual agents.

{\bf NVIDIA’s Cosmos series} exemplifies this direction. Cosmos~\cite{agarwal2025cosmos} proposes a unified platform for building foundation video models as general-purpose world simulators, adaptable via fine-tuning for domains such as robotics and autonomous driving. It offers a full-stack pipeline: from video filtering and tokenization to pretraining and downstream adaptation, designed for efficient, open-source world model development.

Extending this foundation, Cosmos-Transfer1~\cite{Cosmos-Transfer1_2024} introduces a spatially conditioned, multi-modal video generator. Through adaptive fusion control, it incorporates structured inputs such as segmentation maps, depth, and edges to guide video synthesis. Its spatial weighting mechanism enhances generation controllability and diversity, supporting applications like sim-to-real transfer, data augmentation, and robot perception with fine-grained control.

Beyond general-purpose models, domain-specific simulators like {\bf Wayve’s GAIA series} focus on realistic and controllable traffic simulation. GAIA-1~\cite{hu2023gaia} formulates unsupervised sequence modeling of driving videos using multi-modal inputs, such as language, vision, and action, to generate discrete video tokens. GAIA-2~\cite{russell2025gaia2} improves generation fidelity and control by combining structured priors (vehicle states, road layout, scene semantics) with a latent diffusion backbone, enabling high-resolution, multi-camera-consistent video synthesis across diverse traffic settings.

In addition to image-based models, emerging work explores 3D-structured neural simulators that explicitly model physical occupancy or scene geometry. DriveWorld~\cite{Min2024DriveWorld4P} builds a city-scale traffic simulator with causal interactions between agents, offering a richly structured environment for embodied planning. Similarly, DOME~\cite{gu2024dome} proposes a diffusion-based world model that predicts future 3D occupancy frames instead of RGB pixels, achieving high-fidelity and long-horizon predictions with fine-grained controllability. AETHER \cite{team2025aether} introduces a single geometry-aware framework, simultaneously preserving scene geometric consistency and ensuring the inferability of generated content, thereby establishing a geometric foundation for a unified world model. Complementarily, DeepVerse\cite{chen2025deepverse} recasts world modeling as a 4D autoregressive video-generation task, produces future predictions that are both coherent and controllable. These approaches highlight that neural simulators need not produce RGB videos: structured 3D/4D world representations can equally serve as simulation backbones in control-intensive domains. 

Overall, video and 3D generative models as neural simulators enable controllable, high-fidelity, and richly structured world synthesis, serving as scalable alternatives to traditional simulators for training intelligent agents. As model expressivity and control improve, their role in simulation, rare event synthesis, and data-driven decision-making is expected to grow across domains such as autonomous driving, humanoid robotics, and beyond.

\subsubsection{World Models as Dynamic Models}
\label{World_Models_as_Dynamic_Models}

\begin{figure}[tp!] 
    \vspace{-1mm}
    \centering
    \includegraphics[width=0.8\columnwidth]{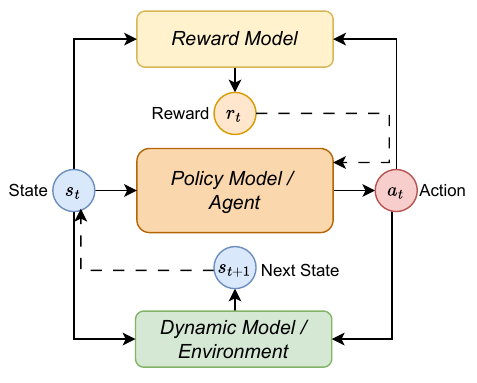}  % 宽度改为单栏宽度
    \caption{The general framework of Model-based RL. The agent learns a dynamic model $f: (s_t, a_t) \rightarrow s_{t+1}$ and a reward model $r: (s_t, a_t) \rightarrow r_t$, which are used to simulate interactions and improve policy learning.}
    \label{fig:mbrl}
    \vspace{-1mm}
\end{figure}

In model-based reinforcement learning (MBRL), the agent builds an internal model of the environment. This model usually includes \textit{dynamic model}, \textit{reward model}, and \textit{policy model}. The agent uses this model to simulate interactions with the environment, which helps the agent to make better decisions, as shown in Fig.~\ref{fig:mbrl}.
Instead of relying solely on real-world interactions, the agent learns a dynamic model and a reward model from collected experience, and subsequently performs planning or policy learning within the simulated environment. This decoupling of environment modeling and policy optimization significantly enhances sample efficiency, which is especially valuable in scenarios with costly, slow, or high-risk data collection.

\begin{figure*}[t]
    \centering
    \includegraphics[width=0.85\linewidth]{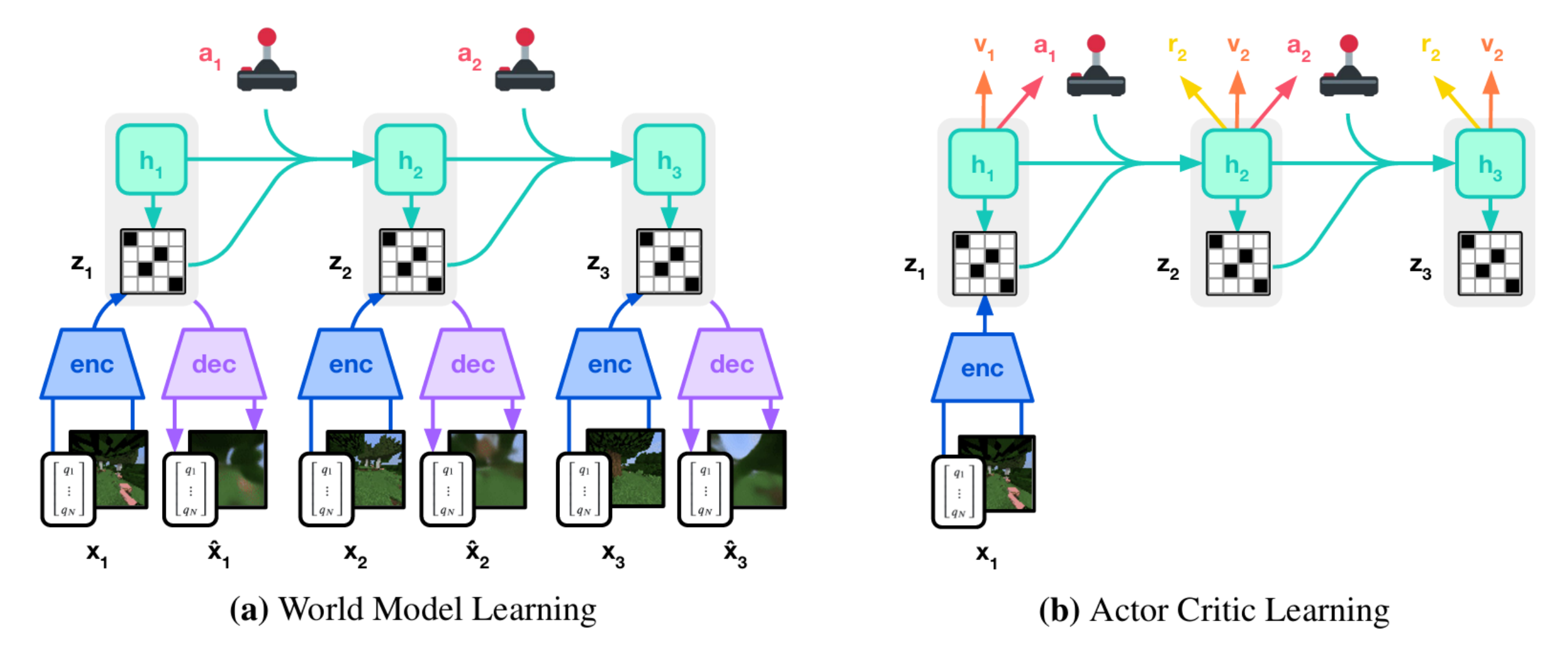}
    \caption{Core idea of the Dreamer series\cite{hafner2023mastering}: learn a latent world model as a dynamics model and optimize policies through model-based RL}
    \label{fig:dreamer}
\end{figure*}
World models can serve as general-purpose dynamic models in MBRL by learning to predict future states or observations conditioned on past interactions. Rather than relying on handcrafted symbolic rules or low-dimensional physics simulators, world models learn internal representations of environment dynamics directly from data. These models enable the agent to simulate hypothetical futures, perform rollouts for planning, and optimize behavior through imagined experiences. Depending on the design, world models may operate in pixel space, latent space, or structured representations, but their shared objective is to capture the temporal evolution of the environment in a way that supports efficient policy learning. This modeling flexibility allows world models to generalize across tasks, incorporate rich perceptual inputs, and serve as a scalable foundation for decision-making in complex domains.

\textbf{The Dreamer series} systematically explores latent-space modeling of environment dynamics from visual input. As shown in Fig.~\ref{fig:dreamer}, its central idea is to encode high-dimensional observations into compact, predictive latent trajectories using variational autoencoding and state-space modeling, enabling efficient policy learning through imagined rollouts. DreamerV1~\cite{dreamerv1} introduced the Recurrent State-Space Model (RSSM), demonstrating strong performance on pixel-based control benchmarks such as DMC. DreamerV2~\cite{dreamerv2} extended this framework with discrete latent variables to improve modeling capacity and generalization, achieving human-level scores on Atari. DreamerV3~\cite{hafner2023dreamerv3} further incorporated normalization and training stabilization mechanisms, becoming the first general world model to reach state-of-the-art performance across a broad range of visual control tasks. DayDreamer~\cite{wu2023daydreamer} validated the real-world applicability of this approach by deploying a Dreamer-style model on physical robots.

% While Dreamer demonstrates the effectiveness of recurrent latent modeling, its reliance on RNN-based architectures imposes limitations on generalization and long-horizon prediction. To address this, TransDreamer~\cite{chen2022transdreamer} replaces the recurrent dynamics with a Transformer-based transition model, leveraging self-attention over entire latent histories. This improves the model’s ability to capture long-range dependencies and enhances planning stability and predictive performance in visually complex environments.

To improve model generalization and transfer, recent work explores pretraining world models on real-world video data. ContextWM~\cite{contextwm} uses natural videos to learn generalizable visual dynamics in an unsupervised manner. By introducing a context modulation mechanism, the model selectively attends to predictable spatial-temporal regions, enabling more sample-efficient fine-tuning in downstream robotics and driving tasks. These results suggest that high-quality visual pretraining can endow world models with transferrable priors across domains.

Beyond latent modeling, token-based approaches directly model visual dynamics in discrete video space. iVideoGPT~\cite{wu2024ivideogpt} exemplifies this line of work, using VQ-VAE~\cite{vqvae} to tokenize videos, actions, and rewards into multi-modal sequences. A Transformer is then trained to autoregressively predict future tokens conditioned on past context. Unlike Dreamer-style methods, iVideoGPT bypasses explicit state construction and instead samples full video rollouts to augment offline policy learning via standard actor-critic methods~\cite{konda1999actor}. This direct modeling approach offers greater flexibility and demonstrates strong generalization in complex, multi-step planning tasks.

In conclusion, world models serve as dynamic models by enabling agents to learn, simulate, and plan based on environment dynamics. Whether implemented in latent space or high-dimensional visual space, they form the backbone of modern MBRL systems by unifying perception, prediction, and decision-making in a single generative framework.

\subsubsection{World Models as Reward Models}

Designing effective reward signals remains a fundamental challenge in reinforcement learning (RL), particularly in open-ended or complex environments. Traditional methods often rely on manually crafted reward functions, which are expensive to define and may fail to capture meaningful behavioral cues. This limitation has motivated the development of methods that automatically infer rewards from weakly supervised or unlabeled data, such as raw videos.

Recent advances in generative world models, especially those trained for video prediction, offer a promising direction for reward inference. These models learn to capture the underlying dynamics and structure of expert demonstrations, enabling them to serve as implicit reward models. The key insight is that if an agent’s behavior leads to trajectories that are easier for the model to predict, it likely aligns with the implicit preferences embedded in the training data. Thus, the model’s prediction confidence can be interpreted as a learned reward signal.

VIPER~\cite{viper} exemplifies this approach. It trains an autoregressive video world model on expert demonstrations, and then uses the model’s prediction likelihood during online agent behavior as the reward. When the agent acts in a way that produces highly predictable trajectories, it receives higher rewards. This enables learning high-quality policies without manually defined rewards. VIPER demonstrates strong performance on visual control benchmarks such as DMC, Atari, and RLBench. Additionally, due to the generalization ability of the learned world model, it can transfer reward inference across different embodiments and environments, such as variations in robotic arms or scene layouts.

Looking ahead, with continued progress in long-horizon modeling and multi-modal conditional generation, generative world models are poised to play a central role in reward modeling, such as humanoid control, multi-agent collaboration, and embodied interaction. Unifying dynamic modeling and reward inference within a single generative framework holds the potential to advance general-purpose agents that learn directly from raw visual experience.

\section{World Models for Intelligent Agents}
\label{sec:world_models_for_AD_robot}
Autonomous driving and articulated robots (including robotic arms, quadruped robots, and humanoid robots) are two critical applications of artificial intelligence and embodied intelligence.
Autonomous driving vehicles, can be seen as an intelligent robot with four wheels together with a smaller action space comparing to humanoid robots. Also, autonomous driving is a critical application of artificial intelligence, imposes stringent requirements on world modeling. 
%This section explores the applications and challenges of world models, particularly video generation-based models, in this domain. 
Further, autonomous driving systems require real-time comprehension and forecasting of complex and dynamic road environments.
As discussed in the previous section, video generation models as world models have demonstrated capacity for capturing physics and dynamic interactions, making them well-suited for the highly dynamic, high-stakes context of autonomous driving. 
%They are emerging as a unifying framework for perception, prediction, and decision-making.
Articulated robots (including robotic arms, quadruped robots, and humanoid robots), as the core carriers of Embodied Intelligence, pose unique and stringent requirements for world modeling. 
Similar to autonomous driving, robot systems also need to predict and understand its surroundings to conduct complex loco-manipulation tasks in human-centered environment. 
%As shown in the Fig.~\ref{fig:robotic_wm}, this section systematically explores the applications and challenges of world models in the field of articulated robot control, focusing on how methods based on robot simulation and multi-modal learning are driving a paradigm shift in robot technology.

Both autonomous driving and articulated robots require rich, long-term, and safe interactions with the real world during their applications, which places high demands on the precise and predictable modeling of the real world. 
%Autonomous driving systems require real-time comprehension and forecasting of complex and dynamic road environments, and world models offer promising solutions to these challenges through their comprehensive environmental representation. 
This section explores the applications and challenges of world models in autonomous drive and articulated robots, particularly in the context of video generation-based models.

%\xiao{将这一小节合并到6.2的前言里面把，这样比较对称。这些limitation可以简单summarize一些，和下面合并}

\begin{figure*}[ht!]
    \centering
    \includegraphics[width=1.9\columnwidth]{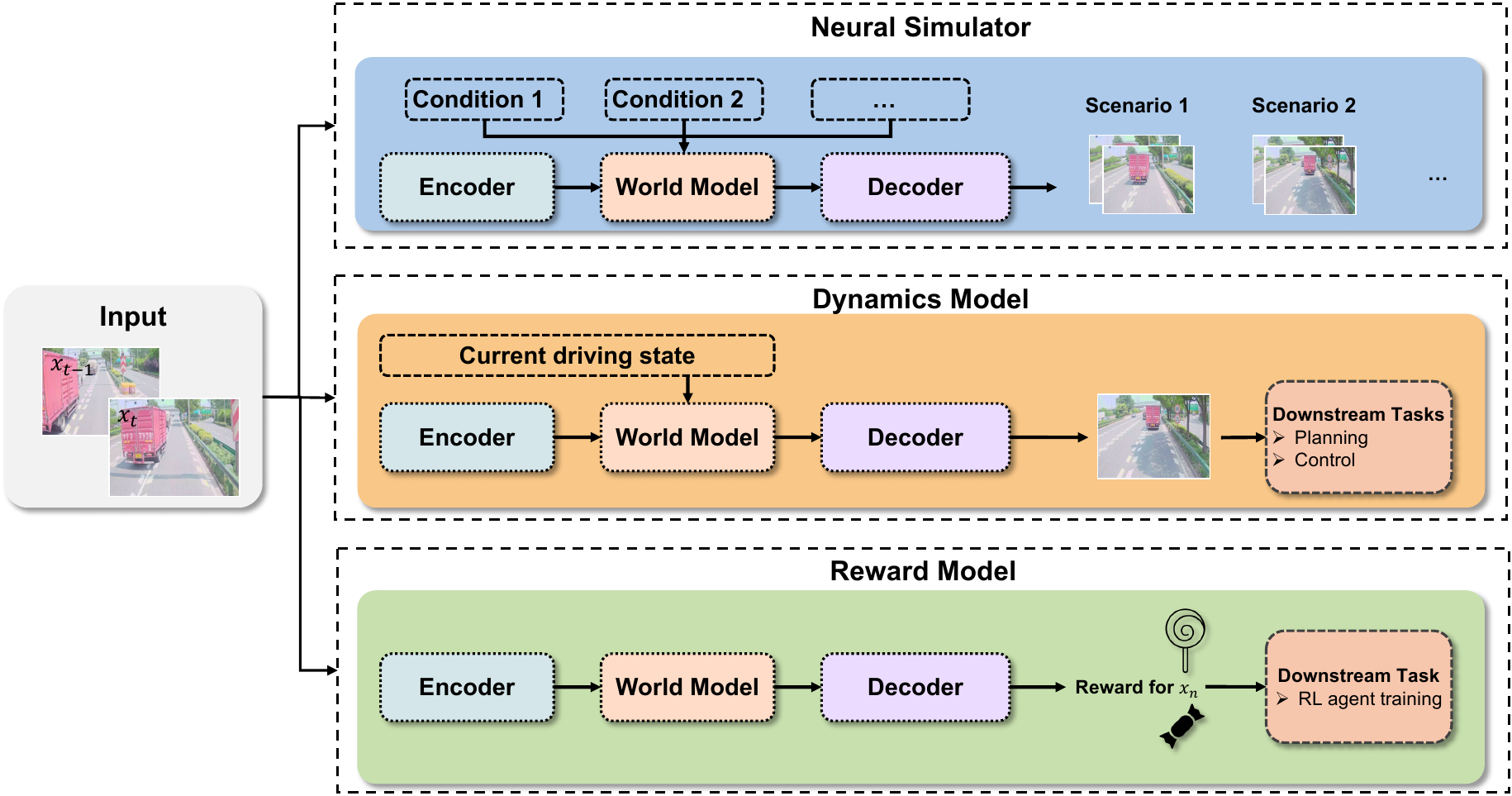}
    \caption{Illustration of the world model's three roles in autonomous driving: Neural Simulator, Dynamics
    Model and Reward Model.}
    \label{fig:wm_overview}
    \vspace{-1mm}
\end{figure*}

\subsection{World Models for Autonomous Driving}
~\label{subsec:WM_for_AD}
Traditional autonomous driving architectures employ a modular design with independent modules for perception, prediction, planning, and control \cite{mcallister2017concrete, Yurtsever2020}. While facilitating development and testing, this design exhibits critical limitations: errors in perception modules compound and amplify during processing \cite{zhang2023perception}, temporal modeling of long-term dependencies remains challenging \cite{Peng2019}, and performance deteriorates significantly in unfamiliar scenarios, as demonstrated by the 2018 Uber autonomous vehicle accident where the system repeatedly failed to correctly identify a pedestrian despite detecting them 5.6 seconds before impact \cite{UberCrash2018}.

Tesla share similar visual encoding architectures in the research and development of autonomous driving and robots~\cite{tesla_2022}, reflecting the commonality of perception technology.
Video generation-based world models have become a key area of research in
autonomous driving, progressing from early foundational models to sophisticated systems
that have achieved significant advancements in scene generation, multi-view
consistency, closed-loop simulation, and reasoning. Unlike general video generation
world models discussed previously, those designed for autonomous driving
prioritize the unique characteristics and safety demands of traffic scenarios, leading
to the development of several notable technical approaches and application paradigms.

\begin{figure*}[ht!]
    \centering
    \includegraphics[width=2.0\columnwidth]{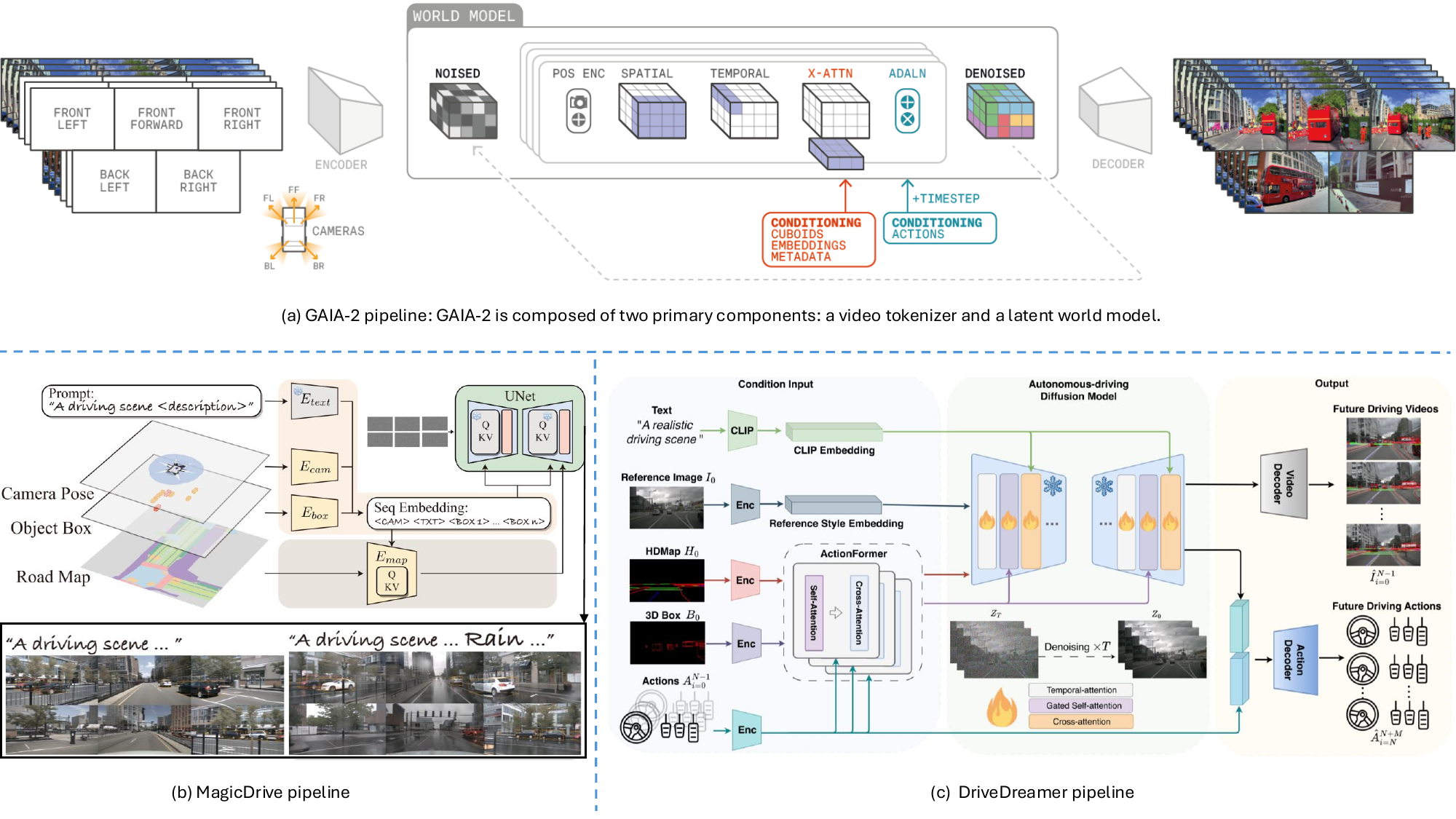}
    \caption{Illustration of the world model's representative pipelines: GAIA-2~\cite{russell2025gaia}, MagicDrive~\cite{gao2023magicdrive}, and DriveDreamer~\cite{wang2024drivedreamer}}
    \label{fig:wm_ad_pipelines}
    \vspace{-1mm}
\end{figure*}

We follow the previous categorization in section \ref{sec:wm_role}, and categorize autonomous driving world models into three main types as shown in Fig.~\ref{fig:wm_overview}: Neural Simulator, Dynamic Model, and Reward Model.
Three representative pipelines of autonomous driving world models are illustrated in Fig.~\ref{fig:wm_ad_pipelines}.

\subsubsection{WMs as Neural Simulators for Autonomous Driving}
Neural simulators focus on generating realistic driving scenarios for training
and testing autonomous systems. These models typically take multi-modal inputs (images, text,
actions, trajectories) and produce high-fidelity video sequences that simulate diverse driving
conditions for data augmentation and safety validation.

\textbf{GAIA-1}~\cite{hu2023gaia} pioneered treating world modeling as sequence prediction in autonomous driving, integrating video, text, and action inputs through an autoregressive transformer architecture to generate realistic driving scenarios. The model leverages a 9-billion parameter transformer trained on 4,700 hours of proprietary driving data, demonstrating emergent behaviors such as predicting diverse future scenarios from the same context and reasoning about interactions with dynamic agents. A key innovation is its ability to generate long, diverse driving scenes purely from learned understanding, with controllable generation through action conditioning and text prompts that can influence environmental aspects like weather and time of day.

Building upon this foundation, and also transitioning from autoregressive to diffusion-based approaches, \textbf{GAIA-2}~\cite{russell2025gaia} advanced controllable generation with structured conditioning for ego-vehicle dynamics, multi-agent interactions, and environmental factors. The model integrates both structured conditioning inputs and external latent embeddings from proprietary driving models, enabling fine-grained control over weather, lighting, and scene geometry across diverse geographical environments including the UK, US, and Germany. This enhanced controllability facilitates scalable simulation of both common and rare safety-critical driving scenarios through high-resolution, spatiotemporally consistent multi-camera video generation. 

\textbf{DriveDreamer}~\cite{wang2024drivedreamer} also introduced diffusion-based generation with structured traffic constraints, representing a major advancement in driving scenario generation. The model demonstrates superior performance on real-world data by learning from actual driving scenarios rather than simulated environments, employing a two-stage training pipeline where the first stage learns structured traffic constraints and the second stage enables future state anticipation. This approach allows for generating precise and controllable videos that faithfully capture the structural constraints of real-world traffic.
\textbf{DriveDreamer-2}~\cite{DriveDreamer2_2024} enhanced the DriveDreamer framework by integrating Large Language Models (LLMs) to enable natural language-driven scenario generation. The system operates through an LLM interface that converts textual user queries into agent trajectories, followed by HDMap generation that adheres to traffic regulations, and finally employs a Unified Multi-View Model (UniMVM) to generate driving videos with enhanced temporal and spatial coherence. This represents the first world model capable of generating customized driving videos based on user descriptions, achieving improvements in quality while enabling generation of diverse and uncommon driving scenarios crucial for robust autonomous driving training.
\textbf{DriveDreamer4D}~\cite{DriveDreamer4D_2024} further enhanced 4D driving scene representation by leveraging world model priors to synthesize novel trajectory videos with explicit spatiotemporal consistency control. The framework uses existing world models as "data machines" to synthesize novel trajectory videos with structured conditions that explicitly control the spatiotemporal consistency of traffic elements. A key innovation is the "cousin data training strategy" that effectively merges real and synthetic data for optimizing 4D Gaussian Splatting-based reconstruction, significantly improving generation quality for novel trajectory views and spatial temporal coherence.

Aiming to address the challenge of 3D geometry control, \textbf{MagicDrive}~\cite{gao2023magicdrive} introduced a novel street view generation framework that accepts diverse inputs including camera poses, road maps, 3D bounding boxes, and textual descriptions. The system employs tailored encoding strategies for varied inputs and incorporates a cross-view attention module to ensure consistency across multiple camera views, enabling high-fidelity street-view synthesis that accurately captures nuanced 3D geometry and diverse scene characteristics such as lighting and weather.
\textbf{MagicDrive3D}~\cite{MagicDrive3D_2024} extended this capability to enable controllable 3D generation for any-view rendering through a novel two-step "generation-first, reconstruction-later" pipeline. The framework first trains a conditional multi-view video generation model, then reconstructs 3D scenes from generated video data using Deformable Gaussian Splatting with monocular depth initialization and appearance modeling. This approach represents the first framework to effectively integrate geometry-free view synthesis with geometry-focused reconstruction for controllable 3D street scene generation using common driving datasets.
\textbf{MagicDrive-V2}~\cite{gao2024magicdrivedit} scaled the framework to high-resolution, long-duration videos using Diffusion Transformer architecture integrated with 3D Variational Autoencoders. The system incorporates an MVDiT block for multi-view video generation, novel spatial-temporal conditional encoding for precise geometric control, and a three-stage progressive bootstrapping training strategy that enables synthesis of videos up to 848×1600 resolution and 241 frames. This represents a significant advancement in output resolution, frame count, and control fidelity for autonomous driving applications. \textbf{Panacea}~\cite{wen2024panacea} focused on panoramic video generation with multi-view consistency mechanisms and super-resolution capabilities. The framework employs a two-stage video generation model that first synthesizes multi-view images, then constructs multi-view videos from these images, incorporating multi-view appearance noise prior mechanisms and super-resolution modules to enable generation of high-quality, high-resolution 360-degree videos.

Focusing on multi-camera consistency, \textbf{WoVoGen}~\cite{lu2024wovogen} introduced explicit 4D world volumes as foundational elements, operating in two phases to envision future 4D temporal world volumes and generate multi-camera videos with sensor interconnectivity. The model addresses the challenge of maintaining intra-world consistency and inter-sensor coherence by incorporating 4D world volumes that adeptly integrate time with spatial data. The generated 4D world feature is geometrically transformed to sample 3D image volumes for each camera, enabling high-quality street-view video generation that responds to vehicle control inputs while maintaining consistency across sensors.

Recent works have also explored occupancy representations for more structured scene understanding. \textbf{OccSora}~\cite{wang2024occsora} employed a diffusion-based 4D occupancy generation model to simulate 3D world development, generating 16-second videos with authentic 3D layouts and temporal consistency. The framework uses a 4D scene tokenizer to obtain compact, discrete spatio-temporal representations from 4D occupancy input, followed by a diffusion transformer trained on these representations for trajectory-conditioned 4D occupancy generation.
\textbf{DriveWorld}~\cite{min2024driveworld} used occupancy-based Memory State-Space Models (MSSM) for 4D scene understanding by learning from multi-camera driving videos. The framework comprises a Dynamic Memory Bank for temporal-aware latent dynamics and a Static Scene Propagation module for spatial-aware latent statics. This approach enables comprehensive 4D scene understanding that benefits diverse vision-centric perception, prediction, and planning tasks, demonstrating improvements across multiple autonomous driving benchmarks.
\textbf{Drive-OccWorld}~\cite{yang2025driving} combined occupancy forecasting with end-to-end planning through semantic and motion-conditional normalization techniques. The framework features a memory module that accumulates semantic and dynamic information from historical BEV embeddings, enabling controllable generation of future states through flexible action conditions. This integration of forecasting capabilities with end-to-end planning opens new avenues for driving world generation and integrated planning systems.

Targeting at address time horizon challenges of generated result,
\textbf{InfinityDrive}~\cite{guo2024infinitydrive} enabled generation of indefinitely long driving sequences through generative models that maintain coherence over extended periods. The framework addresses challenges in sampling speed, long-duration generation, and temporal consistency, representing efforts to achieve controllable, efficient, and realistic video-based world modeling for autonomous driving applications.

\textbf{ReconDreamer}~\cite{ni2024recondreamer} enhanced driving scene reconstruction for closed-loop simulation through online restoration and progressive data update strategies. The framework introduces "DriveRestorer," a world model-based component that mitigates ghosting artifacts through online restoration, coupled with progressive data updates to ensure high-quality rendering especially for large and complex maneuvers. This represents the first method to effectively render large maneuvers such as multi-lane shifts spanning up to 6 meters.

\begin{table*}
    [htp] \center
    \caption{Comparison of \textbf{Key Research} for World Models in Autonomous Driving}
    \begin{threeparttable}
        \small 
        \renewcommand{\arraystretch}{1.45}
        \resizebox{1.0\textwidth}{!}{
        \begin{tabular}{llcccccccclc}
            \toprule \multirow{4}{*}{\textbf{Category}}                        & \multirow{4}{*}{\textbf{Paper}}                                 & \multicolumn{6}{c}{\textbf{Input}} & \multirow{4}{*}{\textbf{Output}} & \multirow{4}{*}{\textbf{\makecell{World Model\\Architecture}}} & \multirow{4}{*}{\textbf{Dataset}} & \multirow{4}{*}{\textbf{\makecell{Code}}\tnote{2}} \\
            \cmidrule(lr){3-8}                                                  %& & Image & Text & Action & trajectory & Geometry & Map &            &         \\
                                                                               &                                                                 & \rotatebox{90}{Image}              & \rotatebox{90}{Text}             & \rotatebox{90}{Action}                                         & \rotatebox{90}{Trajectory}        & \rotatebox{90}{Geometry}\tnote{1}                               & \rotatebox{90}{Map}   &                                                  &                                   \\
            \midrule \multirow{28}{*}{\textbf{\shortstack{Neural\\Simulator}}} & \cellcolor{LightGray}GAIA-1~\cite{hu2023gaia} & \cellcolor{LightGray}\cmark & \cellcolor{LightGray}\cmark & \cellcolor{LightGray}\cmark & \cellcolor{LightGray} & \cellcolor{LightGray} & \cellcolor{LightGray} & \cellcolor{LightGray}Image & \cellcolor{LightGray}Transformer & \cellcolor{LightGray}In-house & \cellcolor{LightGray} \\
          & DriveDreamer~\cite{wang2024drivedreamer} &\cmark    &\cmark    &\cmark    &       &\cmark    &\cmark    & Image & Diffusion & nuScenes &\cmark\\
          & \cellcolor{LightGray}ADriver-I~\cite{ADriver-I_2023} & \cellcolor{LightGray}\cmark & \cellcolor{LightGray}\cmark & \cellcolor{LightGray} & \cellcolor{LightGray} & \cellcolor{LightGray} & \cellcolor{LightGray} & \cellcolor{LightGray}Image & \cellcolor{LightGray}Diffusion & \cellcolor{LightGray}nuScenes~\cite{caesar2020nuscenes} \& In-house & \cellcolor{LightGray} \\
          & GAIA-2~\cite{russell2025gaia} &\cmark    &\cmark    &\cmark    &\cmark    &\cmark    &       & Image & Diffussion & In-house &  \\
          & \cellcolor{LightGray}DriveDreamer-2~\cite{DriveDreamer2_2024} & \cellcolor{LightGray}\cmark & \cellcolor{LightGray}\cmark & \cellcolor{LightGray}\cmark & \cellcolor{LightGray}\cmark & \cellcolor{LightGray}\cmark & \cellcolor{LightGray}\cmark & \cellcolor{LightGray}Image & \cellcolor{LightGray}Diffusion & \cellcolor{LightGray}nuScenes & \cellcolor{LightGray} \\
          & DriveDreamer4D~\cite{DriveDreamer4D_2024} &\cmark    &\cmark    &       &       &\cmark    &\cmark    & Image & Diffusion & Waymo Open dataset (WOD)~\cite{sun2020waymoopen} &\cmark\\
          & \cellcolor{LightGray}DrivingWorld~\cite{hu2024drivingworld} & \cellcolor{LightGray}\cmark & \cellcolor{LightGray} & \cellcolor{LightGray} & \cellcolor{LightGray}\cmark & \cellcolor{LightGray} & \cellcolor{LightGray} & \cellcolor{LightGray}Image & \cellcolor{LightGray}Transformer & \cellcolor{LightGray}NuPlan~\cite{karnchanachari2024nuplan} \& In-house & \cellcolor{LightGray} \\
          & MagicDrive~\cite{gao2023magicdrive} &\cmark    &\cmark    &       &       &\cmark    &\cmark    & Image & DiT   & nuScenes &\cmark\\
          & \cellcolor{LightGray}MagicDrive3D~\cite{MagicDrive3D_2024} & \cellcolor{LightGray} & \cellcolor{LightGray}\cmark & \cellcolor{LightGray} & \cellcolor{LightGray}\cmark & \cellcolor{LightGray}\cmark & \cellcolor{LightGray}\cmark & \cellcolor{LightGray}Image & \cellcolor{LightGray}DiT & \cellcolor{LightGray}nuScenes & \cellcolor{LightGray} \\
          & MagicDrive-V2~\cite{gao2024magicdrivedit} &\cmark    &\cmark    &       &\cmark    &\cmark    &\cmark    & Image & DiT   & nuScenes &  \\
          & \cellcolor{LightGray}WoVoGen~\cite{lu2024wovogen} & \cellcolor{LightGray}\cmark & \cellcolor{LightGray}\cmark & \cellcolor{LightGray} & \cellcolor{LightGray} & \cellcolor{LightGray}\cmark & \cellcolor{LightGray}\cmark & \cellcolor{LightGray}Image & \cellcolor{LightGray}Diffusion & \cellcolor{LightGray}nuScenes, Occ3d~\cite{tian2023occ3d}, nuScenes-lidarseg & \cellcolor{LightGray} \\
          & ReconDreamer~\cite{ni2024recondreamer} &\cmark    &       &       &\cmark    &\cmark    &\cmark    & Image & Diffusion & Waymo Open dataset (WOD) &\cmark\\
          & \cellcolor{LightGray}DualDiff+~\cite{DualDiff_2024} & \cellcolor{LightGray}\cmark & \cellcolor{LightGray}\cmark & \cellcolor{LightGray} & \cellcolor{LightGray} & \cellcolor{LightGray}\cmark & \cellcolor{LightGray}\cmark & \cellcolor{LightGray}Image & \cellcolor{LightGray}Transformer & \cellcolor{LightGray}nuScenes & \cellcolor{LightGray} \\
          & Panacea~\cite{wen2024panacea} &\cmark    &\cmark    &       &       &\cmark    &\cmark    & Image & Diffusion & nuScenes &  \\
          & \cellcolor{LightGray}Cosmos-Transfer1~\cite{Cosmos-Transfer1_2024} & \cellcolor{LightGray}\cmark & \cellcolor{LightGray}\cmark & \cellcolor{LightGray} & \cellcolor{LightGray} & \cellcolor{LightGray}\cmark & \cellcolor{LightGray} & \cellcolor{LightGray}Image & \cellcolor{LightGray}DiT & \cellcolor{LightGray}Cosmos~\cite{agarwal2025cosmos} & \cellcolor{LightGray}\cmark \\
          & GeoDrive~\cite{chen2025geodrive} &\cmark    &       &       &\cmark    &\cmark    &       & Image & Diffusion & nuScenes &  \\
          & \cellcolor{LightGray}DriveWorld~\cite{min2024driveworld} & \cellcolor{LightGray}\cmark & \cellcolor{LightGray}\cmark & \cellcolor{LightGray}\cmark & \cellcolor{LightGray} & \cellcolor{LightGray} & \cellcolor{LightGray} & \cellcolor{LightGray}Occupancy & \cellcolor{LightGray}Transformer & \cellcolor{LightGray}nuScenes, Openscene~\cite{openscene2023} & \cellcolor{LightGray} \\
          & OccSora~\cite{wang2024occsora} &       &       &       &\cmark    &       &       & Occupancy & Diffusion & nuScenes &\cmark\\
          & \cellcolor{LightGray}Drive-OccWorld~\cite{yang2025driving} & \cellcolor{LightGray}\cmark & \cellcolor{LightGray}\cmark & \cellcolor{LightGray}\cmark & \cellcolor{LightGray}\cmark & \cellcolor{LightGray} & \cellcolor{LightGray} & \cellcolor{LightGray}Occupancy & \cellcolor{LightGray}Transformer & \cellcolor{LightGray}nuScenes, Lyft-Level5~\cite{houston2021one} & \cellcolor{LightGray} \\
          & DOME~\cite{gu2024dome} &       &       &       &\cmark    &\cmark    &       & Occupancy & Diffusion & nuScenes &\cmark\\
          & \cellcolor{LightGray}RenderWorld~\cite{yan2024renderworld} & \cellcolor{LightGray} & \cellcolor{LightGray} & \cellcolor{LightGray} & \cellcolor{LightGray} & \cellcolor{LightGray}\cmark & \cellcolor{LightGray} & \cellcolor{LightGray}Occupancy & \cellcolor{LightGray}Transformer & \cellcolor{LightGray}nuScenes & \cellcolor{LightGray} \\
          & OccLLama~\cite{wei2024occllama} &       &\cmark    &       &\cmark    &\cmark    &       & Occupancy & Transformer & NuScenes, Occ3D, NuScenes-QA~\cite{qian2024nuscenes} &  \\
          & \cellcolor{LightGray}BEVWorld~\cite{zhang2024bevworld} & \cellcolor{LightGray}\cmark & \cellcolor{LightGray} & \cellcolor{LightGray}\cmark & \cellcolor{LightGray} & \cellcolor{LightGray}\cmark & \cellcolor{LightGray} & \cellcolor{LightGray}Image, point cloud & \cellcolor{LightGray}Diffusion & \cellcolor{LightGray}nuScenes, Carla simulation~\cite{dosovitskiy2017carla} & \cellcolor{LightGray} \\
          & HoloDrive~\cite{wu2024holodrive} &\cmark    &\cmark    &       &       &\cmark    &\cmark    & Image, point cloud & Transformer & NuScenes &  \\
          & \cellcolor{LightGray}GEM~\cite{hassan2024gem} & \cellcolor{LightGray}\cmark & \cellcolor{LightGray} & \cellcolor{LightGray} & \cellcolor{LightGray}\cmark & \cellcolor{LightGray}\cmark & \cellcolor{LightGray} & \cellcolor{LightGray}Image, depth & \cellcolor{LightGray}Diffusion & \cellcolor{LightGray}BDD \cite{xu2017end}, etc \cite{yang2024generalized, grauman2024ego} & \cellcolor{LightGray}\cmark \\
          & DriveArena~\cite{DriveArena_2023} &\cmark    &\cmark    &       &       &\cmark    &\cmark    & Image & Diffusion & nuScenes &\cmark\\
          & \cellcolor{LightGray}ACT-Bench~\cite{arai2024act} & \cellcolor{LightGray}\cmark & \cellcolor{LightGray}\cmark & \cellcolor{LightGray}\cmark & \cellcolor{LightGray}\cmark & \cellcolor{LightGray} & \cellcolor{LightGray} & \cellcolor{LightGray}Image & \cellcolor{LightGray}Transformer & \cellcolor{LightGray}nuScenes, ACT-Bench & \cellcolor{LightGray} \\
          & InfinityDrive~\cite{guo2024infinitydrive} &\cmark    &\cmark    &\cmark    &       &       &       & Image & Transformer & nuScenes &  \\
          &
          \cellcolor{LightGray}Epona~\cite{zhang2025epona} &\cellcolor{LightGray}\cmark & \cellcolor{LightGray} & \cellcolor{LightGray} & \cellcolor{LightGray}\cmark & \cellcolor{LightGray} & \cellcolor{LightGray} & \cellcolor{LightGray}Image & \cellcolor{LightGray}Transformer & \cellcolor{LightGray}NuPlan &\cellcolor{LightGray}\cmark \\ 
          & DrivePhysica~\cite{yang2024pysical} &\cmark    &\cmark    &\cmark    &       &\cmark    &\cmark    & Image & Diffusion & nuScenes &  \\
          & \cellcolor{LightGray}Cosmos-Drive~\cite{ren2025cosmos} &\cellcolor{LightGray}\cmark & \cellcolor{LightGray}\cmark & \cellcolor{LightGray}\cmark & \cellcolor{LightGray}\cmark & \cellcolor{LightGray}\cmark & \cellcolor{LightGray}\cmark & \cellcolor{LightGray}Image & \cellcolor{LightGray}DiT & \cellcolor{LightGray}In-house, WOD &\cellcolor{LightGray}\cmark \\
          \midrule \multirow{24}{*}{\textbf{\shortstack{Dynamic\\Model}}} & MILE~\cite{hu2022model} &\cmark    &       &\cmark    &       &       &       & Image, BEV Semantics  & RNN   & Carla simulation &\cmark\\
          & \cellcolor{LightGray}Cosmos-Reason1~\cite{Cosmos-Reason1_2025} & \cellcolor{LightGray}\cmark & \cellcolor{LightGray}\cmark & \cellcolor{LightGray}\cmark & \cellcolor{LightGray} & \cellcolor{LightGray} & \cellcolor{LightGray} & \cellcolor{LightGray}Text & \cellcolor{LightGray}Transformer & \cellcolor{LightGray}Cosmos-Reason-1-Dataset & \cellcolor{LightGray} \\
          & TrafficBots~\cite{zhang2023trafficbots} &       &       &\cmark    &\cmark    &\cmark    &\cmark    & Trajectory & Transformer & Waymo Open Motion Dataset &\cmark\\
          & \cellcolor{LightGray}Uniworld~\cite{min2023uniworld} & \cellcolor{LightGray}\cmark & \cellcolor{LightGray} & \cellcolor{LightGray} & \cellcolor{LightGray} & \cellcolor{LightGray} & \cellcolor{LightGray} & \cellcolor{LightGray}Occupancy & \cellcolor{LightGray}Transformer & \cellcolor{LightGray}nuScenes & \cellcolor{LightGray} \\
          & Copilot4D~\cite{zhang2023copilot4d} &       &       &       &       &\cmark    &       & Point cloud & Diffusion & nuScenes, KITTI~\cite{geiger2013vision}  &  \\
          & \cellcolor{LightGray}MUVO~\cite{bogdoll2023muvo} & \cellcolor{LightGray}\cmark & \cellcolor{LightGray} & \cellcolor{LightGray}\cmark & \cellcolor{LightGray} & \cellcolor{LightGray}\cmark & \cellcolor{LightGray} & \cellcolor{LightGray}Image, Occupancy, Point cloud & \cellcolor{LightGray}RNN & \cellcolor{LightGray}Carla simulation & \cellcolor{LightGray} \\
          & OccWorld~\cite{zheng2024occworld} &       &       &       &\cmark    &\cmark    &       & Occupancy & Transformer & nuScenes, Occ3D &\cmark\\
          & \cellcolor{LightGray}ViDAR~\cite{yang2024visual} & \cellcolor{LightGray}\cmark & \cellcolor{LightGray} & \cellcolor{LightGray}\cmark & \cellcolor{LightGray} & \cellcolor{LightGray} & \cellcolor{LightGray} & \cellcolor{LightGray}Point cloud & \cellcolor{LightGray}Transformer & \cellcolor{LightGray}nuScenes & \cellcolor{LightGray} \\
          & Think2Drive~\cite{li2024think2drive} &       &       &       &       &\cmark    &\cmark    & BEV Semantics  & RNN   & CARLA simulation &  \\
          & \cellcolor{LightGray}LidarDM~\cite{zyrianov2024lidardm} & \cellcolor{LightGray} & \cellcolor{LightGray} & \cellcolor{LightGray} & \cellcolor{LightGray} & \cellcolor{LightGray}\cmark & \cellcolor{LightGray} & \cellcolor{LightGray}Point cloud & \cellcolor{LightGray}Diffusion & \cellcolor{LightGray}KITTI-360, WOD & \cellcolor{LightGray} \\
          & LAW~\cite{li2024enhancing} &\cmark    &       &\cmark    &       &       &       & BEV Semantics  & Transformer & nuScenes, NAVSIM~\cite{dauner2024navsim}, CARLA simulation &\cmark\\
          & \cellcolor{LightGray}UnO~\cite{agro2024uno} & \cellcolor{LightGray} & \cellcolor{LightGray} & \cellcolor{LightGray} & \cellcolor{LightGray} & \cellcolor{LightGray}\cmark & \cellcolor{LightGray} & \cellcolor{LightGray}Occupancy & \cellcolor{LightGray}Transformer & \cellcolor{LightGray}nuScenes, Argoverse 2, KITTI Odometry & \cellcolor{LightGray} \\
          & CarFormer~\cite{hamdan2024carformer} &       &       &\cmark    &\cmark    &\cmark    &\cmark    & Trajectory & Transformer & CARLA simulation &\cmark\\
          & \cellcolor{LightGray}NeMo~\cite{huang2024neural} & \cellcolor{LightGray}\cmark & \cellcolor{LightGray} & \cellcolor{LightGray} & \cellcolor{LightGray} & \cellcolor{LightGray} & \cellcolor{LightGray} & \cellcolor{LightGray}Image, Occupancy & \cellcolor{LightGray}Transformer & \cellcolor{LightGray}nuScenes & \cellcolor{LightGray} \\
          & Covariate Shift~\cite{popov2024mitigating} &\cmark    &       &\cmark    &       &       &       & Image, BEV Semantics  & Transformer & CARLA simulation &  \\
          & \cellcolor{LightGray}Imagine-2-Drive~\cite{garg2024imagine} & \cellcolor{LightGray}\cmark & \cellcolor{LightGray} & \cellcolor{LightGray} & \cellcolor{LightGray}\cmark & \cellcolor{LightGray} & \cellcolor{LightGray} & \cellcolor{LightGray}Image & \cellcolor{LightGray}Diffusion & \cellcolor{LightGray}CARLA simulation & \cellcolor{LightGray} \\
          & Doe-1~\cite{zheng2024doe} &\cmark    &\cmark    &       &\cmark    &       &       & Image & Transformer & nuScenes &\cmark\\
          & \cellcolor{LightGray}GaussianWorld~\cite{zuo2025gaussianworld} & \cellcolor{LightGray}\cmark & \cellcolor{LightGray} & \cellcolor{LightGray}\cmark & \cellcolor{LightGray} & \cellcolor{LightGray}\cmark & \cellcolor{LightGray} & \cellcolor{LightGray}Occupancy & \cellcolor{LightGray}TransFormer & \cellcolor{LightGray}nuScenes & \cellcolor{LightGray} \\
          & DFIT-OccWorld~\cite{zhang2024efficient} &\cmark    &       &       &\cmark    &\cmark    &       & Image, Occupancy & Transformer & Occ3D, nuScenes, OpenScene &  \\
          & \cellcolor{LightGray}DrivingGPT~\cite{chen2024drivinggpt} & \cellcolor{LightGray}\cmark & \cellcolor{LightGray} & \cellcolor{LightGray}\cmark & \cellcolor{LightGray} & \cellcolor{LightGray} & \cellcolor{LightGray} & \cellcolor{LightGray}Image & \cellcolor{LightGray}Transformer & \cellcolor{LightGray}nuPlan  & \cellcolor{LightGray} \\
          & AdaWM~\cite{wang2025adawm} &       &\cmark    &       &       &\cmark    &       & Trajectory & RNN   & CARLA simulation &  \\
          & \cellcolor{LightGray}AD-L-JEPA~\cite{zhu2025ad} & \cellcolor{LightGray} & \cellcolor{LightGray} & \cellcolor{LightGray} & \cellcolor{LightGray} & \cellcolor{LightGray}\cmark & \cellcolor{LightGray} & \cellcolor{LightGray}Point cloud & \cellcolor{LightGray}JEPA & \cellcolor{LightGray}KITTI3D, WOD & \cellcolor{LightGray} \\
          & HERMES~\cite{zhou2025hermes} &\cmark    &\cmark    &\cmark    &       &       &       & Point cloud & Transformer & nuScenes, OmniDrive-nuScenes &  \\
          \midrule \multirow{5}{*}{\textbf{\shortstack{Reward\\Model}}} & \cellcolor{LightGray}SEM2~\cite{gao2024enhance} & \cellcolor{LightGray}\cmark & \cellcolor{LightGray} & \cellcolor{LightGray}\cmark & \cellcolor{LightGray} & \cellcolor{LightGray}\cmark & \cellcolor{LightGray} & \cellcolor{LightGray}Image, Point cloud & \cellcolor{LightGray}RNN & \cellcolor{LightGray}nuScenes & \cellcolor{LightGray} \\
          & Iso-Dream~\cite{pan2022iso} &\cmark    &       &\cmark    &       &       &       & Image & RNN   & nuScenes &\cmark\\
          & \cellcolor{LightGray}Vista~\cite{gao2024vista} & \cellcolor{LightGray}\cmark & \cellcolor{LightGray}\cmark & \cellcolor{LightGray}\cmark & \cellcolor{LightGray}\cmark & \cellcolor{LightGray} & \cellcolor{LightGray} & \cellcolor{LightGray}Image & \cellcolor{LightGray}Diffusion & \cellcolor{LightGray}nuScenes, etc \cite{sun2020waymoopen, yang2024generalized, li2022coda}. & \cellcolor{LightGray} \\
          & Drive-WM~\cite{Drive-WM2024} &\cmark    &\cmark    &\cmark    &       &\cmark    &\cmark    & Image & Diffusion & nuScenes &  \\
          & \cellcolor{LightGray}WoTE~\cite{li2025end} &\cellcolor{LightGray}       &\cellcolor{LightGray}       &\cellcolor{LightGray}       &\cellcolor{LightGray}       &\cellcolor{LightGray}\cmark    &\cellcolor{LightGray}\cmark    & \cellcolor{LightGray}BEV Semantics  & \cellcolor{LightGray}Transformer & \cellcolor{LightGray}NAVSIM &\cellcolor{LightGray}\cmark\\
          & IRL-VLA \cite{jiang2025irl} & \cmark & \cmark & \cmark & & & \cmark &Trajectory, Action Sequence & Diffusion & NAVSIM \\
    \bottomrule
    \end{tabular}%
    }
    \begin{tablenotes}
        \scriptsize \item[1] Geometry mean 3D geomtric representation which includes:
        3D voxel occupancy, 3D bounding box, 3D depth, 3D segmentation and 3D
        point cloud. \scriptsize \item[2]
        \begin{minipage}[t]{0.69\linewidth}
            Code availability: " " means code is not announced to be
            released in the paper or is announced to be released in the paper, but the full
            code (training and inference) is still no available yet, "\cmark" means code is released. 
        \end{minipage}\\
    \end{tablenotes}
\end{threeparttable}
\label{tab:ad_wm_comparison}
\end{table*}

\subsubsection{WMs as Dynamic Models for Autonomous Driving}
Dynamic models focus on learning the underlying physics and motion patterns in
driving environments, primarily serving perception, prediction, and planning tasks rather
than high-fidelity generation. These models learn environment dynamics to enable 
better decision-making and long-term planning.

\textbf{MILE}~\cite{hu2022model} pioneered model-based imitation learning for urban driving by jointly learning a predictive world model and driving policy from expert demonstrations. The framework achieves significant improvements in driving scores through high-resolution video inputs and 3D geometry as inductive bias, demonstrating how world models can serve as effective foundations for policy learning in complex urban environments. The joint learning approach enables the model to capture both environmental dynamics and appropriate driving behaviors simultaneously.

\textbf{TrafficBots}~\cite{zhang2023trafficbots} addressed multi-agent traffic simulation with configurable agent personalities through conditional variational autoencoders, enabling scalable simulation of diverse driving behaviors. The system introduces destination-based navigation and time-invariant personality latents to control agent behavior styles from aggressive to cautious, providing a comprehensive framework for simulating realistic multi-agent interactions in traffic scenarios.

Building on occupancy-based representations for better 3D understanding, \textbf{UniWorld}~\cite{min2023uniworld} employed 4D geometric occupancy prediction as a foundational pre-training task, achieving notable improvements in motion prediction, 3D object detection, and semantic scene completion. The framework demonstrates how 4D occupancy prediction can serve as a unified pre-training objective that benefits multiple downstream autonomous driving tasks through comprehensive spatiotemporal scene understanding.
\textbf{OccWorld}~\cite{zheng2024occworld} used vector-quantized variational autoencoders to learn discrete scene tokens from 3D occupancy data, enabling GPT-like spatial-temporal generative modeling. This approach reformulates scene understanding as a sequence modeling problem, allowing for efficient autoregressive prediction of future occupancy states while maintaining spatial and temporal coherence across extended horizons.
\textbf{GaussianWorld}~\cite{zuo2025gaussianworld} reformulated 3D occupancy prediction as 4D occupancy forecasting, using Gaussian world models to infer scene evolution considering ego motion, dynamic objects, and newly-observed areas. The framework leverages Gaussian representations to model uncertainty and temporal dynamics, enabling more robust forecasting in complex driving scenarios with multiple moving agents and changing environmental conditions.

\textbf{DFIT-OccWorld}~\cite{zhang2024efficient} introduced an efficient occupancy world model through decoupled dynamic flow, reformulating occupancy forecasting as voxel warping processes with image-assisted training paradigms. This approach improves computational efficiency while maintaining prediction accuracy by separating static and dynamic scene elements, enabling more scalable deployment in real-time autonomous driving systems.

Addressing sensor fusion and geometric understanding, \textbf{MUVO}~\cite{bogdoll2023muvo} incorporated spatial voxel representations to learn sensor-agnostic geometric understanding from both camera and LiDAR data. The framework addresses the neglect of physical and geometric attributes in existing world models by integrating multi-modal sensor inputs into unified 3D representations, enabling more comprehensive scene understanding that bridges the gap between different sensor modalities.
\textbf{ViDAR}~\cite{yang2024visual} introduced visual point cloud forecasting as a pre-training task, predicting future LiDAR point clouds based solely on historical visual input to foster synergic learning of semantics, 3D structures, and temporal dynamics. This cross-modal approach demonstrates how world models can learn to translate between different sensor modalities while maintaining spatiotemporal consistency and semantic understanding.

\textbf{LAW}~\cite{li2024enhancing} proposed self-supervised learning without perception labels by predicting future latent features based on current observations and ego-vehicle actions. This approach significantly reduces reliance on expensive manual annotations while enabling effective learning of environment dynamics through latent feature prediction. The framework demonstrates how world models can be trained efficiently without requiring dense supervision, making them more practical for large-scale deployment.
\textbf{Think2Drive}~\cite{li2024think2drive} demonstrated efficient reinforcement learning in latent space by training a neural planner within a compact latent world model, achieving expert-level proficiency in complex urban scenarios. The framework shows how latent world models can enable efficient planning and decision-making by abstracting complex high-dimensional observations into manageable latent representations that preserve essential information for control.
\textbf{HERMES}~\cite{zhou2025hermes} unified 3D scene understanding and generation within a single framework using Bird's-Eye View representations and "world queries" that incorporate world knowledge through causal attention mechanisms. The framework demonstrates how world models can integrate understanding and generation capabilities, enabling both perception and simulation within a unified architecture that leverages causal attention for improved temporal modeling.

\textbf{Cosmos-Reason1}~\cite{Cosmos-Reason1_2025} represents cutting-edge exploration in incorporating physical common sense and embodied reasoning, generating scenes that conform more closely to physical laws while reasoning about possible physical interactions. This work advances the field by integrating physical reasoning capabilities into world models, enabling more realistic simulation of complex physical interactions in driving scenarios.
\textbf{Doe-1}~\cite{zheng2024doe} formulated autonomous driving as a next-token generation problem using multi-modal tokens (observation, description, and action), enabling unified perception, prediction, and planning through autoregressive generation. This approach demonstrates how language model architectures can be adapted for autonomous driving by treating all modalities as tokens in a unified sequence, enabling end-to-end learning across perception and control tasks.
\textbf{DrivingGPT}~\cite{chen2024drivinggpt} combined driving world modeling and trajectory planning using multi-modal autoregressive transformers, treating interleaved discrete visual and action tokens as a unified "driving language." The framework demonstrates how driving can be formulated as a language modeling problem, enabling the application of large-scale language model techniques to autonomous driving while maintaining the ability to reason about both visual inputs and control outputs.

\subsubsection{WMs as Reward Models for Autonomous Driving} 
Reward models evaluate the quality and safety of driving behaviors, often integrated
with reinforcement learning for policy optimization. These models use world model 
predictions to assess trajectory safety and guide decision-making without requiring 
manually engineered reward functions.

\textbf{Vista}~\cite{gao2024vista} demonstrated generalizable reward functions using the model's own simulation capabilities, establishing a novel approach where the world model itself assesses the quality or safety of potential driving maneuvers by simulating their outcomes. The system supports versatile action controllability from high-level intentions to low-level maneuvers. This self-assessment capability enables the model to provide intrinsic motivation for safe driving behaviors without requiring manual reward engineering.

\textbf{WoTE}~\cite{li2025end} focused on trajectory evaluation using Bird's-Eye View world models for real-time safety assessment in end-to-end autonomous driving systems. The framework leverages BEV world models to predict future states and evaluate trajectory safety more efficiently than image-level approaches, demonstrating state-of-the-art performance on NAVSIM and Bench2Drive benchmarks. This approach enables continuous safety monitoring and trajectory assessment in real-time autonomous driving applications.

\textbf{Drive-WM}~\cite{Drive-WM2024} enabled multi-future trajectory exploration with image-based reward evaluation through joint spatial-temporal modeling and view factorization. The system can simulate multiple possible futures based on distinct driving maneuvers and determine optimal trajectories by evaluating imagined futures using image-based rewards. This capability supports safe driving planning through world model-based "what-if" reasoning, enabling the system to explore and evaluate multiple potential actions before execution.

\textbf{Iso-Dream}~\cite{pan2022iso} addressed the challenge of controllable versus non-controllable dynamics separation in driving environments. The approach enhances model-based reinforcement learning by isolating controllable dynamics (ego-vehicle actions) from non-controllable dynamics (other vehicles, environmental changes), enabling more effective long-horizon planning and decision-making in complex traffic scenarios. This separation enables more focused learning on controllable aspects while properly modeling environmental uncertainty.

\subsubsection{Technical Trends and Implications}

The evolution of autonomous driving world models reveals four major technological trends that are reshaping how we approach vehicle simulation and testing:

\noindent
\textbf{Generative Architecture Evolution from Autoregressive to Diffusion Models:} 
Early autonomous driving world models like GAIA-1 employed autoregressive transformer architectures that predicted future driving scenarios through sequential token generation. While these approaches demonstrated strong capabilities in learning high-level scene structures, they faced significant computational challenges when generating long-duration, high-fidelity driving videos due to the sequential nature of token prediction. The field has since witnessed a paradigm shift toward diffusion-based models, exemplified by the DriveDreamer series, GAIA-2, and WoVoGen, which offer superior control over generation quality. Modern hybrid architectures are beginning to emerge that combine autoregression with diffusion-based scene generation. The integration of diffusion transformers (DiT) in models like MagicDrive-V2 represents the latest evolution, combining the strengths of transformer attention mechanisms with diffusion-based generation. 

\noindent
\textbf{Multi-Modal Integration and Controllable Scenario Generation:} 
Recent autonomous driving world models have evolved from simple image-to-image generation to sophisticated multi-modal systems that integrate diverse input types including camera images, LiDAR point clouds, textual descriptions, vehicle trajectories, and high-definition maps. This evolution addresses the fundamental challenge in autonomous vehicle testing: the need to generate specific, controllable driving scenarios that can stress-test different aspects of the driving system under precisely defined conditions. Models like GAIA-2 and DriveDreamer-2 exemplify this trend by accepting structured inputs such as ego-vehicle dynamics, multi-agent configurations, environmental factors (weather, time of day), and road semantics to enable fine-grained control over generated scenarios. This multi-modal approach also facilitates the generation of synchronized multi-camera views can cover 360-degree. These advances have transformed world models into active simulation tools that can explore the vast space of possible driving scenarios, enabling more efficient validation of autonomous driving systems.

\noindent
\textbf{3D Spatial-Temporal Understanding and Occupancy-Based Representations:} 
A fundamental branch that has occurred in autonomous driving world models is to have a comprehensive 3D spatial-temporal modeling that better captures the true nature of driving environments. Early models focused primarily on generating realistic camera images, but this approach failed to provide the geometric consistency and 3D understanding necessary for training robust perception systems that must reason about object depths, occlusions, and spatial relationships in real driving scenarios. The transition to 3D-aware modeling is exemplified by frameworks like OccSora, Drive-OccWorld, and OccWorld, which represent driving scenes through 4D occupancy grids that encode both spatial structure and temporal dynamics in a unified representation. This 3D-aware modeling capability enables world models to serve not just as data generators but as comprehensive simulators that can predict how driving scenes evolve in response to ego-vehicle actions. The integration of techniques like Gaussian Splatting in models such as GaussianWorld and MagicDrive3D further enhances the geometric fidelity of generated scenes, enabling novel view synthesis and supporting the development of more robust perception algorithms.

\noindent
\textbf{End-to-End Integration with Autonomous Driving Pipelines:} 
Modern autonomous driving world models are increasingly designed not as standalone simulation tools but as a prediction component of modular End-to-End autonomous driving. Models like MILE, LAW, Think2Drive and WoTE exemplify this integrated approach by jointly learning world dynamics and driving policies, enabling end-to-end optimization that can minimize the accumulation of errors across different system components. The integration extends to reward modeling capabilities, where frameworks like Vista and Drive-WM use their own simulation capabilities to evaluate trajectory safety and guide policy learning without requiring manually engineered reward functions. Advanced integrated systems like Doe-1 and DrivingGPT demonstrate how world models can unify perception (scene understanding), prediction (future state forecasting), and planning (action generation) within a single neural architecture that treats all modalities as tokens in a unified sequence modeling problem. The ultimate goal of this integration trend is to create autonomous driving systems that can reason about their environment, predict future states, and plan safe actions within a unified learned representation.

%These advances collectively address the most critical challenges in autonomous driving development: enabling systematic validation through comprehensive scenario generation, providing realistic simulation environments that can replace expensive real-world testing, improving generalization across diverse geographical and environmental conditions, and supporting the development of more robust and safety-critical autonomous vehicle systems. The integration of these advanced world models into autonomous driving development pipelines promises to accelerate the deployment of safe and reliable autonomous vehicles by providing unprecedented capabilities for simulation-based testing, validation, and training.

\subsection{World Models for Articulated Robots}
~\label{subsec:WM_for_AR}
Articulated robots (including robotic arms, quadruped robots, and humanoid robots), as the core carriers of Embodied Intelligence, pose unique and stringent requirements for world modeling. This section systematically explores the applications and challenges of world models in the field of articulated robot control, focusing on how methods based on robot simulation and multi-modal learning are driving a paradigm shift in robot technology.

\begin{table*}
    [htp] \center
    \caption{Comparison of Researches for World Models in Robotics (Part I)}
    \begin{threeparttable}
        \footnotesize  
        \renewcommand{\arraystretch}{1.45}
        \resizebox{1.0\textwidth}{!}{
        \begin{tabular}{llccccccccccclc}
            \toprule \multirow{4}{*}{\textbf{Category}} & \multirow{4}{*}{\textbf{Paper}}   & \multirow{4}{*}{\textbf{Year}} & \multicolumn{6}{c}{\textbf{Input}} & \multirow{4}{*}{\textbf{Architecture}\tnote{2}} & \multirow{4}{*}{\textbf{Experiments}\tnote{3}} & \multirow{4}{*}{\textbf{\makecell{Code\\Availability}}\tnote{4}} \\
            \cmidrule(lr){4-9}                                                  
            &   &   & \rotatebox{90}{\footnotesize Image} & \rotatebox{90}{\footnotesize Proprioception} & \rotatebox{90}{\footnotesize Text} & \rotatebox{90}{\footnotesize Action} & \rotatebox{90}{\footnotesize Geometry}\tnote{1}   & \rotatebox{90}{\footnotesize Tactile} & & \\
            
            \midrule \multirow{13}{*}{\textbf{\shortstack{Neural\\Simulators}}} 
            & \cellcolor{LightGray}WhALE \cite{zhang2024whale}  & \cellcolor{LightGray}2024 & \cellcolor{LightGray}\cmark & \cellcolor{LightGray} & \cellcolor{LightGray} & \cellcolor{LightGray}\cmark & \cellcolor{LightGray} & \cellcolor{LightGray} & \cellcolor{LightGray}Transformer & \cellcolor{LightGray}Robotic arm & \cellcolor{LightGray}\xmark \\
            & RoboDreamr \cite{zhou2024robodreamer} & 2024 & \cmark & & \cmark & & & & Diffusion & Robotic arm & \cmark \\
            & \cellcolor{LightGray}DreMa \cite{barcellona2024dream} & \cellcolor{LightGray}2024 & \cellcolor{LightGray}\cmark & \cellcolor{LightGray} & \cellcolor{LightGray} &\cellcolor{LightGray} &\cellcolor{LightGray} &\cellcolor{LightGray} &\cellcolor{LightGray} GS reconstruction &\cellcolor{LightGray} Robotic arm &\cellcolor{LightGray} \cmark \\
            & DreamGen \cite{jang2025dreamgen} & 2025 & \cmark & &\cmark &  & & & DiT & Dual robotic arm & $\circ$ \\
            & \cellcolor{LightGray}EnerVerse \cite{huang2025enerverse} & \cellcolor{LightGray}2025 & \cellcolor{LightGray}\cmark & \cellcolor{LightGray} & \cellcolor{LightGray}\cmark & \cellcolor{LightGray} & \cellcolor{LightGray} \cmark & \cellcolor{LightGray} & \cellcolor{LightGray}Diffusion & \cellcolor{LightGray}Robotic arm & \cellcolor{LightGray}\cmark \\
            & WorldEval \cite{li2025worldeval} & 2025 & \cmark & & \cmark & & & & DiT & Robotic arm & \cmark \\
            & \cellcolor{LightGray}Cosmos (NVIDIA) \cite{agarwal2025cosmos} & \cellcolor{LightGray}2025 & \cellcolor{LightGray}\cmark & \cellcolor{LightGray} & \cellcolor{LightGray}\cmark & \cellcolor{LightGray} & \cellcolor{LightGray} & \cellcolor{LightGray} & \cellcolor{LightGray}Diffusion + Autoregressive & \cellcolor{LightGray}Robots & \cellcolor{LightGray}\cmark \\
            & Pangu \cite{HuaweiPangu}  & 2025 & \cmark & & \cmark & & & & / & / & \xmark \\
            & \cellcolor{LightGray}RoboTransfer \cite{liu2025robotransfer} & \cellcolor{LightGray}2025 & \cellcolor{LightGray}\cmark & \cellcolor{LightGray} & \cellcolor{LightGray}\cmark & \cellcolor{LightGray} & \cellcolor{LightGray}\cmark & \cellcolor{LightGray} & \cellcolor{LightGray}Diffusion & \cellcolor{LightGray}Robotic arm& \cellcolor{LightGray}$\circ$ \\
            & TesserAct \cite{zhen2025tesseract} & 2025 & \cmark & & \cmark & & & &DiT & Robotic arm &\cmark \\ 
            & \cellcolor{LightGray}3DPEWM\cite{zhou2025learning} & \cellcolor{LightGray}2025 & \cellcolor{LightGray}\cmark \cellcolor{LightGray}& \cellcolor{LightGray}& \cellcolor{LightGray}& \cellcolor{LightGray}\cmark &\cellcolor{LightGray} &\cellcolor{LightGray} &\cellcolor{LightGray} DiT &\cellcolor{LightGray} Mobile robot &\cellcolor{LightGray}\xmark \\
            & SGImageNav \cite{hu2025imaginative} & 2025 & \cmark & \cmark & \cmark & & & & LLM & Quadroped robot & \xmark \\
            & \cellcolor{LightGray}EmbodieDreamer \cite{wang2025embodiedreamer} & \cellcolor{LightGray}2025 & \cellcolor{LightGray}\cmark & \cellcolor{LightGray}\cmark & \cellcolor{LightGray} & \cellcolor{LightGray} & \cellcolor{LightGray} & \cellcolor{LightGray} & \cellcolor{LightGray}Diffusion + ACT & \cellcolor{LightGray} Robotic arm & \cellcolor{LightGray} $\circ$ \\
                       
            \midrule \multirow{27}{*}{\textbf{\shortstack{Dynamics\\Models}}}
            & PlaNet \cite{hafner2018planet} & 2018 & \cmark & & & & & & RSSM  & DMC & \cmark \\
            & \cellcolor{LightGray}Plan2Explore \cite{sekar2020planning} & \cellcolor{LightGray}2020 & \cellcolor{LightGray}\cmark & \cellcolor{LightGray} & \cellcolor{LightGray} & \cellcolor{LightGray} & \cellcolor{LightGray} & \cellcolor{LightGray} & \cellcolor{LightGray}RSSM & \cellcolor{LightGray}DMC & \cellcolor{LightGray}\cmark \\
            & Dreamer \cite{hafner2019dream} & 2020 & \cmark & & & & & & RSSM & DMC & \cmark \\
            & \cellcolor{LightGray}DreamerV2 \cite{okada2022dreamingv2} & \cellcolor{LightGray}2021 & \cellcolor{LightGray}\cmark & \cellcolor{LightGray} & \cellcolor{LightGray} & \cellcolor{LightGray} & \cellcolor{LightGray} & \cellcolor{LightGray} & \cellcolor{LightGray}RSSM & \cellcolor{LightGray}DMC & \cellcolor{LightGray}\cmark \\
            & DreamerV3 \cite{hafner2023dreamerv3} & 2023 & \cmark & & & & & & RSSM & DMC & \cmark \\
            & \cellcolor{LightGray}DayDreamer \cite{wu2023daydreamer} & \cellcolor{LightGray}2024 & \cellcolor{LightGray}\cmark & \cellcolor{LightGray}\cmark & \cellcolor{LightGray} & \cellcolor{LightGray} & \cellcolor{LightGray} & \cellcolor{LightGray}\cmark & \cellcolor{LightGray}RSSM & \cellcolor{LightGray}Robotic arm & \cellcolor{LightGray}\cmark \\
            & Dreaming \cite{okada2021dreaming} & 2021 & \cmark & & & & & & RSSM & DMC & \cmark \\
            & \cellcolor{LightGray}Dreaming V2 \cite{okadadreamingv2} & \cellcolor{LightGray}2021 & \cellcolor{LightGray}\cmark & \cellcolor{LightGray} & \cellcolor{LightGray} & \cellcolor{LightGray} & \cellcolor{LightGray} & \cellcolor{LightGray} & \cellcolor{LightGray}RSSM & \cellcolor{LightGray}DMC + RoboSuite & \cellcolor{LightGray}\cmark \\
            & DreamerPro \cite{deng2022dreamerpro} & 2022 & \cmark & & \cmark & & & & RSSM & DMC & \cmark \\
            & \cellcolor{LightGray}TransDreamer \cite{chen2022transdreamer} & \cellcolor{LightGray}2024 & \cellcolor{LightGray} & \cellcolor{LightGray} & \cellcolor{LightGray} & \cellcolor{LightGray}\cmark & \cellcolor{LightGray} & \cellcolor{LightGray} & \cellcolor{LightGray}TSSM & \cellcolor{LightGray}2D Simulation & \cellcolor{LightGray}\cmark \\
            & LEXA \cite{mendonca2021discovering} & 2021 & \cmark & & & \cmark & & & RSSM & Simulated robotic arm & \cmark \\
            & \cellcolor{LightGray}FOWM \cite{feng2023finetuning} & \cellcolor{LightGray}2023 & \cellcolor{LightGray}\cmark & \cellcolor{LightGray}\cmark & \cellcolor{LightGray} & \cellcolor{LightGray} & \cellcolor{LightGray} & \cellcolor{LightGray} & \cellcolor{LightGray}TD-MPC & \cellcolor{LightGray}Robotic arm & \cellcolor{LightGray}\cmark \\
            & SWIM \cite{mendonca2023structured} & 2023 & \cmark & & & & & & RSSM & Robotic arm & \xmark \\
            & \cellcolor{LightGray}ContextWM \cite{contextwm} & \cellcolor{LightGray}2023 & \cellcolor{LightGray}\cmark & \cellcolor{LightGray} & \cellcolor{LightGray} & \cellcolor{LightGray} & \cellcolor{LightGray} & \cellcolor{LightGray} & \cellcolor{LightGray}RSSM & \cellcolor{LightGray}DMC + CARLA + Meta-world & \cellcolor{LightGray}\cmark \\
            & iVideoGPT \cite{wu2024ivideogpt} & 2023 & \cmark & \cmark & \cmark & \cmark & & \cmark & AutoreGressive Transformer & Robotic arm & \cmark \\
            & \cellcolor{LightGray}DWL \cite{gu2024advancing} & \cellcolor{LightGray}2024 & \cellcolor{LightGray} & \cellcolor{LightGray}\cmark & \cellcolor{LightGray} & \cellcolor{LightGray} & \cellcolor{LightGray} & \cellcolor{LightGray} & \cellcolor{LightGray}Recurrent encoder & \cellcolor{LightGray}Humanoid robot & \cellcolor{LightGray}\xmark \\
            & Surfer \cite{ren2023surfer} & 2024 & \cmark &\cmark & \cmark &  & & & Transformer & Robotic arm & \cmark \\
            & \cellcolor{LightGray}GAS \cite{lin2024world} & \cellcolor{LightGray}2024 & \cellcolor{LightGray}\cmark & \cellcolor{LightGray} & \cellcolor{LightGray} & \cellcolor{LightGray} & \cellcolor{LightGray} & \cellcolor{LightGray} & \cellcolor{LightGray}RSSM & \cellcolor{LightGray}Surgical robot & \cellcolor{LightGray}\cmark \\
            & Puppeteer \cite{hansen2024hierarchical} & 2024 & \cmark & \cmark & & & & & TD-MPC2 & Simulation (56-DoF humanoid) & \cmark \\
            & \cellcolor{LightGray}TWIST \cite{yamada2024twist} & \cellcolor{LightGray}2024 & \cellcolor{LightGray}\cmark & \cellcolor{LightGray} & \cellcolor{LightGray} & \cellcolor{LightGray} & \cellcolor{LightGray} & \cellcolor{LightGray} & \cellcolor{LightGray}RSSM & \cellcolor{LightGray}Robotic arm & \cellcolor{LightGray}\xmark \\
            & PIVOT-R \cite{zhang2024pivot} & 2024 & \cmark & \cmark & \cmark & & & & Transformer & Robotic arm & \cmark \\
            & \cellcolor{LightGray}HarmonyDream \cite{ma2023harmonydream} & \cellcolor{LightGray}2024 & \cellcolor{LightGray}\cmark & \cellcolor{LightGray} & \cellcolor{LightGray} & \cellcolor{LightGray} & \cellcolor{LightGray} & \cellcolor{LightGray} & \cellcolor{LightGray}RSSM & \cellcolor{LightGray}Robotic arm & \cellcolor{LightGray}\cmark \\
            & SafeDreamer \cite{huang2023safedreamer} & 2024 & \cmark & & & & \cmark & & OSRP & simulation (Safety-Gymnasium) & \cmark \\
            & \cellcolor{LightGray}WMP \cite{lai2024world} & \cellcolor{LightGray}2024 & \cellcolor{LightGray} & \cellcolor{LightGray}\cmark & \cellcolor{LightGray} & \cellcolor{LightGray} & \cellcolor{LightGray}\cmark & \cellcolor{LightGray} & \cellcolor{LightGray}RSSM & \cellcolor{LightGray}Quadruped robot & \cellcolor{LightGray}\cmark \\
            & RWM\cite{li2025robotic} & 2025 & & \cmark & & \cmark & & & GRU+MLP & Quadruped robot & \xmark \\
            & \cellcolor{LightGray}RWM-O \cite{li2025offline} & \cellcolor{LightGray}2025 & \cellcolor{LightGray} & \cellcolor{LightGray}\cmark & \cellcolor{LightGray} & \cellcolor{LightGray} & \cellcolor{LightGray} & \cellcolor{LightGray} & \cellcolor{LightGray}GRU+MLP & \cellcolor{LightGray}Robotic arm & \cellcolor{LightGray}\xmark \\
            & SSWM \cite{aljalbout2025accelerating} & 2025 & \cmark & & & & & & SSM & Quadrotor & \xmark \\
            & \cellcolor{LightGray}WMR \cite{sun2025learning} & \cellcolor{LightGray}2025 & \cellcolor{LightGray} & \cellcolor{LightGray}\cmark & \cellcolor{LightGray} & \cellcolor{LightGray} & \cellcolor{LightGray} & \cellcolor{LightGray} & \cellcolor{LightGray}LSTM & \cellcolor{LightGray}Humanoid robot & \cellcolor{LightGray}\xmark \\
            & PIN-WM \cite{li2025pin} & 2025 & \cmark & & & & & & GS reconstruction & Robotic arm & $\circ$ \\
            & \cellcolor{LightGray}LUMOS \cite{nematollahi2025lumos} & \cellcolor{LightGray}2025 & \cellcolor{LightGray}\cmark & \cellcolor{LightGray} & \cellcolor{LightGray}\cmark & \cellcolor{LightGray} & \cellcolor{LightGray} & \cellcolor{LightGray} & \cellcolor{LightGray}RSSM & \cellcolor{LightGray}Robotic arm & \cellcolor{LightGray}\cmark \\
            & OSVI-WM \cite{goswami2025osvi} & 2025 & \cmark & & & & & & Transformer & Robotic arm & \xmark \\
            & \cellcolor{LightGray}FOCUS \cite{ferraro2025focus} & \cellcolor{LightGray}2025 & \cellcolor{LightGray}\cmark & \cellcolor{LightGray}\cmark & \cellcolor{LightGray} & \cellcolor{LightGray} & \cellcolor{LightGray} & \cellcolor{LightGray} & \cellcolor{LightGray}RSSM & \cellcolor{LightGray}Robotic arm & \cellcolor{LightGray}\cmark \\
            & FLIP \cite{gaoflip} & 2025 & \cmark & & \cmark & & & & DiT & Robotic arm & \cmark \\
            & \cellcolor{LightGray}EnerVerse-AC \cite{jiang2025enerverse} & \cellcolor{LightGray}2025 & \cellcolor{LightGray}\cmark & \cellcolor{LightGray}\cmark & \cellcolor{LightGray} & \cellcolor{LightGray}\cmark & \cellcolor{LightGray} & \cellcolor{LightGray} & \cellcolor{LightGray}Diffusion & \cellcolor{LightGray}Robotic arm & \cellcolor{LightGray}\cmark \\
            & FlowDreamer \cite{guo2025flowdreamer} & 2025 & \cmark & & & \cmark & \cmark & & Diffusion & Robotic arm & $\circ$ \\
            & \cellcolor{LightGray}HWM \cite{ali2025humanoid} & \cellcolor{LightGray}2025 & \cellcolor{LightGray}\cmark & \cellcolor{LightGray} & \cellcolor{LightGray} & \cellcolor{LightGray} & \cellcolor{LightGray} & \cellcolor{LightGray} & \cellcolor{LightGray}MVM + FM & \cellcolor{LightGray}Humanoid robot & \cellcolor{LightGray}\xmark \\
            & MoDemV2 \cite{lancaster2024modem} & 2023 & \cmark &\cmark & & & & & RSSM & Robotic arm & \cmark \\
            & \cellcolor{LightGray}V-JEPA 2 \cite{assran2025vjepa2} & \cellcolor{LightGray}2025 & \cellcolor{LightGray}\cmark & \cellcolor{LightGray}\cmark & \cellcolor{LightGray} & \cellcolor{LightGray}\cmark & \cellcolor{LightGray} & \cellcolor{LightGray} & \cellcolor{LightGray}JEPA & \cellcolor{LightGray}Robotic arm &\cellcolor{LightGray}\cmark \\
            & AdaWorld \cite{gao2025adaworld} & 2025 &\cmark & & & & &  & DiT & Robotic arm & \cmark \\
            \bottomrule
        \end{tabular}
        }
        \begin{tablenotes}
        \setlength{\parskip}{0pt}
            \tiny \item[1] 
            \begin{minipage}[t]{0.9\linewidth}
                Geometry means 3D geometric representation, which includes: 3D voxel occupancy, 3D bounding box, 3D depth, 3D segmentation and 3D point cloud.
            \end{minipage}
            \tiny \item[2] 
            \begin{minipage}[t]{0.85\linewidth}
                DiT = Diffusion Transformers; 
                RSSM = Recurrent State-Space Model; 
                TSSM = Transformer State-Space Model;
                SSM = State-Space Model;
                TD-MPC = Temporal Difference learning for Model Predictive Control; 
                OSRP = online safety-reward planning;
                GRU = Gated Recurrent Unit; MLP = Multilayer Perceptron;
                LSTM = Long Short-Term Memory; 
                GS = Gaussian Splatting; 
                MVM = Masked Video Modelling; FM = Flow Matching; 
                JEPA = Joint Embedding-Predictive Architecture.
            \end{minipage}
            \tiny \item[3] 
            \begin{minipage}[t]{0.9\linewidth}
                DMC = DeepMind Control suite (Simulation).
            \end{minipage}
            \tiny \item[4]
            \begin{minipage}[t]{0.9\linewidth}
                Code availability: "\xmark" means code is not announced to be released in the paper, "\cmark" means code is released, "$\circ$" means code is announced to be released in the paper, but the full code (training and inference) is not yet available.
            \end{minipage}
        \end{tablenotes}
    \end{threeparttable}
\end{table*}

% 第二个表格
\begin{table*}
    [htp] \center
    \caption{Comparison of Researches for World Models in Robotics (Part II)}
    \begin{threeparttable}
        \footnotesize  
        \renewcommand{\arraystretch}{1.45}
        \resizebox{1.0\textwidth}{!}{
        \begin{tabular}{llccccccccccclc}
            \toprule \multirow{4}{*}{\textbf{Category}} & \multirow{4}{*}{\textbf{Paper}}   & \multirow{4}{*}{\textbf{Year}} & \multicolumn{6}{c}{\textbf{Input}} & \multirow{4}{*}{\textbf{Architecture}\tnote{2}} & \multirow{4}{*}{\textbf{Experiments}\tnote{3}} & \multirow{4}{*}{\textbf{\makecell{Code\\Availability}}\tnote{4}} \\
            \cmidrule(lr){4-9}                                                  
            &   &   & \rotatebox{90}{\footnotesize Image} & \rotatebox{90}{\footnotesize Proprioception} & \rotatebox{90}{\footnotesize Text} & \rotatebox{90}{\footnotesize Action} & \rotatebox{90}{\footnotesize Geometry}\tnote{1}   & \rotatebox{90}{\footnotesize Tactile} & & \\
            
            \midrule \multirow{3}{*}{\textbf{\shortstack{Dynamics\\Models}}}
            &\cellcolor{LightGray}MoSim \cite{hao2025neural} &\cellcolor{LightGray}2025 &\cellcolor{LightGray} &\cellcolor{LightGray} \cmark &\cellcolor{LightGray} &\cellcolor{LightGray} \cmark &\cellcolor{LightGray} &\cellcolor{LightGray} &\cellcolor{LightGray} rigid-body dynamics + ODE &\cellcolor{LightGray}Robotic arm &\cellcolor{LightGray}\cmark \\
            & DALI \cite{roder2025dynamics} & 2025 & \cmark & \cmark & & \cmark & & & RSSM & DMC & \cmark \\
           &\cellcolor{LightGray} GWM \cite{lu2025gaussianworldmodel} & \cellcolor{LightGray} 2025 & \cellcolor{LightGray} \cmark & \cellcolor{LightGray} & \cellcolor{LightGray} & \cellcolor{LightGray} \cmark & \cellcolor{LightGray} &\cellcolor{LightGray} &\cellcolor{LightGray} DiT & \cellcolor{LightGray}Robotic arm &\cellcolor{LightGray} \cmark \\
            
            \midrule \multirow{2}{*}{\textbf{\shortstack{Reward\\Models}}} 
            & VIPER \cite{Escontrela23arXiv_VIPER}& 2023 & \cmark & \cmark & & \cmark & & & Autoregressive Transformer & DMC + RLBench + Atari & \cmark\\
            & \cellcolor{LightGray}PlaNet \cite{hafner2018planet} & \cellcolor{LightGray}2018 & \cellcolor{LightGray}\cmark & \cellcolor{LightGray} & \cellcolor{LightGray} & \cellcolor{LightGray} & \cellcolor{LightGray} & \cellcolor{LightGray} & \cellcolor{LightGray}RSSM  & \cellcolor{LightGray}DMC & \cellcolor{LightGray}\cmark \\
            \bottomrule
        \end{tabular}
        }
        \begin{tablenotes}
        \setlength{\parskip}{0pt}
            \tiny \item[1] 
            \begin{minipage}[t]{0.9\linewidth}
                Geometry means 3D geometric representation, which includes: 3D voxel occupancy, 3D bounding box, 3D depth, 3D segmentation and 3D point cloud.
            \end{minipage}
            \tiny \item[2] 
            \begin{minipage}[t]{0.85\linewidth}
                DiT = Diffusion Transformers; 
                RSSM = Recurrent State-Space Model; 
                TSSM = Transformer State-Space Model;
                SSM = State-Space Model;
                TD-MPC = Temporal Difference learning for Model Predictive Control; 
                OSRP = online safety-reward planning;
                GRU = Gated Recurrent Unit; MLP = Multilayer Percentron;
                LSTM = Long Short-Term Memory; 
                GS = Gaussian Splatting; 
                MVM = Masked Video Modelling; FM = Flow Matching; 
                JEPA = Joint Embedding-Predictive Architecture.
            \end{minipage}
            \tiny \item[3] 
            \begin{minipage}[t]{0.9\linewidth}
                DMC = DeepMind Control suite (Simulation).
            \end{minipage}
            \tiny \item[4]
            \begin{minipage}[t]{0.9\linewidth}
                Code availability: "\xmark" means code is not announced to be released in the paper, "\cmark" means code is released, "$\circ$" means code is announced to be released in the paper, but the full code (training and inference) is not yet available.
            \end{minipage}
        \end{tablenotes}
    \end{threeparttable}
\end{table*}

\subsubsection{WMs as Neural Simulators for Articulated Robots}

World models serve as neural simulators by learning to generate temporally coherent and semantically rich representations of physical environments from multimodal inputs (e.g., text, images, trajectories). These generative models provide a scalable, data-driven alternative to traditional physics-based simulators, enabling efficient training and evaluation of autonomous agents. 

\begin{figure}[tp!] 
    \vspace{-1mm}
    \centering
    \includegraphics[width=1.0\columnwidth]{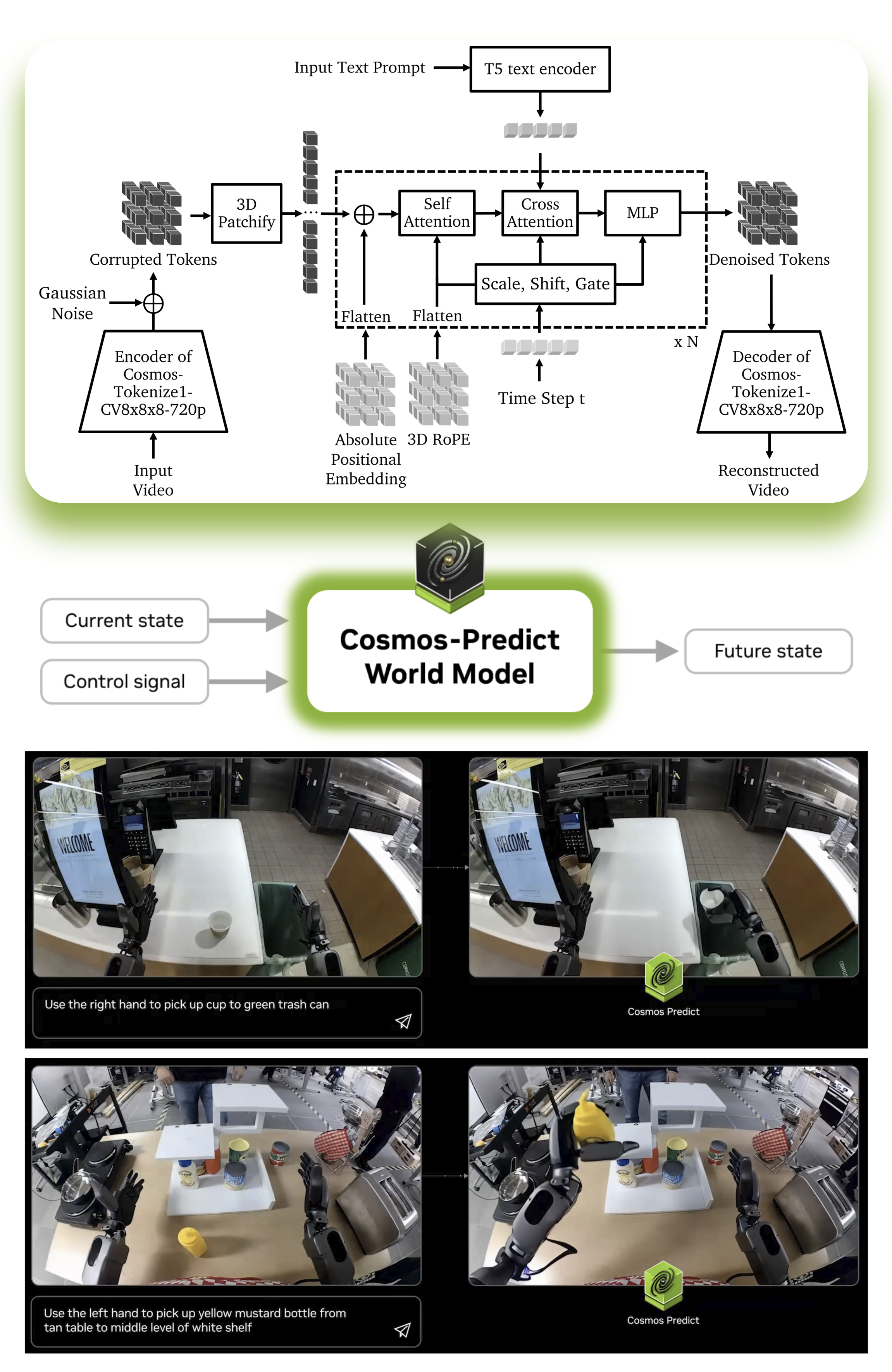}  % 宽度改为单栏宽度
    \caption{The Cosmos-Predict World Foundation Model processes input videos through Cosmos-Tokenize1-CV8×8×8-720p, encoding them into latent representations perturbed with Gaussian noise. A 3D patchification step structures these latents, followed by iterative self-attention, cross-attention (conditioned on text), and MLP blocks, modulated by adaptive layer normalization. Finally, the decoder reconstructs high-fidelity video output from the refined latent space. This architecture enables robust spatiotemporal modeling for diverse Physical AI applications\cite{agarwal2025cosmos}.}
    \label{fig:cosmos_predict}
    \vspace{-1mm}
\end{figure}

A prominent example is NVIDIA’s \textbf{Cosmos} World Foundation Model Platform \cite{agarwal2025cosmos}, which establishes a unified framework for building foundation world models capable of producing physics-accurate 3D video predictions through diffusion and autoregressive architectures. By synthesizing realistic, controllable environments from structured inputs (e.g., segmentation maps, depth), Cosmos facilitates sim-to-real transfer, data augmentation, and perception training for robotics. Fig.~\ref{fig:cosmos_predict} shows the \textbf{Cosmos-Predict} World Foundation Model. The platform’s modular design supports task-specific fine-tuning via post-training on specialized datasets, significantly reducing data requirements through transfer learning from large-scale pretraining. This approach bridges the gap between simulation and reality, enhancing robot learning in dynamic environments such as autonomous driving and robotic manipulation. Future advancements in neural simulation fidelity and control will further expand their role in embodied AI systems.

\textbf{WHALE}\cite{zhang2024whale} proposes a generalizable world model framework with behavior-conditioning and retracing-rollout for OOD generalization and uncertainty estimation. Whale-ST (spatial-temporal transformer) and Whale-X\cite{zhangwhale} (414 M-parameter model) demonstrate enhanced scalability and performance in simulation and real-world manipulation tasks.

\textbf{RoboDreamer}\cite{zhou2024robodreamer} introduces a compositional world model for robotic decision-making by factorizing video generation into primitives. It leverages language compositionality to generalize to unseen object-action combinations and multimodal goals, synthesizing plans for novel tasks in RT-X and outperforming monolithic baselines in simulation.

\textbf{DreMa}\cite{barcellona2024dream} introduces a compositional world model combining Gaussian Splatting and physics simulation to explicitly replicate real-world dynamics, enabling photorealistic future prediction and data-efficient imitation learning via equivariant transformations, achieving one-shot policy learning on a Franka robot with improved accuracy and generalization.

\textbf{DreamGen}\cite{jang2025dreamgen} introduces a 4-stage pipeline for training generalizable robot policies via neural trajectories, leveraging video world models to synthesize photorealistic data across behaviors and environments. It recovers pseudo-actions from generated videos, achieving zero-shot generalization with minimal real-world data, validated by DreamGen Bench benchmark.

\textbf{EnerVerse}\cite{huang2025enerverse} introduces a generative foundation model for robotics manipulation, employing autoregressive video diffusion and Free Anchor Views (FAVs) for 3D world modeling. The framework integrates 4D Gaussian Splatting in EnerVerse-D to reduce sim-to-real gaps, while EnerVerse-A translates 4D representations into actions, achieving state-of-the-art performance in simulated and real-world tasks.

\textbf{WorldEval}\cite{li2025worldeval} introduces a world model-based pipeline for online robot policy evaluation, using Policy2Vec to generate action-following videos via latent action conditioning. It enables scalable, reproducible ranking of policies and safety detection, demonstrating strong real-world correlation and outperforming real-to-sim methods.

Huawei Cloud’s \textbf{Pangu} World Model\cite{HuaweiPangu} is a neural simulator that synthesizes high-fidelity digital environments (e.g., camera videos, LiDAR point clouds) for training intelligent driving and embodied AI systems. By modeling physical dynamics and multimodal sensor data, it bypasses costly real-world data collection. Integrated with GAC Group, it enables rapid corner-case generation and 2D-to-3D pixel mapping. The CloudRobo Platform extends this with embodied AI models (generation, planning, execution) and R2C Protocol standardization, aiming to unify robotics development via cloud-based simulation and deployment.

\textbf{RoboTransfer} \cite{liu2025robotransfer} proposes a geometry-consistent video diffusion framework for robotic visual policy transfer, integrating multi-view geometry with explicit scene control. By enforcing cross-view feature interactions and depth/normal conditions, it synthesizes geometrically consistent multi-view videos, improving sim-to-real policy performance. The method enables fine-grained scene editing while maintaining visual fidelity.

\subsubsection{WMs as Dynamic Models for Articulated Robots}
\begin{figure*}[htbp] 
    \vspace{-1mm}
    \centering
    \includegraphics[width=2\columnwidth]{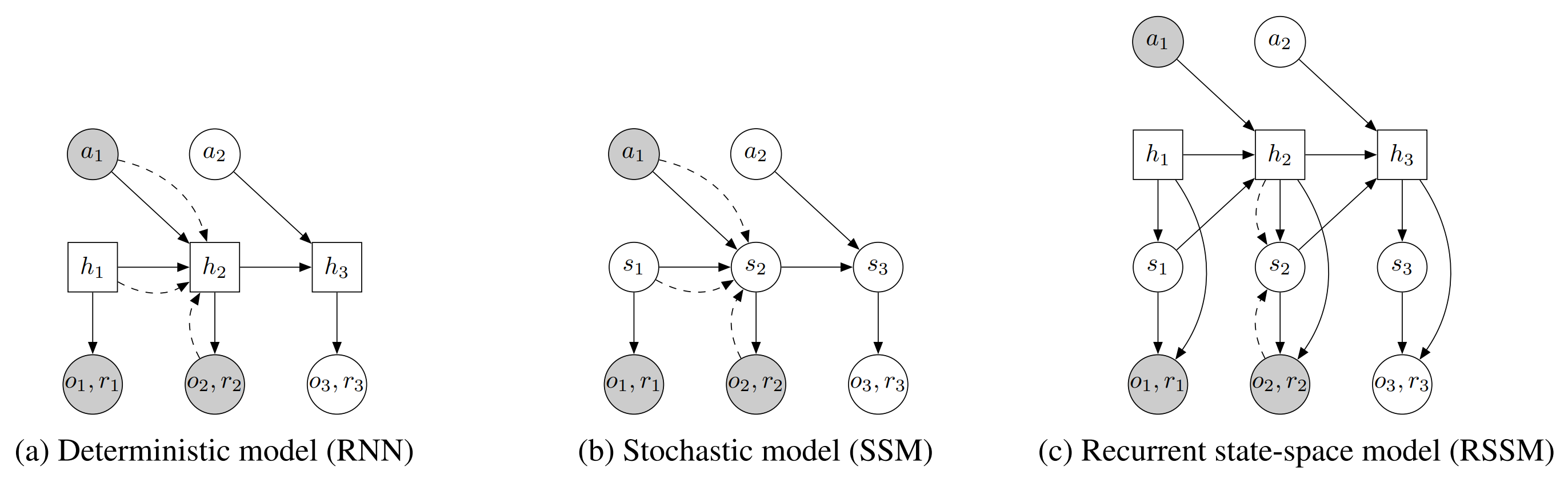}  % 宽度改为单栏宽度
    \caption{Latent dynamic models employ distinct transition mechanisms for temporal prediction. (a) RNN-based models use deterministic transitions, limiting multi-modal future prediction and enabling planner exploitation. (b) State-space models (SSM) rely solely on stochastic transitions, hindering long-term memory retention. (c) RSSM combines stochastic and deterministic state components, balancing multi-future prediction with temporal coherence. This hybrid approach enhances robustness in learning diverse futures while maintaining sequential consistency, as demonstrated in \cite{hafner2019learning}.}
    \label{fig:wm_RSSM}
    \vspace{-1mm}
\end{figure*}

World models serve as dynamic models in model-based reinforcement learning (MBRL) by learning predictive representations of environmental dynamics from observed data. Instead of relying on handcrafted physics engines or sparse reward signals, these models enable agents to simulate future states and plan actions through imagined rollouts. PlaNet was one of the earliest to use the RSSM architecture, as shown in Fig.~\ref{fig:wm_RSSM}. The Dreamer series exemplifies this approach by learning latent-state dynamics from high-dimensional observations using variational autoencoders and RSSM. Extensions like TransDreamer replace RNNs with Transformers to better capture long-range dependencies, while ContextWM and iVideoGPT explore pretraining on real-world videos and discrete token-based modeling, respectively. These advancements enhance generalization, long-horizon prediction, and transferability across tasks. By decoupling perception and planning, world models reduce sample complexity and improve decision-making in complex, high-dimensional environments. Recent work further demonstrates their real-world applicability, validating their potential for robotics and autonomous systems.

\textbf{PlaNet}\cite{hafner2019learning} presents a latent dynamic model for pixel-based planning, combining deterministic and stochastic transitions with latent overshooting for multi-step prediction. It solves complex continuous control tasks with fewer episodes than model-free methods, demonstrating high sample efficiency in unknown environments.

\textbf{Plan2Explore}\cite{sekar2020planning} introduces a self-supervised reinforcement learning agent that leverages model-based planning to actively seek future novelty during exploration, enabling zero or few-shot adaptation to unseen tasks. It outperforms prior methods in high-dimensional image-based control tasks without task-specific supervision, approaching oracle-level performance.

The \textbf{Dreamer series} does a lot in world model and has discussed it in Chapter.~\ref {World_Models_as_Dynamic_Models}. For example, \textbf{DreamerV3}\cite{hafner2023mastering} presents a generalist reinforcement learning algorithm that learns a world model to imagine future scenarios, achieving state-of-the-art performance across 150+ diverse tasks with a single configuration. Its robustness techniques enable stable cross-domain learning, demonstrated by first achieving diamond collection in Minecraft without human data or curricula. %The pipeline of DayDreamer of is shown as 

\textbf{Dreaming}\cite{okada2021dreaming} eliminates Dreamer's decoder to mitigate object vanishing, employing a likelihood-free InfoMax contrastive objective with linear dynamics and data augmentation, achieving top performance on 5 robotics tasks. \textbf{DreamingV2}\cite{okadadreamingv2} advances this by fusing DreamerV2's discrete latent states with Dreaming's reconstruction-free learning, creating a hybrid world model that leverages categorical state representation for complex environments and contrastive vision modeling, excelling in 3D robotic arm tasks without reconstruction.

\textbf{DreamerPro}\cite{deng2022dreamerpro} enhances MBRL robustness to visual distractions by integrating prototypical representations into Dreamer's world model, distilling temporal structures from recurrent states. This approach improves performance on DeepMind Control tasks with complex backgrounds, outperforming contrastive methods in both standard and distraction settings.

\textbf{LEXA}\cite{mendonca2021discovering} introduces a unified framework for unsupervised goal-reaching, combining world model-based imagined rollouts with foresight-driven exploration to discover novel states and achieve diverse tasks. It outperforms prior methods on 40 challenging robotic tasks, demonstrating zero-shot generalization and scalability across multiple environments.

\textbf{FOWM} \cite{feng2023finetuning} proposes a framework combining offline world model pretraining with online finetuning, using epistemic uncertainty regularization to mitigate extrapolation errors. It enables few-shot adaptation to seen/unseen visuo-motor tasks with limited offline data, validated on simulation and real-world robotic control benchmarks.

\textbf{SWIM} \cite{mendonca2023structured} proposes an affordance-space world model for robotic manipulation, trained on human videos and fine-tuned with minimal robot data. The model learns structured action representations from human-object interactions, enabling rapid skill acquisition (<30min) across diverse tasks and robots without task-specific supervision.

\textbf{DWL} \cite{gu2024advancing} is an end-to-end RL framework for humanoid locomotion. The world model enables zero-shot sim-to-real transfer, mastering diverse challenging terrains (snow, stairs, uneven ground) with a single policy. The approach demonstrates robustness and generalization without environment-specific tuning.

\textbf{Surfer} \cite{ren2023surfer} introduces a world model-based framework for robot manipulation, decoupling action and scene prediction to enhance generalization in multi-modal tasks. It incorporates explicit world knowledge modeling and is evaluated on SeaWave benchmark, achieving 54.74 $\%$ success rate, outperforming baselines by modeling physics-based state transitions.

\textbf{GAS} \cite{lin2024world} proposes a world-model-based deep reinforcement learning framework tailored for surgical robotic manipulation, employing pixel-level visuomotor policies with uncertainty-aware depth estimation and compact 3-channel image encoding. It achieves a 69$\%$ success rate in handling unseen objects and disturbances within real-world surgical environments, demonstrating superior robustness and generalization compared to prior methods in clinical settings.

\textbf{Puppeteer}\cite{hansen2024hierarchical} presents a hierarchical world model for visual whole-body humanoid control, where a high-level visual policy generates commands for a low-level execution policy, both trained via reinforcement learning. The approach achieves high-performance motion synthesis across 8 tasks for a 56-DoF humanoid without simplifying assumptions or reward engineering.

\textbf{TWIST}\cite{yamada2024twist} proposes a teacher-student world model distillation framework for efficient sim-to-real transfer in vision-based model-based RL. It leverages state-privileged teacher models to supervise image-based student models, accelerating adaptation while bridging the sim-to-real gap through domain-randomized distillation, outperforming naive methods in sample efficiency and task performance.

\textbf{PIVOT-R} \cite{zhang2024pivot} proposes a primitive-driven waypoint-aware world model (WAWM) for language-guided robotic manipulation, decoupling waypoint prediction from action execution with an asynchronous hierarchical executor (AHE). It achieves 19.45\% relative improvement on SeaWave benchmark while improving efficiency by 28× with minimal performance trade-off.

\textbf{HarmonyDream} \cite{ma2023harmonydream} proposes a task harmonization framework for world models, dynamically balancing observation and reward modeling losses to enhance sample-efficient MBRL. It achieves 10\%-69\% performance gains on visual robotic tasks and sets a new Atari 100K benchmark record by resolving task dominance issues in world model learning.

\textbf{SafeDreamer}\cite{huang2023safedreamer} integrates Lagrangian-based methods with world model planning in the Dreamer framework to enhance safe reinforcement learning. The approach achieves near-zero-cost performance on Safety-Gymnasium tasks, demonstrating effective balance between performance and safety for both low-dimensional and vision-only inputs through improved model accuracy and sample efficiency.

\textbf{WMP} \cite{lai2024world} proposes a world model-based perception framework for legged locomotion, eliminating privileged information reliance by learning policies from simulated world model predictions. It achieves state-of-the-art real-world traversability and robustness through cross-domain generalization, validated in simulation and physical environments.

\textbf{RWM}\cite{li2025robotic} introduces a neural network-based robotic world model with dual-autoregressive mechanisms for long-horizon dynamics prediction. The framework enables self-supervised training and robust policy optimization through imagined environments, addressing challenges in partial observability and sim-to-real transfer without domain-specific biases.

\textbf{RWM-O} \cite{li2025offline} introduces an offline robotic world model with explicit epistemic uncertainty estimation, penalizing unreliable transitions to enhance policy stability and generalization. Validated in real-world data settings, it reduces sim-to-real gaps and improves safety without physics simulators, outperforming conventional MBRL methods.

\textbf{SSWM} \cite{krinner2025accelerating}, state-space world models, was used to accelerate model-based reinforcement learning (MBRL). The approach parallelizes dynamic model training and utilizes privileged information, achieving up to 10 times faster world model training and 4 times overall MBRL speedup while maintaining performance in complex quadrotor flight tasks with partial observability.

\begin{figure*}[ht!] 
    \vspace{-1mm}
    \centering
    \includegraphics[width=2\columnwidth]{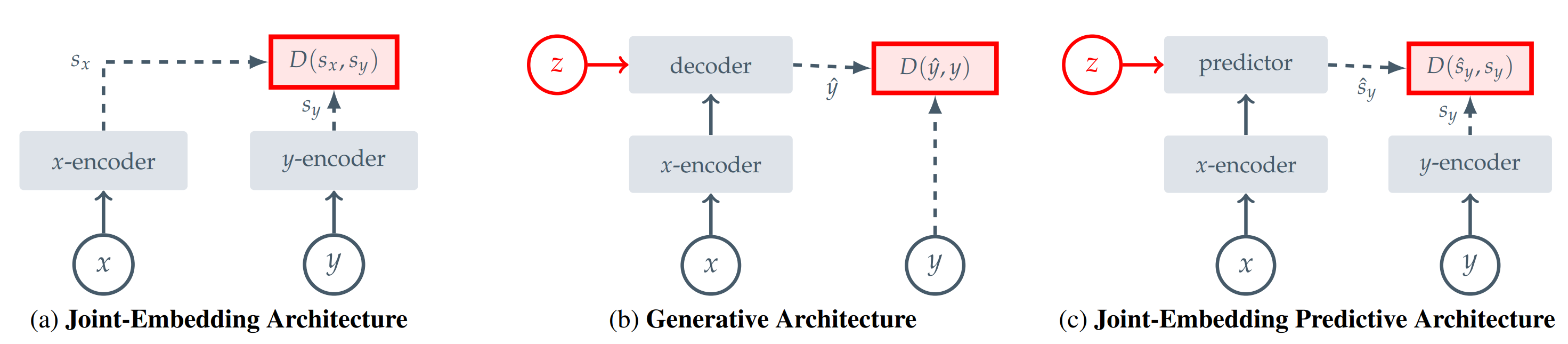}  % 宽度改为单栏宽度
    \caption{Self-supervised learning architectures employ distinct approaches to model input compatibility: (a) Joint-Embedding aligns embeddings of compatible inputs; (b) Generative reconstructs signals via latent-conditioned decoders; (c) Joint-Embedding Predictive forecasts embeddings using latent-conditioned predictors. Each framework optimizes energy assignments—low for compatible, high for incompatible inputs \cite{assran2023self}.}
    \label{fig:wm_JEPA}
    \vspace{-1mm}
\end{figure*}

\textbf{WMR} \cite{sun2025learning} proposes an end-to-end world model reconstruction framework for blind humanoid locomotion, explicitly reconstructing world states to enhance policy robustness. Gradient cutoff ensures independent state reconstruction, enabling adaptive locomotion on complex terrains, validated by 3.2 km hikes across ice, snow, and deformable surfaces.

\textbf{PIN-WM} \cite{li2025pin} presents a physics-informed world model for non-prehensile manipulation, enabling end-to-end 3D rigid body dynamics learning from few-shot visual trajectories via differentiable simulation. It eliminates state estimation with Gaussian Splatting-based observational loss and bridges Sim2Real gaps through physics-aware digital cousins, achieving robust policy transfer.

\textbf{LUMOS}\cite{nematollahi2025lumos} introduces a language-conditioned imitation learning framework leveraging world models for latent-space skill practice. The approach combines latent planning with hindsight goal relabeling and intrinsic rewards, enabling zero-shot transfer to real robots. It demonstrates superior performance on long-horizon tasks while mitigating policy-induced distribution shift in offline settings.

\textbf{OSVI-WM}\cite{gautam2025osvi} proposes a one-shot visual imitation learning framework using world-model-guided trajectory generation. A learned world model predicts latent states and actions from demonstrations, decoded into physical waypoints for execution. Evaluated across simulated and real robotic platforms, the method achieves over 30\% improvement compared to prior approaches.

\textbf{FOCUS}\cite{ferraro2025focus} introduces an object-centric world model for robotic manipulation, representing scenes through structured object interactions. The framework enables object-centric exploration and improves manipulation skills through more accurate scene predictions. Evaluated on robotic tasks, it demonstrates efficient learning and adaptation to sparse-reward scenarios using a Franka Emika robot arm.

\textbf{FLIP}\cite{gaoflip} presents a flow-centric model-based planning framework for language-vision manipulation, integrating multi-modal flow generation, flow-conditioned video dynamics, and vision-language representation modules. It synthesizes long-horizon plans via image flows, guiding low-level policy training with interactive world model properties, validated on diverse benchmarks.

\textbf{EnerVerse-AC} \cite{jiang2025enerverse} proposes an action-conditional world model for robotic evaluation, featuring multi-level action-conditioning and ray map encoding to generate dynamic multi-view observations. The model serves as both data engine and evaluator, synthesizing realistic action-conditioned videos from human-collected trajectories to enable cost-effective policy testing without physical robots or complex simulations.

\textbf{FlowDreamer} \cite{guo2025flowdreamer} proposes an RGB-D world model using explicit 3D scene flow representations for visual prediction and planning. It decouples motion estimation (U-Net) and frame synthesis (diffusion model) while maintaining end-to-end training, outperforming baselines by 7-11$\%$ in semantic quality and success rate across manipulation benchmarks.

\textbf{HWM}\cite{ali2025humanoid} introduces lightweight video-based world models for humanoid robotics, employing Masked Transformers and FlowMatching to forecast action-conditioned egocentric observations. The framework demonstrates efficient parameter sharing strategies, reducing model size by 33\%-53\% while maintaining performance, enabling practical deployment in resource-constrained academic settings.

\textbf{MoDem-V2} \cite{lancaster2024modem} enables real-world contact-rich manipulation learning via a model-based reinforcement learning framework that integrates demonstration bootstrapping and safety-aware exploration strategies (exploration centering, agency handover, actor-critic ensembles), achieving the first successful direct real-world training of vision-based MBRL systems without instrumentation.

\textbf{V-JEPA 2}\cite{assran2025vjepa2} is a 1.2B-parameter world model employing joint-embedding predictive architecture for video-based understanding, prediction and zero-shot planning. The model undergoes two-stage training: actionless pre-training on 1M+ video hours for physical intuition, followed by action-conditioned fine-tuning with minimal robot data (62 hours). Demonstrating state-of-the-art performance on action recognition and anticipation tasks, V-JEPA 2 enables model-predictive control for robotic tasks (65\%-80\% success in novel environments) through visual subgoal planning. The framework includes three new benchmarks for evaluating physical reasoning capabilities.

\subsubsection{WMs as Reward models for Articulated Robots}
A world model as a reward model leverages its learned dynamics to implicitly infer rewards by measuring how well an agent’s behavior aligns with the model’s predictions. For instance, if trajectories are highly predictable (i.e., match the world model’s expectations), they are assigned higher rewards, eliminating manual reward engineering. 

Unlike Dreamer, which implicitly replaces the reward signal through the value function, \textbf{PlaNet} uses an explicitly learned reward predictor. Its Reward Predictor, as part of the dynamics model, is responsible for predicting environmental rewards from compressed latent states, training by minimizing the error between the predicted reward and the true reward, and providing immediate reward signals for multi-step trajectory rolling in the latent space during the online planning stage, thereby replacing the hand-designed reward function.

Such approaches unify environment simulation and reward generation, enabling scalable RL from raw observations. This paradigm is particularly powerful for transfer learning, as seen in VIPER’s cross-embodiment generalization.

\textbf{VIPER}\cite{Escontrela23arXiv_VIPER} proposes using pretrained video prediction models as reward signals for reinforcement learning. The method trains an autoregressive transformer on expert videos and utilizes prediction likelihoods as rewards, enabling expert-level control without task-specific rewards across DMC, Atari and RLBench tasks, while supporting cross-embodiment generalization in tabletop manipulation scenarios.

\subsubsection{Technical Trends}
The world model has broad development prospects in the future, but in the field of robotics, it may have the following development potential:

\textbf{Tactile-Enhanced World Models for Dexterous Manipulation}.
The evolution of tactile-integrated world models is enabling breakthroughs in robotic dexterity, particularly for multi-fingered hands. Cutting-edge approaches now combine high-resolution contact modeling with visuo-tactile fusion, using neural networks to predict slip, deformation, and optimal grip forces in real time. Self-supervised tactile encoders eliminate manual labeling by autonomously learning material and shape representations, while graph/transformer architectures process dynamic spatiotemporal touch signals. These innovations allow robots to handle novel objects with human-like adaptability, overcoming traditional sim-to-real challenges in delicate manipulation tasks.

\textbf{Unified World Models for Cross-Hardware and Cross-Task Generalization}.
Future robotics world models will focus on hardware-agnostic dynamics encoding and task-adaptive latent spaces to generalize across diverse embodiments (e.g., single/double arms, legged/wheeled robots) and tasks (e.g., gripper vs. dexterous-hand manipulation). Key directions include: Modular architectures with shared physical priors for transferable dynamics learning; Meta-reinforcement learning for rapid adaptation to new hardware/task combinations; Object-centric representations enabling skill reuse across scenarios; and Sim-to-real bridges via residual physics modeling. These advances aim to create "one model fits all" solutions for scalable robotic intelligence.

\textbf{Hierarchical World Models for Long-Horizon Task}.
Future robotics world models will focus on hierarchical planning and temporal abstraction to handle complex, multi-stage tasks. Key advancements include: Goal-conditioned latent spaces for dynamic sub-task chaining; Memory-augmented transformers to capture long-term dependencies; Self-supervised skill discovery for reusable primitives; and Interactive human feedback for real-time plan adaptation. These innovations aim to bridge high-level reasoning with low-level control, enabling robust autonomy in open-ended environments.

\subsection{Challenges and Future Perspectives}
~\label{subsec:challeng_WM}

\textbf{High-Dimensionality and Partial Observability}.
Autonomous systems operate on high-dimensional sensory inputs, such as camera images, LiDAR point clouds, and radar signals. Modeling the world from this data imposes significant computational demands. Furthermore, these observations are inherently partial; the agent never perceives the complete state of the environment. This partial observability introduces uncertainty, necessitating robust state estimation techniques or the maintenance of a belief state over possible world states to inform decision-making.

\textbf{Causal Reasoning versus Correlation Learning}.
A fundamental limitation of many current world models is their proficiency at learning correlations rather than causal relationships. For instance, a model can learn that brake lights correlate with deceleration but may lack a deeper understanding of the underlying physics and driver intent. This deficiency hinders true generalization, as it prevents the model from performing counterfactual reasoning—evaluating "what if" scenarios that deviate from the training distribution. Achieving robust performance in novel situations requires a transition from correlational pattern matching to a genuine causal understanding of the environment.

\textbf{Abstract and Semantic Understanding}.
Effective world models must transcend low-level signal prediction and operate on a higher level of semantic and abstract understanding. A robust model should not merely predict future pixels or LiDAR points but also reason about abstract concepts. A major open problem lies in the fusion of fine-grained physical predictions with abstract reasoning about concepts such as traffic laws, pedestrian intent, and object affordances (e.g., that a chair is for sitting). Integrating these different levels of abstraction is crucial for intelligent and context-aware behavior.

\textbf{Systematic Evaluation and Benchmarking}.
The objective evaluation and comparison of world models present a significant research challenge. Conventional metrics like Mean Squared Error on future predictions are often insufficient, as they may not correlate with the performance of downstream tasks. A model generating visually sharper predictions might not necessarily enable a safer or more efficient control policy. The development of new evaluation frameworks is needed, with metrics that assess the model's utility for planning, its robustness in safety-critical scenarios, and its ability to capture causally relevant aspects of the environment.

\textbf{Memory Architecture and Long-Term Dependencies}.
Accurate long-term forecasting is notoriously difficult due to the compounding of prediction errors and the stochastic nature of the real world. A critical challenge is the design of memory architectures capable of retaining and retrieving relevant information over extended timescales, such as remembering a "Road Work Ahead" sign seen several minutes prior. The development of efficient and effective memory systems, leveraging architectures like Transformers or State-Space Models (SSMs), to manage these long-term dependencies remains an active and contested area of research.

\textbf{Human Interaction and Predictability}.
For agents operating in human-centric environments, a world model's role extends beyond mere environmental prediction. It must also facilitate agent behavior that is legible, predictable, and socially compliant to humans. Actions that are technically optimal but appear erratic or counter-intuitive can confuse human counterparts, such as other drivers or pedestrians, potentially leading to unsafe interactions. This social intelligence layer is a subtle but critical component of a functional world model.

\textbf{Interpretability and Verifiability}.
Deep learning-based world models are often opaque "black boxes," making it difficult to discern the rationale behind their predictions. For safety-critical applications like autonomous driving, the ability to audit and understand the model's internal decision-making process is non-negotiable, especially during post-incident analysis. Furthermore, a formidable theoretical and engineering challenge is the formal verification of these models—mathematically proving that they satisfy crucial safety properties (e.g., never hallucinating hazardous obstacles) across the vast space of possible inputs.

\textbf{Compositional Generalization and Abstraction}.
While the sim-to-real gap is a well-known generalization problem, a deeper challenge is compositional generalization. Humans can learn discrete concepts like "cup" and "table" and immediately generalize to novel compositions, such as "a cup on a table." In contrast, current models often require extensive exposure to specific compositional examples. An ideal world model should learn disentangled, abstract representations of entities, their relations, and their physical properties. This would enable it to understand and predict novel scenarios by composing known concepts, rather than relying on end-to-end pattern matching of entire scenes.

\textbf{Data Curation and Bias}.
The performance of a world model is fundamentally contingent on the quality and composition of its training data. Models inevitably inherit and may even amplify biases present in the dataset. For instance, a model trained predominantly on data from one geographical region may underperform in another with different road conventions or environmental conditions. A critical aspect of data curation is addressing the "long tail" of rare but safety-critical events. Systematically identifying, collecting, and ensuring the model learns effectively from these infrequent scenarios is essential for building robust and reliable systems.

\section{Conclusion}
\label{sec:conclusion}
This survey has provided a comprehensive examination of the critical role that physical simulators and world models play in advancing embodied artificial intelligence, revealing a transformative paradigm shift toward truly intelligent robotic systems. Through our analysis, we introduced a five-level classification framework (IR-L0 to IR-L4) for evaluating robot autonomy, conducted extensive comparative studies of mainstream simulation platforms, and explored the evolution of world models from simple recurrent architectures to sophisticated foundation-scale systems. Our investigation demonstrates how modern simulators like Isaac Gym, Genesis, and the emerging Newton platform have revolutionized robot learning through GPU-accelerated physics and photorealistic rendering, while advanced world models have enabled unprecedented capabilities in both autonomous driving and articulated robotics. 

These technologies has not only mitigated the sim-to-real gap but has also unlocked new possibilities for sample-efficient learning, long-horizon planning, and robust generalization across diverse environments. As we advance toward the realization of IR-L4 fully autonomous systems, the integration of physical simulators and world models represents the foundation upon which the next generation of embodied intelligence will be built, promising to transform robotics from task-specific automation to general-purpose intelligence capable of seamless integration into human society.

\bibliographystyle{IEEEtran}
\bibliography{levels,body,simulator,wm}

%{\small
%\bibliographystyle{unsrt2authabbrvpp}
%\bibliography{levels,body,simulator,wm}
%}

\end{document}